\def\paperID{14709}
\def\confName{CVPR}
\def\confYear{2025}

\def\paperTitle{ConsistentFeature: A Plug-and-Play Component for Neural Network Regularization}

\def\authorBlock{
    Ruizhe Jiang$^{1}$\thanks{Equal contribution} \qquad
    Haotian Lei$^{1}$\footnotemark[1] \qquad
    \\
    $^1$University of Chinese Academy of Science \\
    $^2$Sichuan University\\
    {\tt\small jiangruizhe22@mails.ucas.ac.cn} \qquad
    {\tt\small hao@stu.scu.edu.cn}\qquad
}

\newif\ifreview 
\newif\ifarxiv \newcommand{\arxiv}{\arxivtrue}
\newif\ifcamera 
\newif\ifrebuttal

\arxiv 

\pdfoutput=1
\documentclass[10pt,twocolumn,letterpaper]{article}
\usepackage{graphicx}
\usepackage{subcaption}

\ifreview \usepackage[review]{cvpr} \fi
\ifarxiv \usepackage[pagenumbers]{cvpr} \fi
\ifrebuttal \usepackage[rebuttal]{cvpr} \fi
\ifcamera \usepackage{cvpr} \fi

\usepackage{graphicx}	
\usepackage{amsmath}	
\usepackage{amssymb}	
\usepackage{booktabs}
\usepackage{times}
\usepackage{microtype}
\usepackage{epsfig}
\usepackage{caption}
\usepackage{float}
\usepackage{placeins}
\usepackage{color, colortbl}
\usepackage{stfloats}
\usepackage{enumitem}
\usepackage{tabularx}
\usepackage{xstring}
\usepackage{multirow}
\usepackage{xspace}
\usepackage{url}
\usepackage{subcaption}
\usepackage{xcolor}
\usepackage[hang,flushmargin]{footmisc}

\ifcamera \usepackage[accsupp]{axessibility} \fi

% ...

\ifarxiv  \fi

\newcommand{\R}[1]{{%
    \textbf{%
        \ifstrequal{#1}{1}{\textcolor{red}{R#1}}{%
        \ifstrequal{#1}{2}{\textcolor{blue}{R#1}}{%
        \ifstrequal{#1}{3}{\textcolor{magenta}{R#1}}{%
        \ifstrequal{#1}{4}{\textcolor{teal}{R#1}}{%
                           \textcolor{cyan}{R#1}%
        }}}}%
    }%
}}

\usepackage{xr-hyper}

\makeatletter
\newcommand*{\addFileDependency}[1]{
  \typeout{(#1)}
  \@addtofilelist{#1}
  \IfFileExists{#1}{}{\typeout{No file #1.}}
}

\makeatother
\newcommand*{\myexternaldocument}[1]{
    \externaldocument{#1}
    \addFileDependency{#1.tex}
    \addFileDependency{#1.aux}
}

\definecolor{cvprblue}{rgb}{0.21,0.49,0.74}
\usepackage[pagebackref,breaklinks,colorlinks,allcolors=cvprblue]{hyperref}
\usepackage[capitalize]{cleveref}
\crefname{section}{Sec.}{Secs.}
\crefname{table}{Table}{Tables}
\crefname{figure}{Fig.}{Figs.}

\ifarxiv \crefname{appendix}{App.}{Apps.}
\else \crefname{appendix}{Suppl.}{Suppls.} \fi

\unless\ifarxiv \myexternaldocument{_supplementary} \fi
\newtheorem{definition}{Definition}
\newtheorem{assumption}{Assumption}

\begin{document}
\title{\paperTitle}
\author{\authorBlock}
\maketitle

\begin{abstract}
Over-parameterized neural network models often lead to significant performance discrepancies between training and test sets, a phenomenon known as overfitting. To address this, researchers have proposed numerous regularization techniques tailored to various tasks and model architectures. In this paper, we introduce a simple perspective on overfitting: models learn different representations in different i.i.d. datasets. Based on this viewpoint, we propose an adaptive method, \textbf{ConsistentFeature}, that regularizes the model by constraining feature differences across random subsets of the same training set. Due to minimal prior assumptions, this approach is applicable to almost any architecture and task. Our experiments show that it effectively reduces overfitting, with low sensitivity to hyperparameters and minimal computational cost. It demonstrates particularly strong memory suppression and promotes normal convergence, even when the model has already started to overfit. \textbf{Even in the absence of significant overfitting, our method consistently improves accuracy and reduces validation loss.}
\end{abstract}

\section{Introduction}
\label{sec:intro}

\begin{figure}[ht]
    \centering
    \begin{subfigure}[b]{0.4\textwidth}
        \centering
        \includegraphics[width=\textwidth]{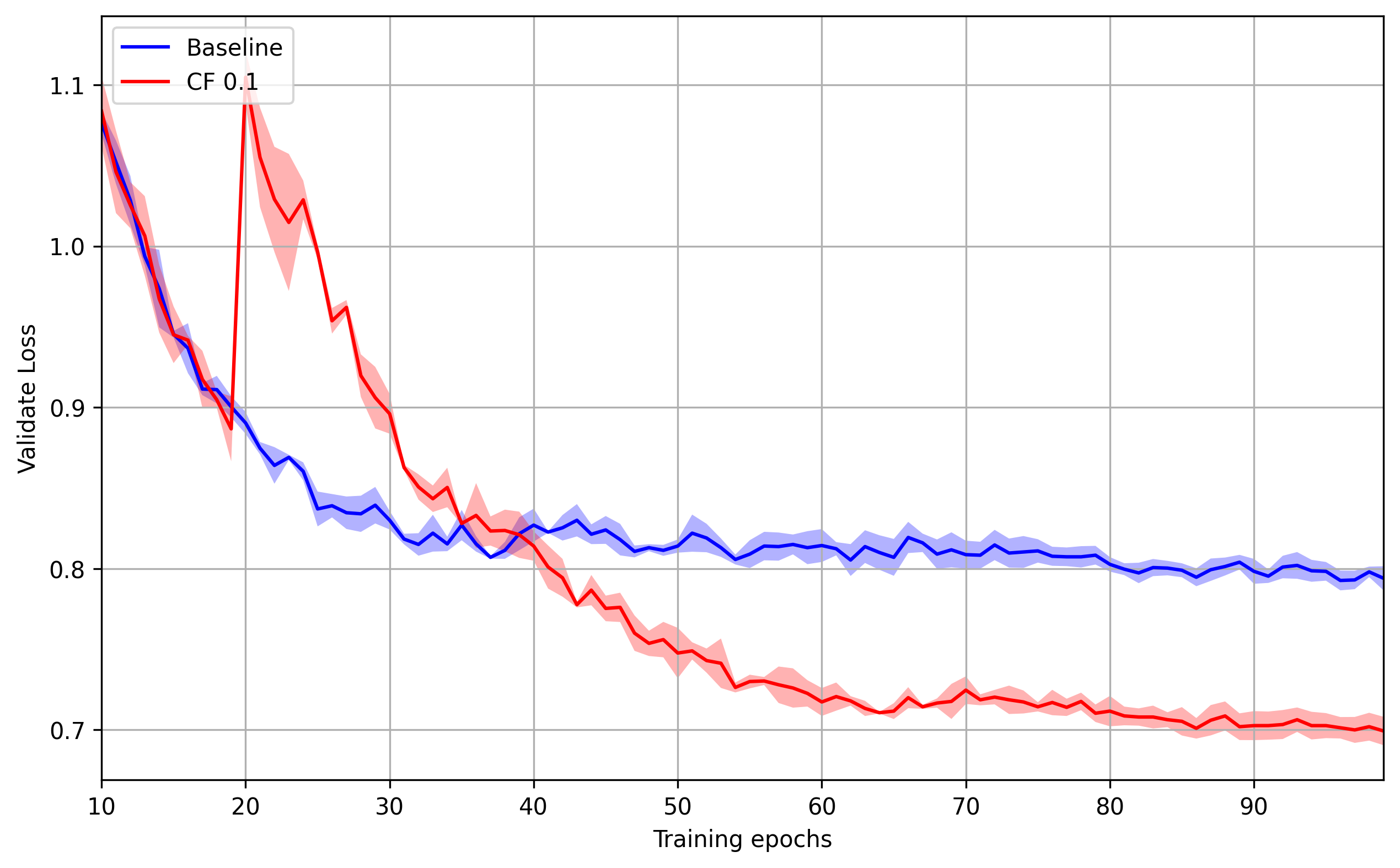}
        \caption{Validate Loss}
        \label{fig:shuffle_loss}
    \end{subfigure}
    \hfill
    \begin{subfigure}[b]{0.4\textwidth}
        \centering
        \includegraphics[width=\textwidth]{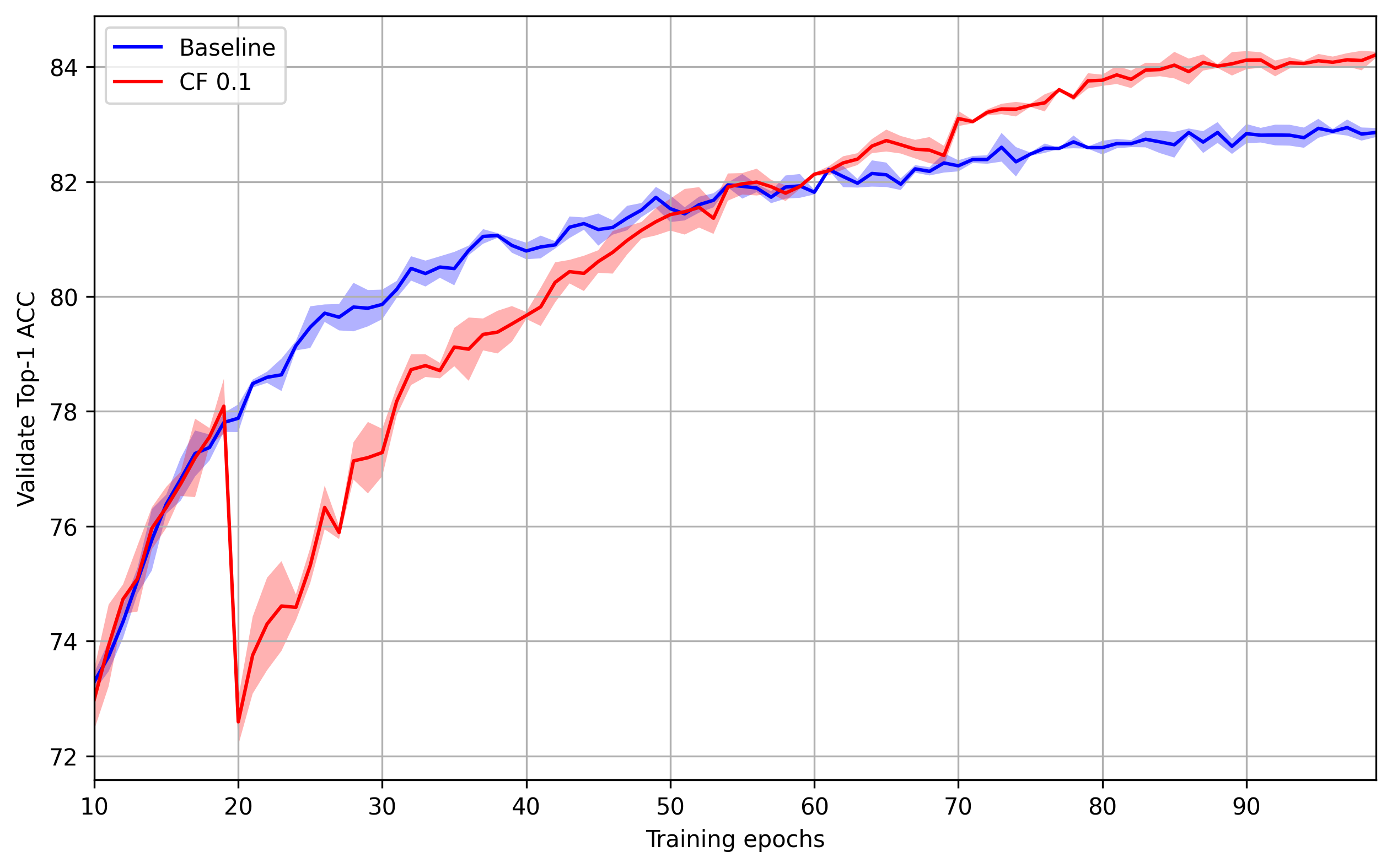}
        \caption{Top-1 ACC}
        \label{fig:shuffle_acc}
    \end{subfigure}
    \caption{Validate Loss/Top-1 ACC with and without Consistent Feature (CF) of ShuffleNetV2 on ImageNet200}
    \label{fig:shuffle_result}
\end{figure}

Deep learning-based image representation methods have made significant progress in recent years, with these advancements largely relying on increased hardware computing power, model architectures and sizes, as well as the scale of available datasets. However, when applied to certain practical tasks, especially those with limited data, large model architectures still face a notable performance gap. Overfitting is one of the key challenges in this context, and is widely recognized as being influenced by multiple factors\cite{zhang2021understanding}, ultimately leading to a misalignment between the knowledge learned by the model and the actual domain knowledge. Regularization methods, which now have a broader definition\cite{kukavcka2017regularization}, are no longer limited to just controlling model complexity. Data augmentation, normalization layers, and even certain optimizers all exhibit regularizing effects\cite{smith2021origin}. By incorporating prior constraints that cannot be directly derived from the data\cite{tian2022comprehensive}, regularization methods can significantly mitigate the overfitting problem.
\par
In recent years, prior assumptions in regularization methods are often specific to certain domains or model architectures. For example, \cite{mescheder2018training} explores the convergence and stability issues in Generative Adversarial Network (GAN)\cite{goodfellow2014generative} training, demonstrating the importance of absolute continuity. By penalizing the gradients of real data, this method prevents the discriminator from deviating from the Nash equilibrium, ensuring stable model training. \cite{merity2017revisiting} encourages stable training by penalizing differences between states that have been previously explored in Recurrent Neural Networks (RNNs). \cite{huang2016deep} further improves dropout by applying it to networks with residual structures, randomly discarding entire residual blocks, which stabilizes the training of networks with up to 1200 layers. In contrast, the method proposed in this work makes no assumptions based on model architecture or domain-specific priors. Inspired by the concept of "domain-invariant features" in domain adaptation\cite{singhal2023domain, ganin2015unsupervised, tzeng2014deep}, we propose a general explicit regularization method that, similar to weight decay\cite{lecun1989generalization}, label smoothing\cite{szegedy2016rethinking}, dropout\cite{JMLR:v15:srivastava14a}, and batch normalization\cite{ioffe2015batch}, can be applied to almost any supervised learning task.
\par
In this work, inspired by the feature constraints for domain invariance in transfer learning, we propose a novel explicit regularization method. This method adds an additional, independently updated discriminator head after the backbone, which randomly splits the original training data into two groups and labels them. Through adversarial updates between the discriminator head and the backbone, we limit the feature differences learned across different subsets of the training set, thereby encouraging the model to use more generalizable features for the corresponding tasks.

Compared to other general explicit regularization methods, our method offers the following advantages: \begin{itemize} \item \textbf{Suppresses overfitting while promoting normal convergence:} Our method exhibits a strong ability to suppress overfitting, particularly on small-scale and noisy datasets, while promoting the model's convergence to a normal state. Even in the absence of significant overfitting, it consistently improves accuracy and reduces validation loss. \item \textbf{Insensitive to hyperparameters:} This method exhibits minimal sensitivity to hyperparameters, ensuring stable performance across different model architectures and data scales. Unlike other regularization techniques, our method rarely suffers from performance degradation due to improper hyperparameter choices. \end{itemize}

To thoroughly understand the effectiveness and properties of our method, we conduct experiments across datasets of various sizes, different visual tasks, and models with varying parameter scales and architectures. We also provide some analysis of how the method affects the features learned by the model. Specifically, Our main contributions can be summarized as follows: \begin{itemize} \item We propose a nearly cost-free, general explicit regularization method: Unlike recent task- or architecture-specific regularization techniques, our method does not rely on any assumptions about the task or model structure. It is applicable to almost any task without requiring modifications to the original training process, offering broad applicability. Furthermore, our method uses a discriminator head with a minimal number of parameters, which has almost no impact on the training speed compared to training a standard model. \item We perform experiments across various data scales and model architectures: We evaluate and analyze the performance of the method under different conditions, including data aggregation sizes and model parameter scales and structures. \item We provide an explanation and analysis of the features learned by the model under this method, due to the adversarial strategy employed, which does not explicitly constrain any specific components. This analysis helps users and researchers gain a more intuitive understanding of the impacts brought by this method. \end{itemize}

\section{Related Work}
\label{sec:related}

\subsection{Domain Adaptation}
Deep neural network models, when trained on a dataset (source domain), often perform poorly when applied to another dataset with similar attributes but from a different domain (target domain). Domain Adaptation (DA) aims to address this issue. A key concept in classical domain adaptation methods is generating domain-invariant representations across the training sets. In some early studies, sample re-weighting algorithms\cite{gong2013connecting, huang2006correcting} were proposed to adjust the decision boundaries learned from the training samples, making them suitable for the target domain. Later, Deep Domain Confusion (DDC)\cite{tzeng2014deep} reduced the assumptions of early methods on the model's feature space and introduced an important idea: the features learned by the model should reside in a space that is devoid of domain-specific information but retains class (task-related) information. DDC uses two parallel networks, one with supervised loss (task-related) and the other as an unsupervised network. A domain confusion loss is used to penalize the feature differences between the domains. Maximum Mean Discrepancy (MMD)\cite{borgwardt2006integrating} is used as a measure of the domain loss's discrepancy. Building on this, work\cite{ganin2015unsupervised} introduced GAN-based methods into unsupervised domain adaptation, improving the model's performance on unsupervised data by constraining the feature differences between supervised and unsupervised data. The specific ideas and differences between this approach and ours are discussed in detail in Section 3. \cite{tzeng2017adversarial} further improved this by separating the training steps for supervised and unsupervised data. The method constrains the feature differences between the supervised data and the features learned from the unsupervised data in the replica model, encouraging the model to learn discriminative features on the unsupervised data. Initially, this method was used only for classification tasks, but later work extended it to image segmentation\cite{hoffman2016fcns} and natural language processing\cite{kim2017adversarial, xu2019adversarial, shen2018wasserstein} . Additionally, there are tasks in other domains that are similar to DA. For example, \cite{tang2024aug} combined data-free distillation with the training of an additional model to modify the knowledge distillation data, thus transferring the domain-invariant knowledge.

\subsection{Regularization}
Overfitting is a significant issue in supervised learning, and it can be influenced by factors such as dataset noise, model architecture, or model size. Initially, the overfitting problem was believed to be related to model complexity\cite{vapnik2015uniform}, and regularization methods were first used as a means of fine-tuning and constraining model complexity\cite{JMLR:v15:srivastava14a, girosi1995regularization, lecun1989generalization}. With the development of deep learning, the definition of regularization has been expanded. By summarizing and generalizing modern regularization techniques, \cite{kukavcka2017regularization} proposed that any method that enhances a model's generalization ability could be considered a regularization method. In recent years, regularization techniques have flourished, and based on whether they directly constrain the model's weights (i.e., by adding an extra penalty term after the target loss), regularization methods can be categorized into explicit and implicit regularization. Implicit regularization techniques, such as data augmentation \cite{zhong2020random}, normalization layers\cite{ioffe2015batch, lei2016layer, wu2018group}, and various noise injection methods\cite{welling2011bayesian, an1996effects}, are generally more universal, while recent explicit regularization methods focus on improvements specific to particular tasks or model architectures or enhance performance in a specific domain by incorporating domain priors\cite{tompson2015efficient, huang2016deep, fan2019reducing, mescheder2018training}.

In simple terms, our work applies ideas from domain adaptation to the regularization field. Inspired by the work of \cite{ganin2015unsupervised}, we propose a general explicit regularization method. To the best of our knowledge, \cite{abuduweili2021adaptive} presents the method most similar to ours, which aims to better leverage unlabeled data to improve the performance of knowledge distillation on pretrained models for a given dataset. This method introduced the ARC module, which aims to constrain the feature representations of the student model on both labeled and unlabeled data and uses MMD for discrimination. In contrast, our method focuses on improving the model's performance in supervised training. By randomly splitting the training set into two independent and identically distributed parts, we use an additional discriminator to constrain the feature differences between the two subsets and apply a GAN-based approach for updates.

\begin{figure}[ht]
    \centering
    \begin{subfigure}[b]{0.23\textwidth}
        \centering
        \includegraphics[width=\textwidth]{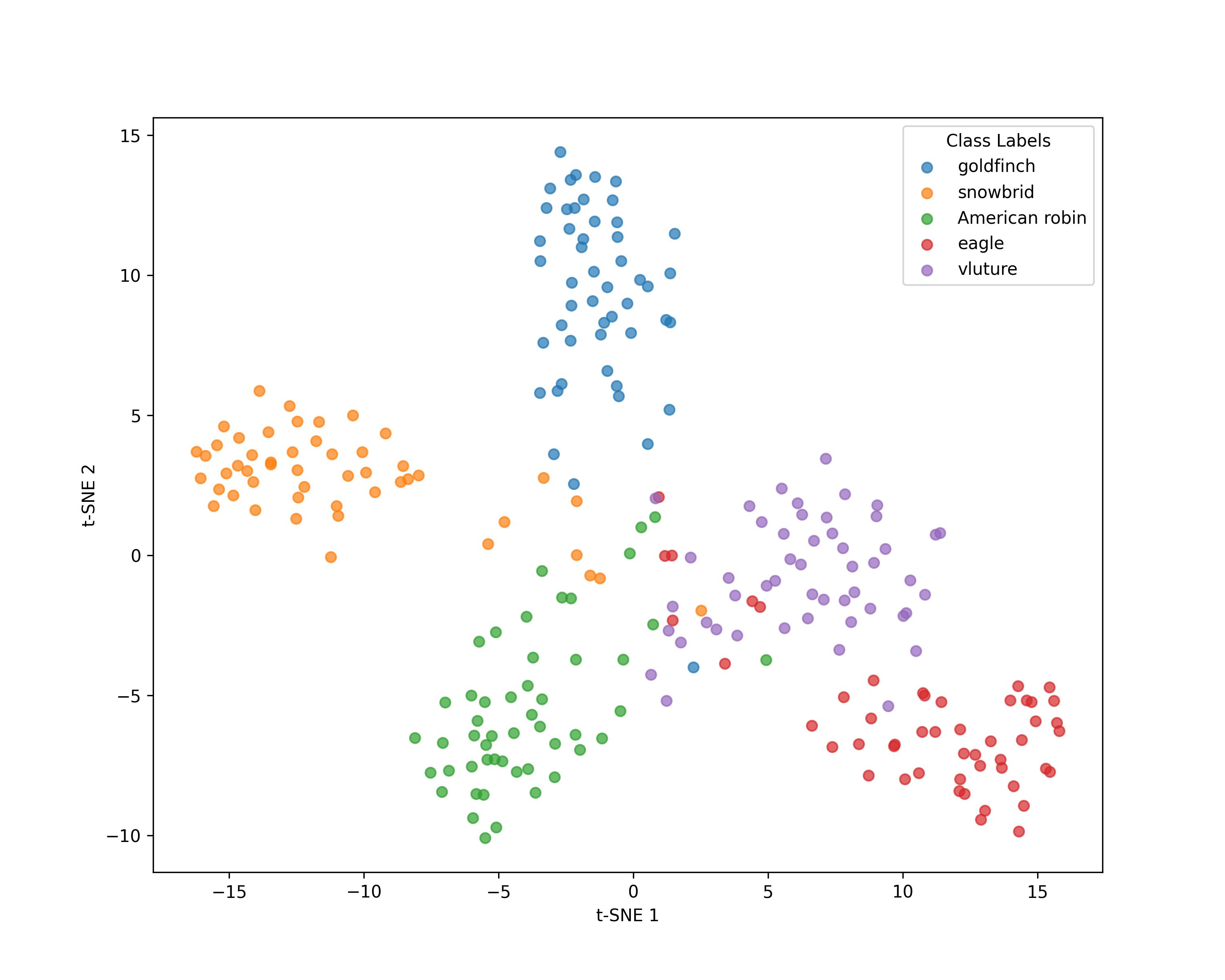}
        \caption{CF 0.1}
        \label{fig:tsne_sim_same}
    \end{subfigure}
    \begin{subfigure}[b]{0.23\textwidth}
        \centering
        \includegraphics[width=\textwidth]{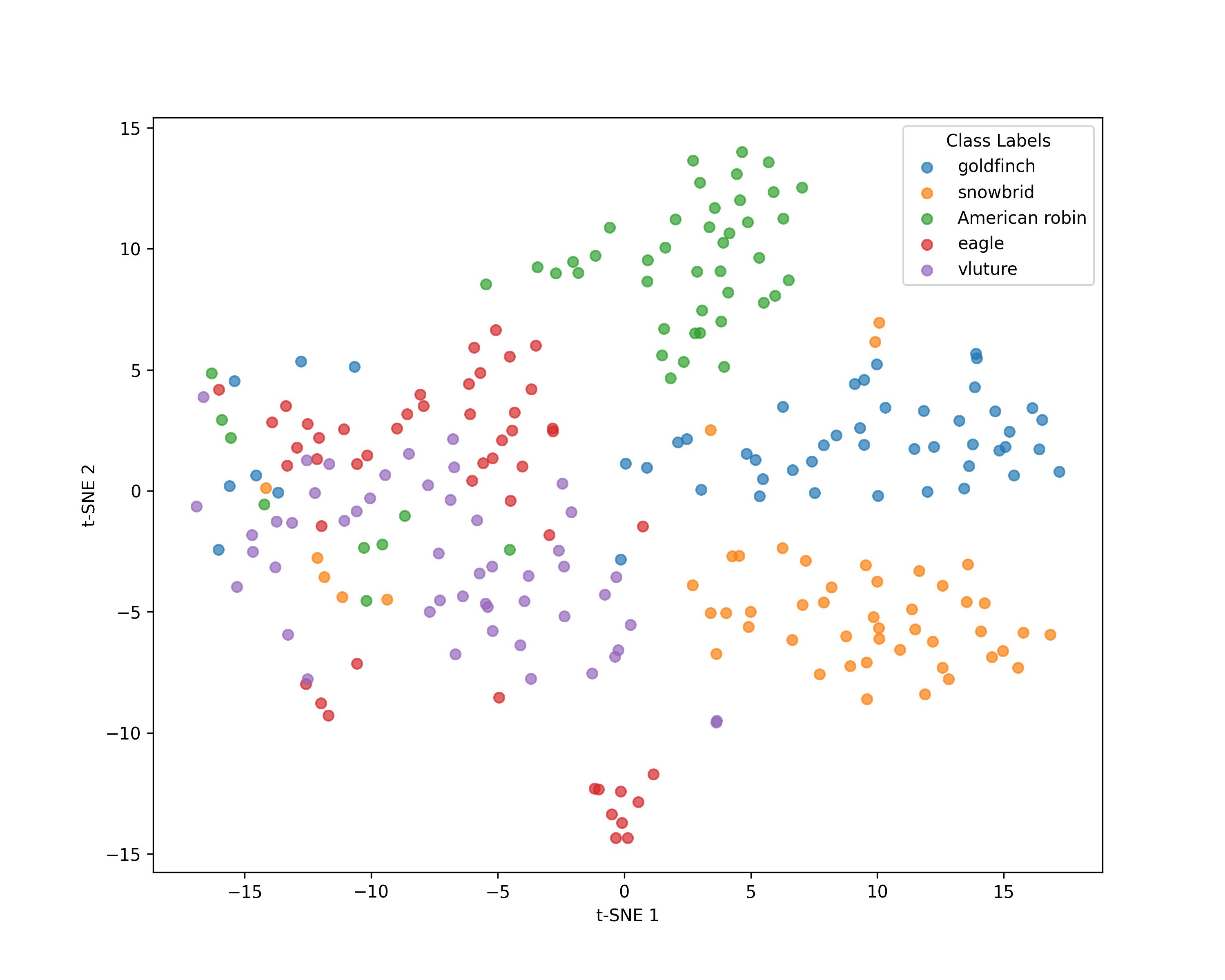}
        \caption{Baseline}
        \label{fig:tsne_sim_base}
    \end{subfigure}
    \vskip\baselineskip 

    \begin{subfigure}[b]{0.23\textwidth}
        \centering
        \includegraphics[width=\textwidth]{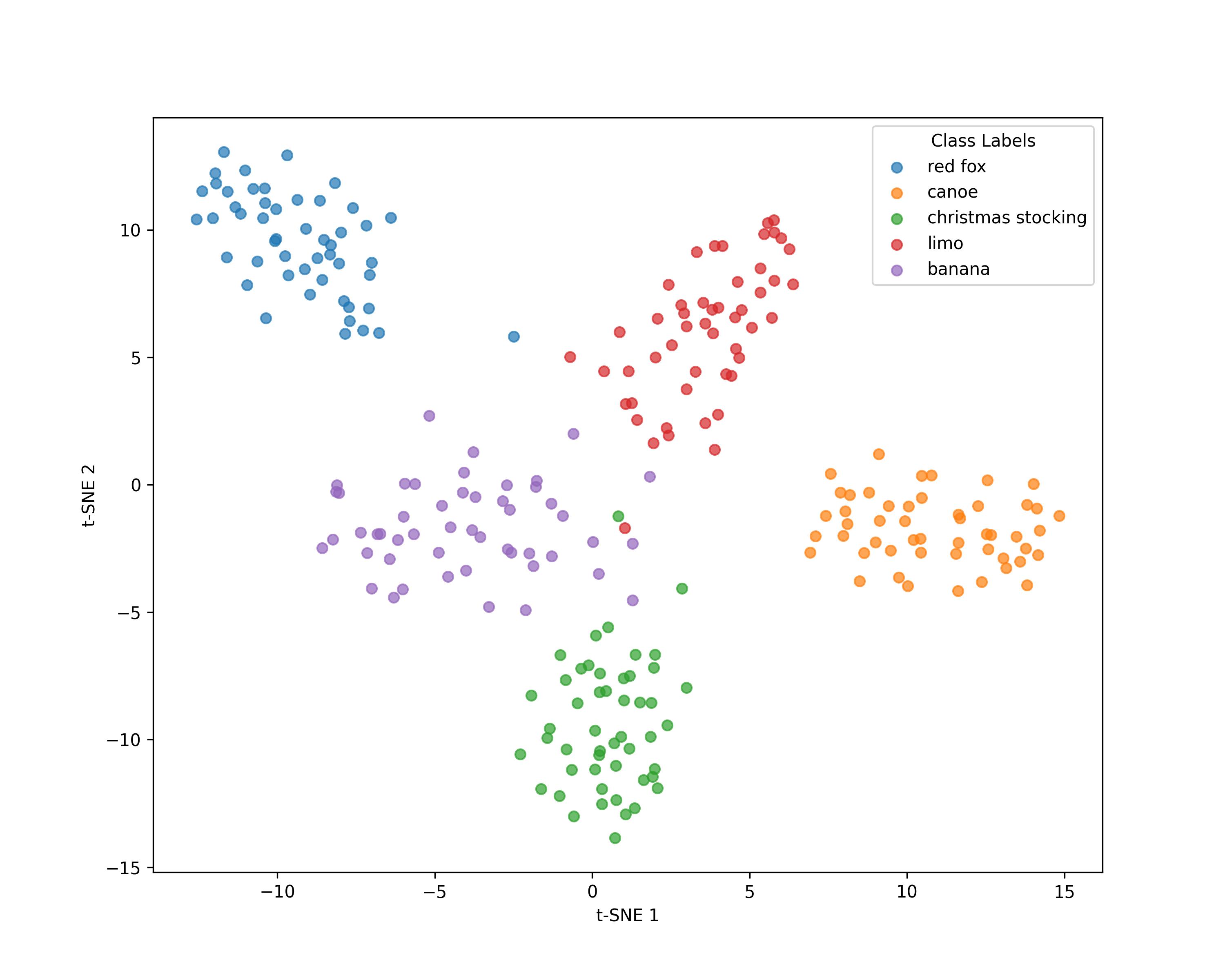}
        \caption{CF 0.1}
        \label{fig:tsne_diff_same}
    \end{subfigure}
    \begin{subfigure}[b]{0.23\textwidth}
        \centering
        \includegraphics[width=\textwidth]{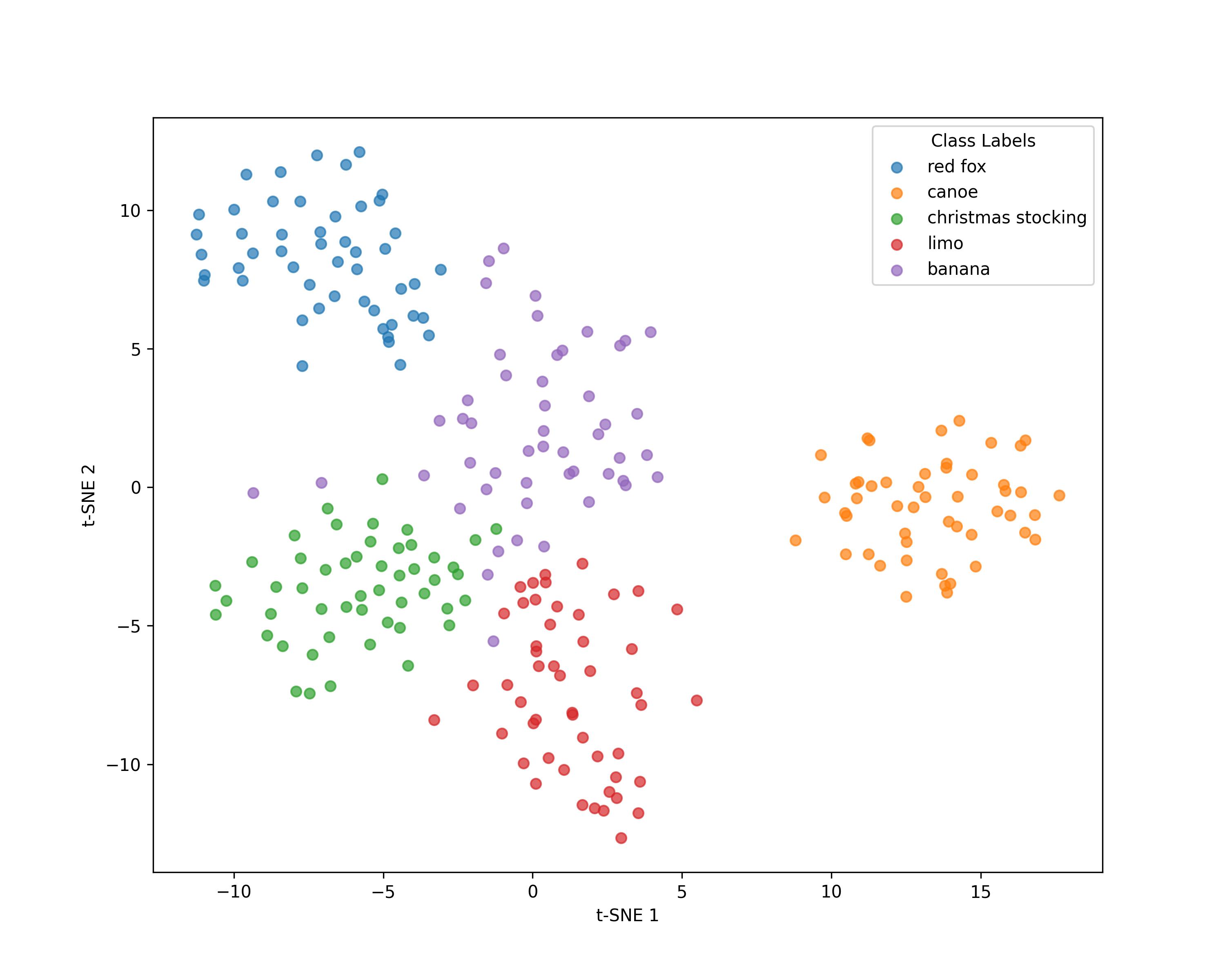}
        \caption{Baseline}
        \label{fig:tsne_diff_base}
    \end{subfigure}
    \caption{T-SNE Visualization of Semantically Similar Categories (a, b) / Semantically Unrelated Categories (c, d)}
    \label{fig:tsne_vis}
\end{figure}
\section{Method}
\subsection{Intuition}
Under the basic assumption of domain adaptation, \cite{ganin2015unsupervised} proposed a method called Domain Adversarial Neural Network (DANN), which aims to address the problem of unsupervised domain adaptation. This method consists primarily of three components: a feature extractor \( G \), a classifier \( C \), and a domain discriminator \( D \). Through these three components, DANN enables the model to learn similar features across both supervised and unsupervised datasets, ensuring the effectiveness of these features by measuring classification accuracy on the supervised data.

Inspired by this approach, we use the perspective of domain adaptation to address the problem of overfitting: if the training set and test set are treated as different domains, a model that learns domain-invariant and robust features should perform consistently across both the training and test sets. 

\begin{definition}
(Overfitting) Let \( S = \{\phi(X) \mid X \in \mathcal{X}\} \) be the feature space extracted by \(\phi\) from the input space \(\mathcal{X}\). The set \( S_{\text{train}} \) consists of patterns \textbf{unique to the training set}, defined as:
\begin{equation}
S_{\text{train}} = \left\{ z \mid \begin{array}{l}
\Pr_{X \sim \hat{P}_{\text{train}}}(\phi(X) = z) > 0 \\
\Pr_{X \sim \hat{P}_{\text{val}}}(\phi(X) = z) = 0
\end{array} \right\}
\end{equation}

These patterns are indicative of overfitting, as they do not generalize beyond the training set.We divide \(\phi\) into the following two parts:
\(\phi_{\text{train}}\) extracts only the patterns in \( S_{\text{train}} \), which are specific to the training set and lead to overfitting.
\(\phi_{\text{inv}}\) represents domain-invariant features, which are more generalizable.
Thus, the loss on the training set is lower than that on the validation set:
\begin{equation}
\begin{split}
\mathbb{E}_{(X, Y) \sim \hat{P}_{\text{train}}} [L(f(\phi(X)), Y)] < \\
\mathbb{E}_{(X, Y) \sim \hat{P}_{\text{val}}} [L(f(\phi(X)), Y)]
\end{split}
\end{equation}

where \( L \) refer to the loss function.
\end{definition}

Similarly, we make the following hypothesis: we sample three independent, identically distributed sets \( A \), \( B \), and \( C \) from the overall population. Suppose we use any two of these sets (here, \( A \) and \( B \)) as the training set, constraining the model to learn domain-invariant patterns between \( A \) and \( B \). This approach, compared to directly learning from \( A \) and \( B \) without such constraints, yields patterns with improved generalization on \( C \). More specifically:

\begin{assumption}
Given three datasets \( D_A \), \( D_B \), and \( D_C \) sampled from the overall distribution \( P(X, Y) \), and following (refer to Equation 1), the model can learn two types of features on each dataset: \( \phi_{\text{inv}} \) and \( \phi_{\text{data}} \), where \( \phi_{\text{data}} \) cannot generalize well to other i.i.d. sets. Additionally, \( \phi_{AB} \) represents patterns that belong exclusively to both \( D_A \) and \( D_B \), but not uniquely to either.

Consequently, the pattern learned directly on \( A \) and \( B \) is given by:
\begin{equation}
\phi_{\text{org}} = \phi_{\text{inv}} + \phi_A + \phi_B + \phi_{AB}.
\end{equation}
By penalizing the feature differences learned between \( A \) and \( B \), the model is forced to suppress \( \phi_A \) and \( \phi_B \), relying instead on a refined pattern:
\begin{equation}
\phi_{\text{reg}} = \phi_{\text{inv}} + \phi_{AB}
\end{equation}
to make predictions. Thus,
\begin{equation}
\begin{split}
\mathbb{E}_{(X, Y) \sim \hat{P}_{\text{C}}} \left[ L(f(\phi_{\text{reg}}(X)), Y) \right] \leq \\ 
\mathbb{E}_{(X, Y) \sim \hat{P}_{\text{C}}} \left[ L(f(\phi_{\text{org}}(X)), Y) \right]
\end{split}    
\end{equation}
indicating that the constrained model achieves better generalization on dataset \( C \).
\end{assumption}
Unlike the method in \cite{abuduweili2021adaptive}, our approach does not discard any supervision labels; instead, it simply adds a regularization term to constrain the feature distribution differences learned by the model on sets \( A \) and \( B \).

\begin{figure*}[ht]
    \centering
    \includegraphics[width=0.8\textwidth]{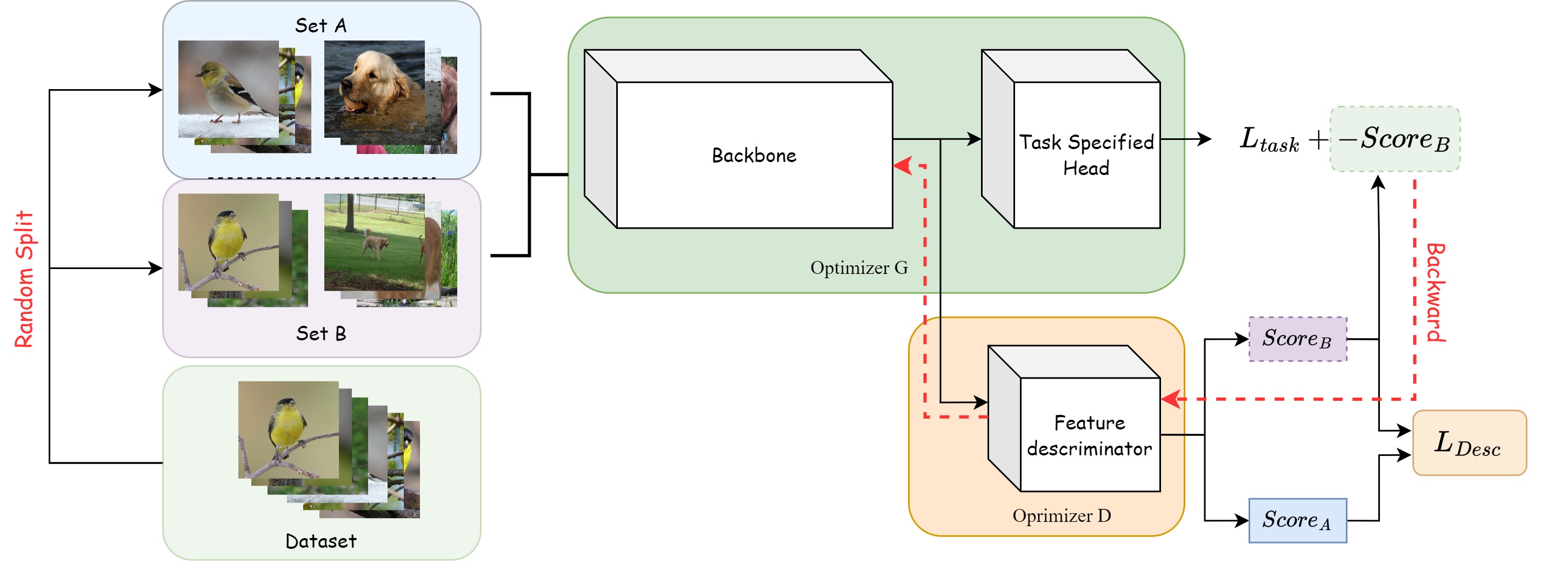} 
    \caption{Illustration of the Proposed Method. By randomly splitting the data into two subsets (i.e., $D_A$, $D_B$), the discriminator attempts to distinguish the data labels based on the model's feature outputs. Meanwhile, the model adversarially interacts with the discriminator using a subset of samples, thereby reducing the discernibility of the feature set.
} 
    \label{fig:method_illustration} 
\end{figure*}

\subsection{Proposed Approach}
In this work, we explicitly constrain the consistency of features obtained by the model across different i.i.d. sets. Specifically, we randomly and evenly divide any supervised training set into two subsets: \( D_A \) and \( D_B \). During the entire training process, we employ the approach from [15] and train an additional feature discriminator \( D_{\text{feature}} \) to distinguish between the feature differences learned by the model on \( D_A \) and \( D_B \). This difference is used as an explicit regularization term, which is added to the model's objective loss to encourage the model to utilize more generalizable features for the task, as illustrated in Figure\ref{fig:method_illustration}. Specifically:

Let \( \phi \) denote the feature extractor, \( H_{\text{task}} \) be the task-specific head, and \( H_{\text{desc}} \) be the discriminator head. The parameters \( \theta_G \) and \( \theta_D \) represent the weights for the backbone network and the discriminator head \( H_{\text{desc}} \), respectively.

Given two subsets \( D_A \) and \( D_B \) sampled from the training set \( D \), the optimization objectives are as follows:
\begin{equation}
\begin{split}
\underset{\theta_G, X \in D, X' \in D_B }{\arg \min} \, & \mathbb{E}\left[ L(H_{\text{task}}(\phi(X)), Y) + H_{\text{desc}}(\phi(X')) \right]
\end{split}
\end{equation}

\begin{equation}
\begin{split}
\underset{\theta_D, X \in D_A, X' \in D_B}{\arg \min} \, & \mathbb{E} \left[ L(H_{\text{desc}}(\phi(X)), H_{\text{desc}}(\phi(X'))) \right]
\end{split}
\end{equation}

where \( L \) represents the hinge loss, which encourages \( D_{\text{desc}}(\phi(X)) \) to approach 1 and \( D_{\text{desc}}(\phi(X')) \) to approach -1.
\par
In our approach, since the discriminator and generator are updated in sync at each step, we limit adversarial training to data $X' \in D_B$ to prevent the generator from overpowering the discriminator, which could otherwise lead to the discriminator providing uninformative feedback. This approach is validated in Table\ref{tab:param_sensitivity}. We observe a certain level of improvement as we increase $p$ (i.e., reduce the sample size of $D_B$).

\label{sec:method}

\section{Experiments} \label{sec:exps}

In this section, we first analyze the characteristics of the proposed method: we apply our approach across different model architectures and datasets, reporting results under various conditions. Next, we visualize the model features to further analyze and demonstrate the additional effects of our method, highlighting its consistency with our stated hypotheses. Finally, we compare our method with commonly used explicit regularization techniques under a standardized setting to examine similarities and differences in their properties.

\subsection{Experimental Setup}

To comprehensively reflect the behavior of our proposed method across various conditions, we selected mainstream model architectures, datasets of different sizes, and diverse tasks. For clarity, the datasets and metrics used may vary slightly across experimental sections, and detailed experimental settings will be provided in each specific experiment.
For the specific default parameters of our method, please refer to Figure \ref{tab:param_sensitivity}. In all experiments, our method will be referred to as \textbf{CF}.

\subsubsection{Datasets}

We used the following datasets across all experiments:

\begin{itemize}
    \item \textbf{ImageNet-A}: The ImageNet-A dataset, introduced in \cite{hendrycks2021natural}, is a subset of the ImageNet-1k dataset that includes additional challenging images across 200 subcategories. These images are intentionally selected to be misclassified by standard ImageNet models, providing a more difficult benchmark for evaluating model robustness and generalization.

    \item \textbf{ImageNet-200}: To more comprehensively validate the effects of our proposed method, we extracted 200 classes from ImageNet-1k, matching the categories of ImageNet-A. As a medium-scale dataset, it prevents the model from easily overfitting, allowing us to analyze the effects of our method when no significant overfitting occurs. Additionally, by training on this dataset, we can directly evaluate feature stability on ImageNet-A.

    \item \textbf{Flowers-102}: Flowers-102\cite{Nilsback08} is a high-resolution dataset containing 102 categories of flowers, with around 40–258 images per category. The dataset’s high resolution and relatively small size make it ideal for quickly evaluating our method’s performance on high-resolution data with limited samples.

    \item \textbf{WebVision-Mini}: WebVision\cite{li2017webvision} includes 2.4 million images across 1,000 classes and is widely used to evaluate methods for learning with noisy labels. Given the dataset's large size, we follow prior works \cite{chen2019understanding, chen2023sample, li2020dividemix} and use only the first 50 classes from the Google image subset for faster experimentation. The noise level in this subset is estimated to be around 20\% \cite{song2019does}.

    \item \textbf{CIFAR-100}: CIFAR-100\cite{krizhevsky2009learning} are widely used benchmark datasets for image classification, containing 32x32 color images.CIFAR-100 includes 100 classes with 600 images per class. These datasets are commonly used to assess model performance on small, diverse image sets and to test regularization effectiveness in preventing overfitting.In our experiments, to avoid modifying the original model's parameters, we resized the images to a resolution of 224.
\end{itemize}

\subsubsection{Optimization Strategy and Data Augmentation}

As noted in \cite{hernandez2018data, zhang2021understanding}, the SGD optimizer has implicit regularization effects, and its combination with explicit regularization methods may impact experimental results. To isolate the effects of our method and allow a fair comparison with other regularization techniques, we use the AdamW optimizer with a learning rate of \( \text{lr} = 1 \times 10^{-4} \) in all experiments, unless otherwise specified. Similarly, no data augmentation is applied apart from data normalization, unless explicitly noted.

\subsection{Cross-architecture Evaluation}
This experiment aims to analyze the effectiveness of our method across different model architectures and further demonstrate its performance in the absence of significant overfitting. In this experiment, we use the ImageNet200 dataset and a variety of popular architectures, ranging from models with 5.5M parameters (e.g., MobileNetV3 Large)\cite{howard2019searching} to those with 28.6M parameters (e.g., ConvNeXt Tiny)\cite{liu2022convnet}. Specifically, we employ data augmentation techniques such as\textbf{ \emph{RandomResizeCrop}} and \textbf{\emph{RandomFlip}}, as well as \textbf{\emph{CosineLearningRateAnnealing}} to minimize overfitting.

We present the training curves of ShuffleNet under our method in this setting, as shown in Figure\ref{fig:shuffle_result}, with detailed results reported in Table\ref{tab:cross_model}.

As observed, our method consistently improves performance across all architectures, achieving lower validation loss and higher accuracy without relying on any additional external information. This further demonstrates the robustness and compatibility of our approach. Additionally, we evaluate the loss on the ImageNet-A natural adversarial dataset under various methods. Our method exhibits stronger feature stability, achieving lower validation loss even on out-of-distribution (OOD) data. On average, our method reduces the validation loss by about 1 compared to the baseline.

To further investigate the impact of our method on model features, we perform t-SNE dimensionality reduction and visualization of ResNet18 model features across different categories in the validation set, as shown in Figure\ref{fig:tsne_vis}. In the ImageNet-200 validation set, we select semantically related categories and semantically unrelated categories for visualization of \textit{ResNet18 baseline} and \textit{ResNet18 with CF 0.1}.

It can be seen that for semantically unrelated categories (e.g., red fox, canoe, Christmas stocking, limo, and banana), both models show strong discrimination ability(Figure \ref{fig:tsne_diff_base}, \ref{fig:tsne_diff_same}). However, for semantically similar categories (e.g., goldfinch, snowbird, American robin, eagle, and vulture), the baseline model struggles to differentiate between them (Figure \ref{fig:tsne_sim_base}), while our method significantly improves the separability (Figure \ref{fig:tsne_sim_same}).

\begin{table*}[ht]
\centering
\renewcommand{\arraystretch}{1.2} 
\begin{tabular}{l c c c c c}
\hline
\textbf{Model} & \textbf{Validate Loss} & \textbf{Top-1 ACC} & \textbf{Top-5 ACC} & \textbf{Validate Loss on ImageNet-A} \\
\hline
ConvNext\_Tiny             & $0.97 \pm 0.00$ & $81.64 \pm 0.15$ & $92.57 \pm 0.05$ & $11.22 \pm 0.08$\\
\textbf{ConvNext\_Tiny + CF 0.1}    & $\mathbf{0.81 \pm 0.02}$ & $\mathbf{82.28 \pm 0.29}$ & $\mathbf{93.28 \pm 0.30}$ & $\mathbf{10.31 \pm 0.20}$\\
\hline
MobileNetV3\_Large         & $0.84 \pm 0.01$ & $81.73 \pm 0.13$ & $93.24 \pm 0.04$ & $8.61 \pm 0.04$\\
\textbf{MobileNetV3\_Large + CF 0.1} & $\mathbf{0.82 \pm 0.00}$ & $\mathbf{82.03 \pm 0.13}$ & $\mathbf{93.30 \pm 0.10}$ & $\mathbf{8.18 \pm 0.03}$\\
\hline
ResNet18                   & $0.89 \pm 0.02$ & $80.89 \pm 0.33$ & $92.83 \pm 0.46$ & $9.40 \pm 0.04$\\
\textbf{ResNet18 + CF 0.1}          & $\mathbf{0.79 \pm 0.02}$ & $\mathbf{81.98 \pm 0.59}$ & $\mathbf{93.30 \pm 0.19}$ & $\mathbf{8.43 \pm 0.11}$\\
\hline
ResNet50                   & $0.76 \pm 0.01$ & $84.32 \pm 0.04$ & $94.29 \pm 0.16$ & $10.34 \pm 0.09$\\
\textbf{ResNet50 + CF 0.1  }        & $\mathbf{0.67 \pm 0.01}$ & $\mathbf{84.95 \pm 0.51}$ & $\mathbf{94.91 \pm 0.26}$ & $\mathbf{9.10 \pm 0.23}$\\
\hline
ShuffleNetV2\_x2           & $0.79 \pm 0.00$ & $83.00 \pm 0.14$ & $93.76 \pm 0.10$ & $9.48 \pm 0.03$\\
\textbf{ShuffleNetV2\_x2 + CF 0.1}  & $\mathbf{0.69 \pm 0.00}$ & $\mathbf{84.23 \pm 0.08}$ & $\mathbf{94.55 \pm 0.07}$ & $\mathbf{8.51 \pm 0.06}$\\
\hline
\end{tabular}
\caption{Performance of our method across different model architectures on ImageNet200. Our method consistently achieves better performance on all models. The standard deviation is 0 because it is less than two decimal places.}
\label{tab:cross_model}
\end{table*}

\begin{figure}[ht]
    \centering
    \begin{minipage}{0.22\textwidth}
        \centering
        \includegraphics[width=\linewidth]{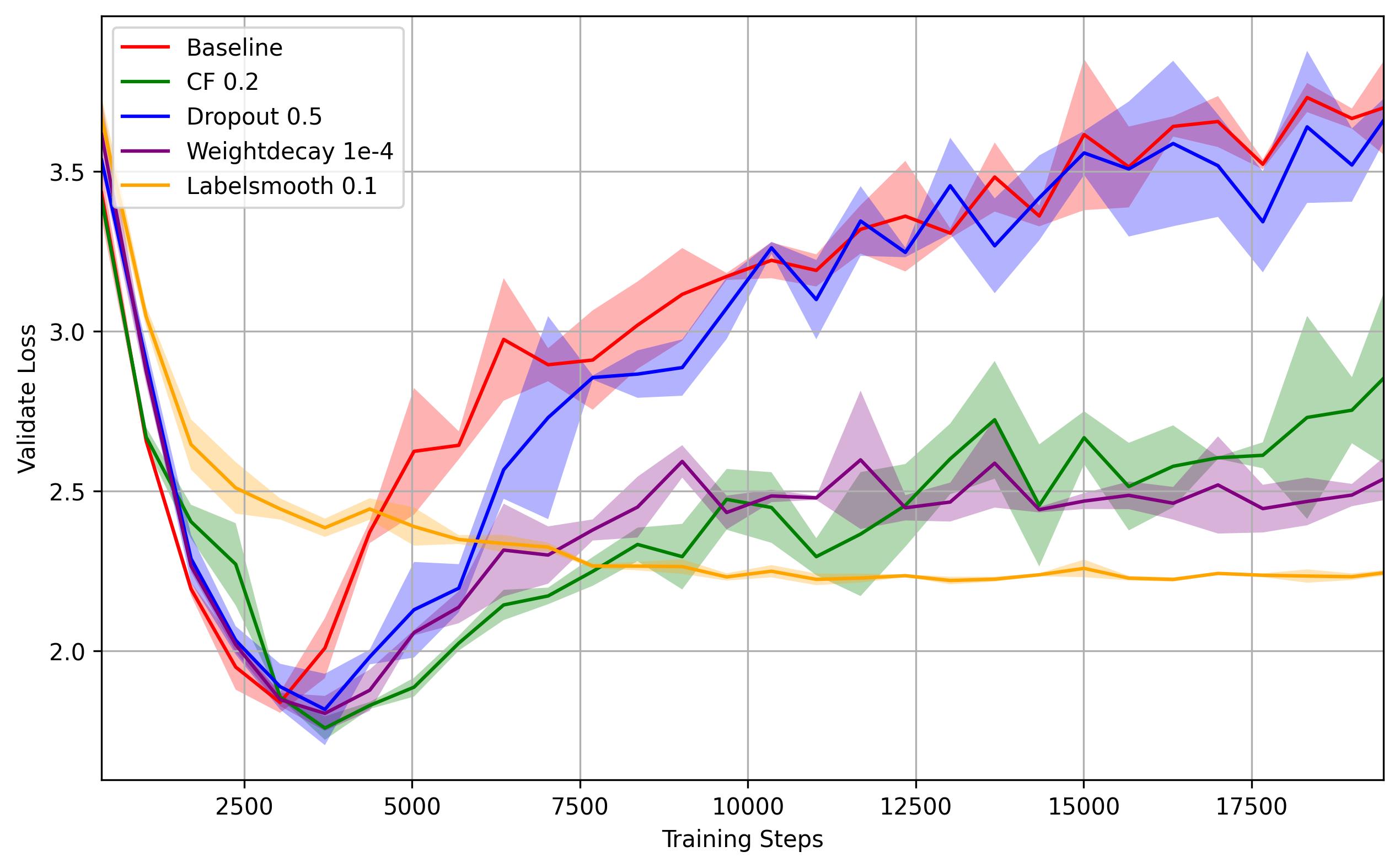} 
        \subcaption{Validate Loss on CIFAR-100.} 
        \label{fig:cifar_val}
    \end{minipage}
    \begin{minipage}{0.22\textwidth}
        \centering
        \includegraphics[width=\linewidth]{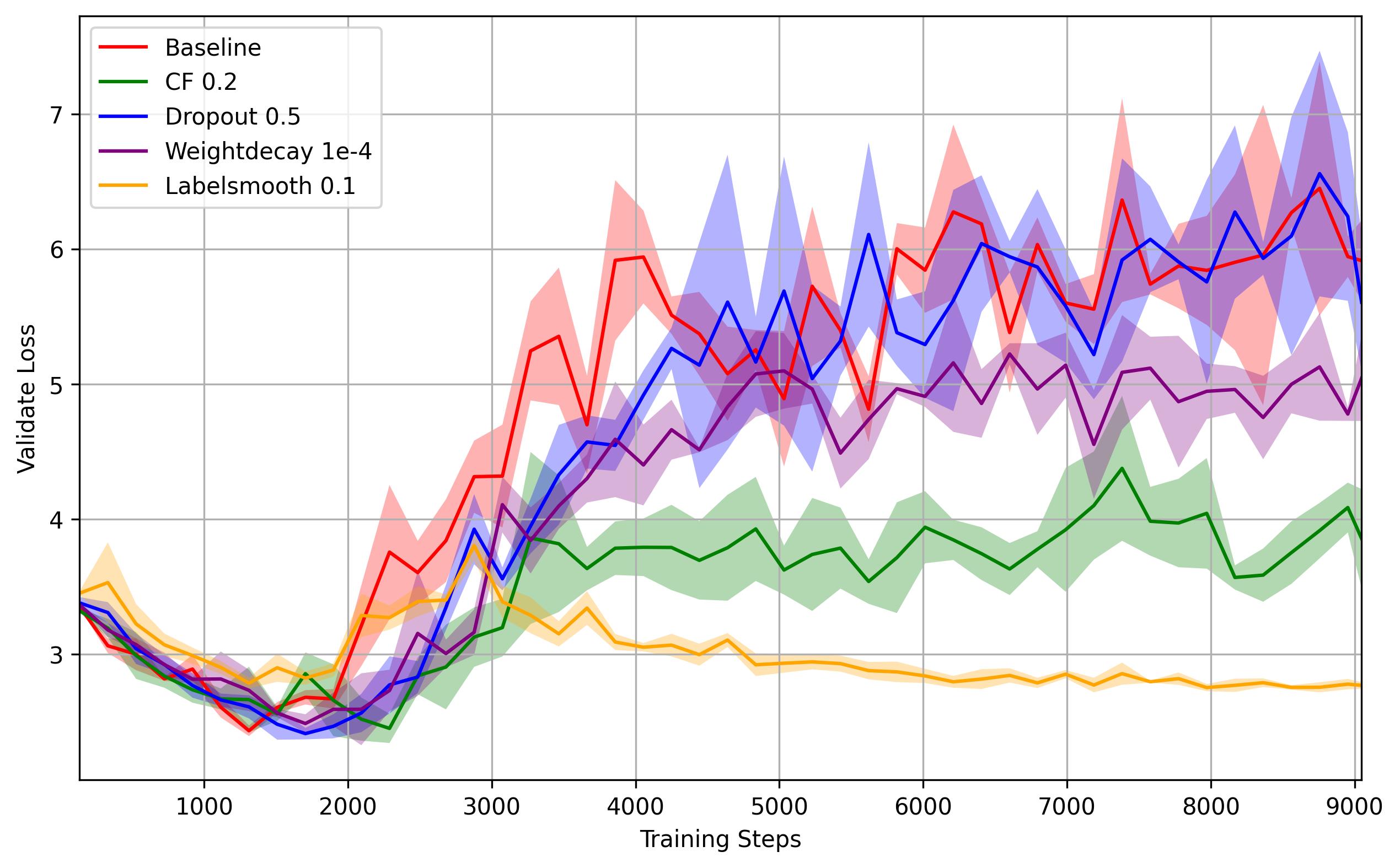} 
        \subcaption{Validate Loss on Webvision-mini.} 
        \label{fig:web_val}
    \end{minipage}

    \caption{Performance of Different Regularization Methods on CIFAR-100 and Webvision-mini.} 
    \label{fig:exp_analysis}
\end{figure}

\subsection{Regularization Empirical Analysis}
In this experiment, we compare our method with current mainstream explicit regularization strategies on both clean (CIFAR-100) and noisy (WebVision-mini) datasets. We use ResNet-18 as the baseline model and evaluate the effect of combining our method with different explicit regularization techniques to analyze the differences and compatibility between methods.

On CIFAR-100, as shown in Figures\ref{fig:cifar_val} and \ref{fig:web_val}, all regularization methods demonstrate some degree of overfitting suppression. Overall, our method yields results similar to weight decay, while label smoothing, which imposes constraints on the model’s embedding space\cite{muller2019does} achieves the highest accuracy and the most stable validation loss. Our method, however, achieves the lowest validation loss. The quantitative results can be found in Appendix A, Table\ref{tab:cifar100_performance}.

On the WebVision dataset, due to the presence of label noise, our method surpasses similar techniques like weight decay, achieving the lowest validation loss (Appendix A, Table\ref{tab:webvision_performance}), with the second-most stable validation loss, after label smoothing. Even without imposing any constraints on the model’s representation space, our method demonstrates excellent overfitting suppression.

\begin{figure*}[htbp]
    \centering
    \begin{subfigure}[b]{0.4\textwidth}
        \centering
        \includegraphics[width=\textwidth]{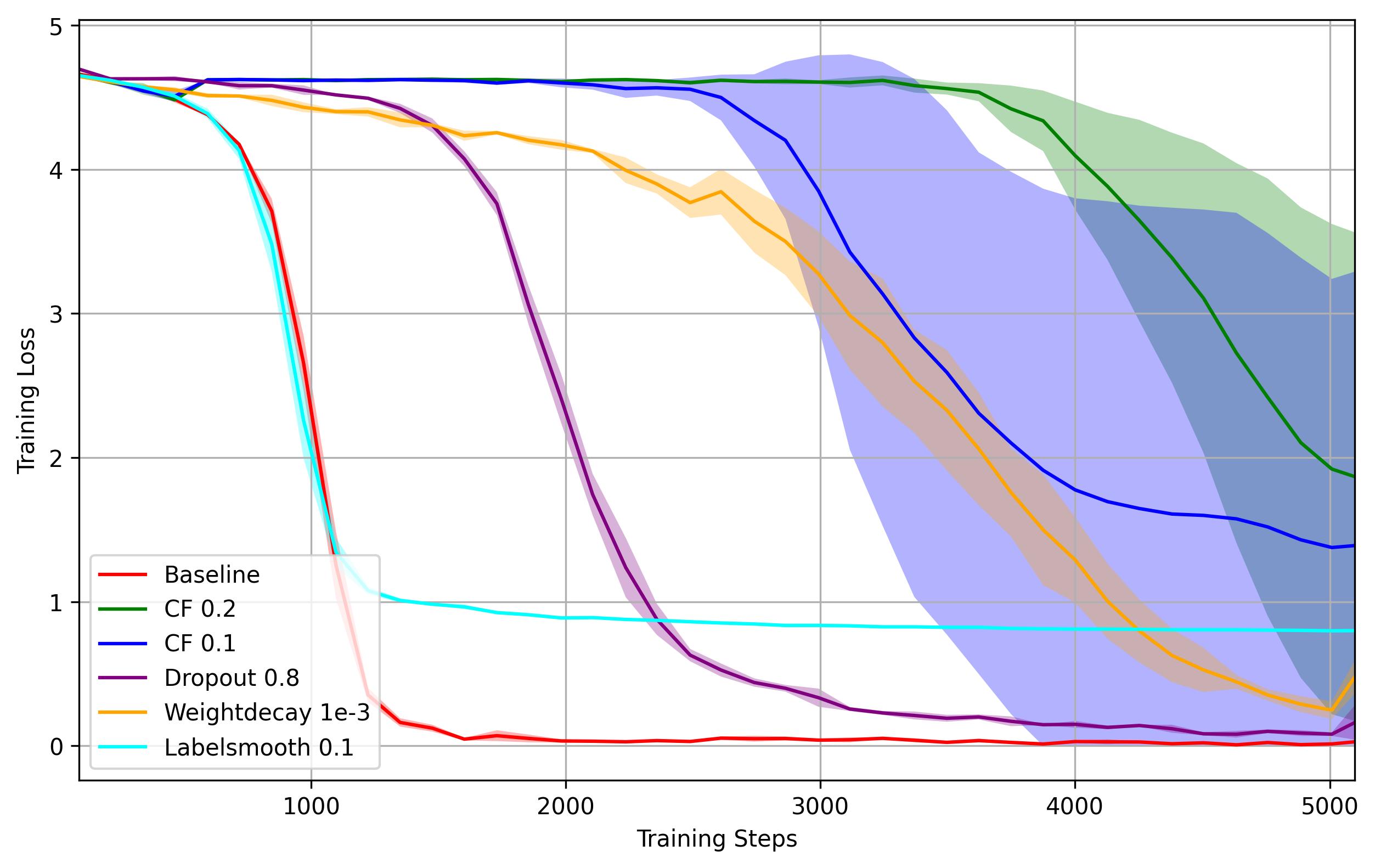} 
        \caption{} 
        \label{fig:memo_noise}
    \end{subfigure}
    \begin{subfigure}[b]{0.4\textwidth}
        \centering
        \includegraphics[width=\textwidth]{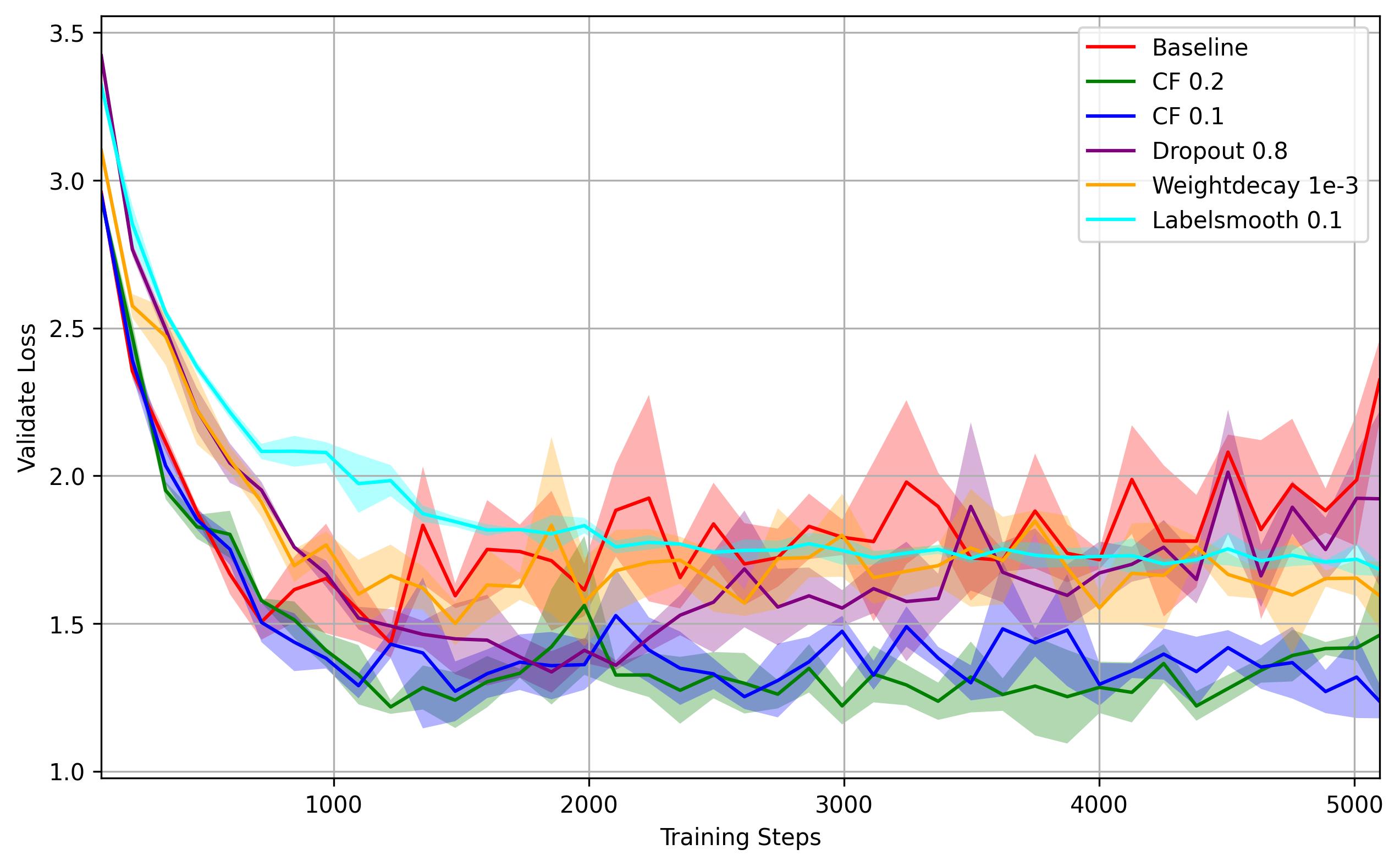} 
        \caption{} 
        \label{fig:memo_normal}
    \end{subfigure}
    \begin{subfigure}[b]{0.4\textwidth}
        \centering
        \includegraphics[width=\textwidth]{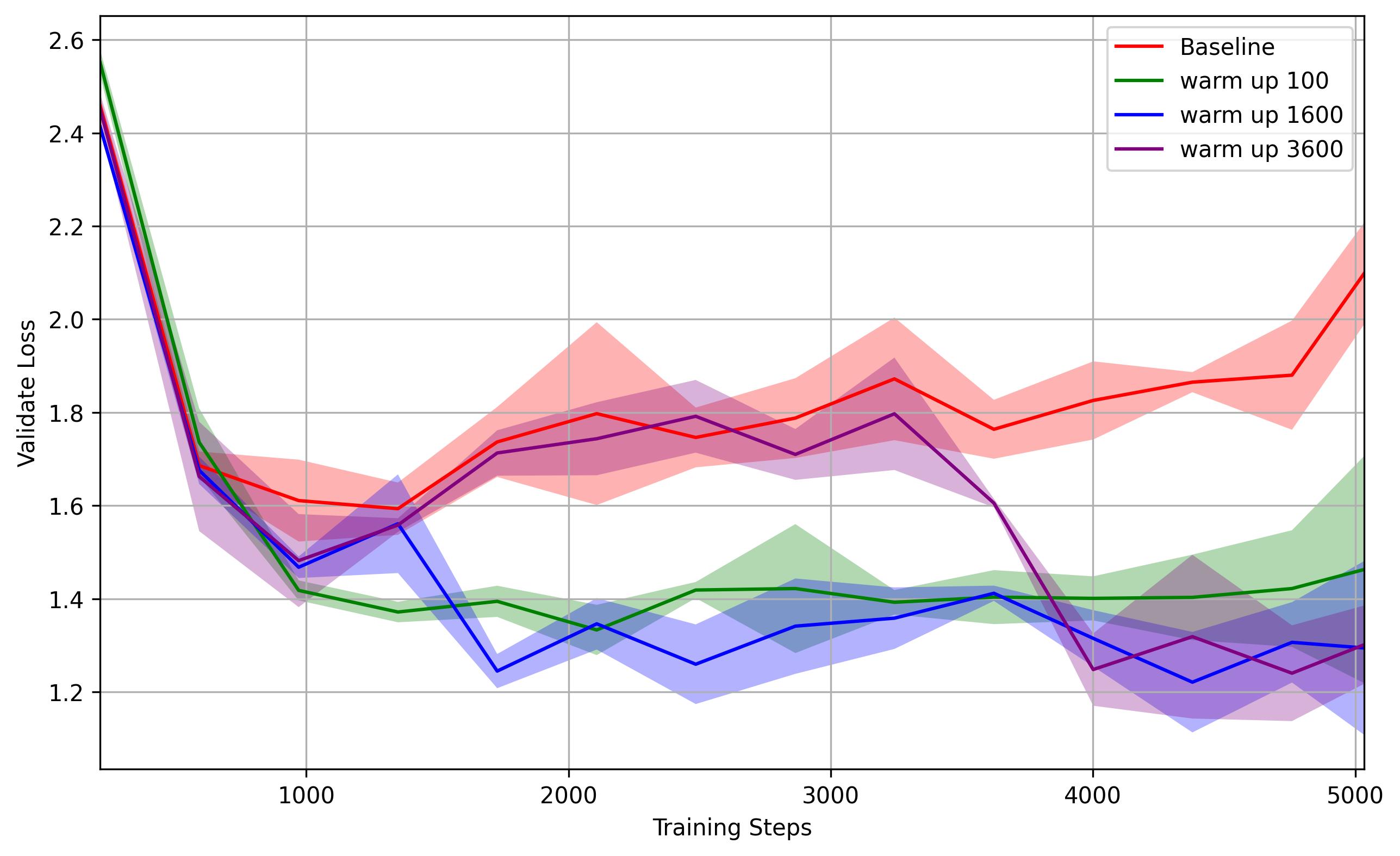} 
        \caption{} 
        \label{fig:warmup}
    \end{subfigure}
    \begin{subfigure}[b]{0.4\textwidth}
        \centering
        \includegraphics[width=\textwidth]{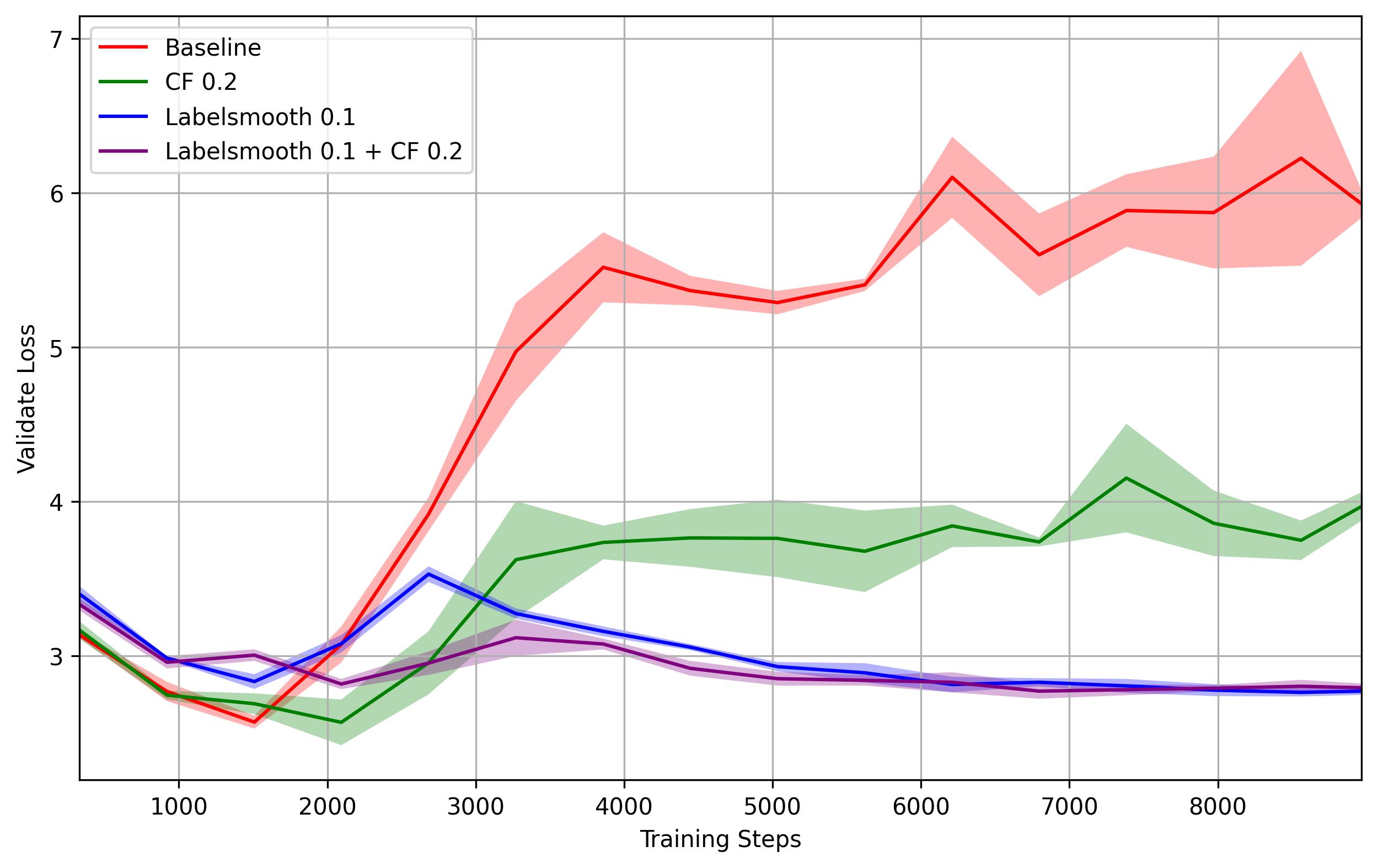} 
        \caption{} 
        \label{fig:addictive}
    \end{subfigure}
    
    \caption{Performance of Different Regularization Methods: (a) Training Loss (Memory) of Different Regularization Methods on Noisy Datasets, (b) Effect of Different Regularization Methods with the Same Hyperparameters on Convergence in the Normal Dataset (Validation Loss), (c) The Effect of Different Warm-up Steps on the Method. It is worth noting that even when the model has started overfitting before the intervention of our method, our approach is still able to return it to normal convergence, and (d) Performance of Our Method Combined with Label Smoothing on Webvision-mini.}
    \label{fig:memo_depress}
\end{figure*}

In summary, this experiment shows that our method provides the most stable overfitting control, balancing validation loss and accuracy, outperforming all methods except label smoothing.
Additionally, we analyzed the composability of our method, and some of the results are shown in Figure\ref{fig:addictive}. As can be seen, the regularization effect is enhanced to varying degrees when different methods are combined. For the complete curves, please refer to Figure 1 in Appendix A

\subsection{Generalization Methods for Regularization Comparison}

Generalization capacity is a metric used to assess a model's performance on both the training set and real-world data. Despite a large body of research offering theoretical explanations for model generalization, a unified definition remains elusive \cite{zhang2021understanding}. In works \cite{zhang2021understanding, arpit2017closer}, researchers controlled a model's generalization by randomizing the training set. In this setup, each sample is assigned a random label, with no inherent structure. The model can only progress through memorization of the training set, without generalizing to real-world distributions.

In \cite{arpit2017closer}, the regularization methods' ability to suppress the model's memorization behavior was used to quantify their effects. Specifically, if a method can suppress convergence on random data but does not negatively impact convergence on real data, the method is considered to enhance the model’s generalization ability.

To further compare the properties of our method with existing approaches, we adopted a similar setup: In this experiment, we used ResNet18 and the Flower102 dataset, generating a randomized dataset, Flower102\_random, by assigning random labels to each sample in the training set. By comparing the performance differences of various regularization methods on both the random data and normal data, we further compared their effects with our method.

In Figure \ref{fig:memo_depress}, we present the training loss curves of different methods on random data, along with the corresponding validation loss curves on real data using the same setup. Prior research suggests that a model initially learns simple patterns to explain the validation set, and then begins memorizing the data \cite{arpit2017closer}. As shown in Figure \ref{fig:memo_noise}, LabelSmoothing, which performed well previously, has almost no memory suppression ability on random data, but it enhances the model’s performance on normal data. On the other hand, Weight Decay overly restricts the model’s search space, leading to reduced generalization on real data (Figure \ref{fig:memo_normal}). Dropout, while having minimal effect on real data performance, also exhibits limited ability to suppress model memorization (even at a dropout rate of 0.8). Our method, however, not only suppresses the model's fitting (memory behavior) on noisy data almost entirely but also enhances the model's generalization ability on real data, resulting in the lowest validation loss (Figure \ref{fig:memo_normal}).

\subsubsection{Parameter Sensitivity Analysis}

Since our method only requires a random split of the dataset and the addition of an extra discriminator, there is limited space for ablation studies. However, hyperparameter selection is crucial for mainstream explicit regularization methods, as improper hyperparameter choices may either have no effect or negatively impact the model's convergence. Therefore, we conduct a variation of all hyperparameters in our proposed method to analyze the sensitivity of the method to different hyperparameter values.

In our method, technically, we use a discriminator structure similar to that in \cite{isola2017image} and apply the historical feature recording technique from \cite{metz2016unrolled} to stabilize the discriminator's training. Since our assumptions are not restricted to a specific type of model or discriminator, any technique could theoretically be used in practice to implement it.

Specifically, our method has the following adjustable parameters:

\begin{itemize}
    \item Dataset split ratio \textbf{$p$}: The dataset is randomly divided into two parts, with sizes \textbf{$p$} and \textbf{$1-p$}.
    \item Loss weight \textbf{$w$}: The weight applied to the regularization loss.
    \item History length \textbf{$history\_len$}: The length of the historical feature sequence used by the discriminator.
    \item Discriminator channel \textbf{$desc\_channel$}: This controls the complexity of the discriminator by adjusting the number of convolutional kernels.
    \item Discriminator warm-up time \textbf{$warm\_up$}: The duration before the loss starts influencing the model during training.
\end{itemize}

We perform a controlled variable analysis by adjusting each parameter individually. The effects of these parameters on the performance are shown in Table \ref{tab:param_sensitivity}. Specifically, val loss last 10 refers to the average validation loss over the last 10 epochs, which reflects the stability of the method's ability to suppress overfitting.

From the results, we observe that, with different parameter selections, accuracy and validation loss experience slight increases or decreases. Overall, smaller discriminators and larger \textbf{$p$} values (since fewer samples are used in the adversarial discriminator when \textbf{$p$} is larger) lead to some improvements, but the overall performance remains stable, which reflects the robustness of our method to varying parameters. Moreover, in our other experiments, we did not conduct extensive hyperparameter search, which sets our method apart from other explicit regularization techniques.

It is also noteworthy that when we varied the number of intervention steps, we found that \textbf{even when the model was overfitting, our method could still pull back the model's validation loss within a certain range}, as shown in Figure \ref{fig:warmup}.

\begin{table*}[htbp]
\centering
\begin{tabular}{c c c c c | c c c}
\hline
p    & desc\_channel & history\_len & warm\_up step & weight & Top-1 ACC    & Validate Loss   & Avg Val\_loss Last10 \\
\hline
n/a   & n/a            & n/a           & n/a            & n/a     & 74.7$\pm$0.73 & 1.26$\pm$0.05   & 1.97$\pm$0.04        \\
0.5  & 64            & 100          & 1600          & 0.1    & 74.3$\pm$0.89 & 0.98$\pm$0.06   & 1.42$\pm$0.01        \\
     &               &              &               & 0.2    & 75.1$\pm$0.60 & 0.99$\pm$0.03   & 1.27$\pm$0.10        \\
     &               &              &               & 0.5    & 74.7$\pm$0.83 & 0.97$\pm$0.02   & 1.31$\pm$0.09        \\
0.2  &               &              &               &        & 75.0$\pm$0.46 & 0.99$\pm$0.01   & 1.35$\pm$0.01        \\
0.5  &               &              &               &        & 74.2$\pm$0.84 & 1.00$\pm$0.03   & 1.38$\pm$0.06        \\
0.8  &               &              &               &        & 76.5$\pm$0.18 & 0.95$\pm$0.01   & 1.05$\pm$0.03        \\
     &               &              & 100           &        & 73.6$\pm$0.38 & 1.08$\pm$0.02   & 1.44$\pm$0.16        \\
     &               &              & 1600          &        & 74.8$\pm$0.76 & 0.98$\pm$0.01   & 1.29$\pm$0.17        \\
     &               &              & 3600          &        & 75.9$\pm$1.20 & 0.94$\pm$0.04   & 1.29$\pm$0.04        \\
     &               & 10           &               &        & 73.3$\pm$0.31 & 1.04$\pm$0.02   & 1.48$\pm$0.13        \\
     &               & 100          &               &        & 74.6$\pm$1.4  & 1.03$\pm$0.04   & 1.36$\pm$0.12        \\
     &               & 500          &               &        & 74.2$\pm$0.48 & 0.99$\pm$0.02   & 1.43$\pm$0.12        \\
     & 32            &              &               &        & 75.5$\pm$0.13 & 0.98$\pm$0.02   & 1.40$\pm$0.09        \\
     & 64            &              &               &        & 74.6$\pm$0.71 & 1.00$\pm$0.03   & 1.48$\pm$0.19        \\
     & 128           &              &               &        & 74.1$\pm$0.44 & 0.99$\pm$0.03   & 1.32$\pm$0.08        \\
\hline
\end{tabular}
\caption{Sensitivity of the Proposed Method to Different Hyperparameters. The first row corresponds to the baseline, where the proposed method is not applied. The second row shows the default hyperparameters used in the proposed method.}

\label{tab:param_sensitivity}
\end{table*}

\section{Conclusion}
\label{sec:conclusion}

In this work, we propose an explicit regularization method for supervised learning, inspired by the concept of domain-invariant features from domain adaptation techniques. We validate the proposed method across different model architectures, and consistently achieve improved results without significant overfitting. Additionally, we use feature dimensionality reduction and visualization to demonstrate that our method helps the model learn better features.

Subsequently, we compare our method with commonly used regularization techniques from different perspectives, and experiments show that our approach has a remarkable ability to suppress memorization and promotes the proper convergence of the model, aligning with our initial hypothesis. We also demonstrate that our method can be combined with other regularization methods to achieve further improvements.

We have conducted a limited parameter search for our method, which shows that it is not sensitive to hyperparameter choices, making it one of the key advantages of our approach. Moreover, even when the model has already started to overfit, our method is still able to restore the model to normal convergence.

Despite these contributions, there are still limitations in our work. Since we use an adversarial model to dynamically constrain the model's features, we cannot exactly identify which features the discriminator is suppressing or promoting. While some of our method's performance aligns with our theoretical assumptions, further analysis is required to fully understand its behavior.

{\small
\bibliographystyle{ieeenat_fullname}
\bibliography{11_references}

\begin{thebibliography}{49}
\providecommand{\natexlab}[1]{#1}
\providecommand{\url}[1]{\texttt{#1}}
\expandafter\ifx\csname urlstyle\endcsname\relax
  \providecommand{\doi}[1]{doi: #1}\else
  \providecommand{\doi}{doi: \begingroup \urlstyle{rm}\Url}\fi

\bibitem[Abuduweili et~al.(2021)Abuduweili, Li, Shi, Xu, and Dou]{abuduweili2021adaptive}
Abulikemu Abuduweili, Xingjian Li, Humphrey Shi, Cheng-Zhong Xu, and Dejing Dou.
\newblock Adaptive consistency regularization for semi-supervised transfer learning.
\newblock In \emph{Proceedings of the IEEE/CVF conference on computer vision and pattern recognition}, pages 6923--6932, 2021.

\bibitem[An(1996)]{an1996effects}
Guozhong An.
\newblock The effects of adding noise during backpropagation training on a generalization performance.
\newblock \emph{Neural computation}, 8\penalty0 (3):\penalty0 643--674, 1996.

\bibitem[Arpit et~al.(2017)Arpit, Jastrzebski, Ballas, Krueger, Bengio, Kanwal, Maharaj, Fischer, Courville, Bengio, et~al.]{arpit2017closer}
Devansh Arpit, Stanis{\l}aw Jastrzebski, Nicolas Ballas, David Krueger, Emmanuel Bengio, Maxinder~S Kanwal, Tegan Maharaj, Asja Fischer, Aaron Courville, Yoshua Bengio, et~al.
\newblock A closer look at memorization in deep networks.
\newblock In \emph{International conference on machine learning}, pages 233--242. PMLR, 2017.

\bibitem[Borgwardt et~al.(2006)Borgwardt, Gretton, Rasch, Kriegel, Sch{\"o}lkopf, and Smola]{borgwardt2006integrating}
Karsten~M Borgwardt, Arthur Gretton, Malte~J Rasch, Hans-Peter Kriegel, Bernhard Sch{\"o}lkopf, and Alex~J Smola.
\newblock Integrating structured biological data by kernel maximum mean discrepancy.
\newblock \emph{Bioinformatics}, 22\penalty0 (14):\penalty0 e49--e57, 2006.

\bibitem[Chen et~al.(2019)Chen, Liao, Chen, and Zhang]{chen2019understanding}
Pengfei Chen, Ben~Ben Liao, Guangyong Chen, and Shengyu Zhang.
\newblock Understanding and utilizing deep neural networks trained with noisy labels.
\newblock In \emph{International conference on machine learning}, pages 1062--1070. PMLR, 2019.

\bibitem[Chen et~al.(2023)Chen, Zhu, and Li]{chen2023sample}
Wenkai Chen, Chuang Zhu, and Mengting Li.
\newblock Sample prior guided robust model learning to suppress noisy labels.
\newblock In \emph{Joint European Conference on Machine Learning and Knowledge Discovery in Databases}, pages 3--19. Springer, 2023.

\bibitem[Fan et~al.(2019)Fan, Grave, and Joulin]{fan2019reducing}
Angela Fan, Edouard Grave, and Armand Joulin.
\newblock Reducing transformer depth on demand with structured dropout.
\newblock \emph{arXiv preprint arXiv:1909.11556}, 2019.

\bibitem[Ganin and Lempitsky(2015)]{ganin2015unsupervised}
Yaroslav Ganin and Victor Lempitsky.
\newblock Unsupervised domain adaptation by backpropagation.
\newblock In \emph{International conference on machine learning}, pages 1180--1189. PMLR, 2015.

\bibitem[Girosi et~al.(1995)Girosi, Jones, and Poggio]{girosi1995regularization}
Federico Girosi, Michael Jones, and Tomaso Poggio.
\newblock Regularization theory and neural networks architectures.
\newblock \emph{Neural computation}, 7\penalty0 (2):\penalty0 219--269, 1995.

\bibitem[Gong et~al.(2013)Gong, Grauman, and Sha]{gong2013connecting}
Boqing Gong, Kristen Grauman, and Fei Sha.
\newblock Connecting the dots with landmarks: Discriminatively learning domain-invariant features for unsupervised domain adaptation.
\newblock In \emph{International conference on machine learning}, pages 222--230. PMLR, 2013.

\bibitem[Goodfellow et~al.(2014)Goodfellow, Pouget-Abadie, Mirza, Xu, Warde-Farley, Ozair, Courville, and Bengio]{goodfellow2014generative}
Ian Goodfellow, Jean Pouget-Abadie, Mehdi Mirza, Bing Xu, David Warde-Farley, Sherjil Ozair, Aaron Courville, and Yoshua Bengio.
\newblock Generative adversarial nets.
\newblock \emph{Advances in neural information processing systems}, 27, 2014.

\bibitem[Hendrycks et~al.(2021)Hendrycks, Zhao, Basart, Steinhardt, and Song]{hendrycks2021natural}
Dan Hendrycks, Kevin Zhao, Steven Basart, Jacob Steinhardt, and Dawn Song.
\newblock Natural adversarial examples.
\newblock In \emph{Proceedings of the IEEE/CVF conference on computer vision and pattern recognition}, pages 15262--15271, 2021.

\bibitem[Hern{\'a}ndez-Garc{\'\i}a and K{\"o}nig(2018)]{hernandez2018data}
Alex Hern{\'a}ndez-Garc{\'\i}a and Peter K{\"o}nig.
\newblock Data augmentation instead of explicit regularization.
\newblock \emph{arXiv preprint arXiv:1806.03852}, 2018.

\bibitem[Hoffman et~al.(2016)Hoffman, Wang, Yu, and Darrell]{hoffman2016fcns}
Judy Hoffman, Dequan Wang, Fisher Yu, and Trevor Darrell.
\newblock Fcns in the wild: Pixel-level adversarial and constraint-based adaptation.
\newblock \emph{arXiv preprint arXiv:1612.02649}, 2016.

\bibitem[Howard et~al.(2019)Howard, Sandler, Chu, Chen, Chen, Tan, Wang, Zhu, Pang, Vasudevan, et~al.]{howard2019searching}
Andrew Howard, Mark Sandler, Grace Chu, Liang-Chieh Chen, Bo Chen, Mingxing Tan, Weijun Wang, Yukun Zhu, Ruoming Pang, Vijay Vasudevan, et~al.
\newblock Searching for mobilenetv3.
\newblock In \emph{Proceedings of the IEEE/CVF international conference on computer vision}, pages 1314--1324, 2019.

\bibitem[Huang et~al.(2016)Huang, Sun, Liu, Sedra, and Weinberger]{huang2016deep}
Gao Huang, Yu Sun, Zhuang Liu, Daniel Sedra, and Kilian~Q Weinberger.
\newblock Deep networks with stochastic depth.
\newblock In \emph{Computer Vision--ECCV 2016: 14th European Conference, Amsterdam, The Netherlands, October 11--14, 2016, Proceedings, Part IV 14}, pages 646--661. Springer, 2016.

\bibitem[Huang et~al.(2006)Huang, Gretton, Borgwardt, Sch{\"o}lkopf, and Smola]{huang2006correcting}
Jiayuan Huang, Arthur Gretton, Karsten Borgwardt, Bernhard Sch{\"o}lkopf, and Alex Smola.
\newblock Correcting sample selection bias by unlabeled data.
\newblock \emph{Advances in neural information processing systems}, 19, 2006.

\bibitem[Ioffe(2015)]{ioffe2015batch}
Sergey Ioffe.
\newblock Batch normalization: Accelerating deep network training by reducing internal covariate shift.
\newblock \emph{arXiv preprint arXiv:1502.03167}, 2015.

\bibitem[Isola et~al.(2017)Isola, Zhu, Zhou, and Efros]{isola2017image}
Phillip Isola, Jun-Yan Zhu, Tinghui Zhou, and Alexei~A Efros.
\newblock Image-to-image translation with conditional adversarial networks.
\newblock In \emph{Proceedings of the IEEE conference on computer vision and pattern recognition}, pages 1125--1134, 2017.

\bibitem[Kim et~al.(2017)Kim, Stratos, and Kim]{kim2017adversarial}
Young-Bum Kim, Karl Stratos, and Dongchan Kim.
\newblock Adversarial adaptation of synthetic or stale data.
\newblock In \emph{Proceedings of the 55th Annual Meeting of the Association for Computational Linguistics (Volume 1: Long Papers)}, pages 1297--1307, 2017.

\bibitem[Krizhevsky et~al.(2009)Krizhevsky, Hinton, et~al.]{krizhevsky2009learning}
Alex Krizhevsky, Geoffrey Hinton, et~al.
\newblock Learning multiple layers of features from tiny images.
\newblock 2009.

\bibitem[Kuka{\v{c}}ka et~al.(2017)Kuka{\v{c}}ka, Golkov, and Cremers]{kukavcka2017regularization}
Jan Kuka{\v{c}}ka, Vladimir Golkov, and Daniel Cremers.
\newblock Regularization for deep learning: A taxonomy.
\newblock \emph{arXiv preprint arXiv:1710.10686}, 2017.

\bibitem[LeCun(1989)]{lecun1989generalization}
Y LeCun.
\newblock Generalization and network design strategies.
\newblock \emph{Connections in Perspective}, 1989.

\bibitem[Lei~Ba et~al.(2016)Lei~Ba, Kiros, and Hinton]{lei2016layer}
Jimmy Lei~Ba, Jamie~Ryan Kiros, and Geoffrey~E Hinton.
\newblock Layer normalization.
\newblock \emph{ArXiv e-prints}, pages arXiv--1607, 2016.

\bibitem[Li et~al.(2020)Li, Socher, and Hoi]{li2020dividemix}
Junnan Li, Richard Socher, and Steven~CH Hoi.
\newblock Dividemix: Learning with noisy labels as semi-supervised learning.
\newblock \emph{arXiv preprint arXiv:2002.07394}, 2020.

\bibitem[Li et~al.(2017)Li, Wang, Li, Agustsson, and Van~Gool]{li2017webvision}
Wen Li, Limin Wang, Wei Li, Eirikur Agustsson, and Luc Van~Gool.
\newblock Webvision database: Visual learning and understanding from web data.
\newblock \emph{arXiv preprint arXiv:1708.02862}, 2017.

\bibitem[Liu et~al.(2022)Liu, Mao, Wu, Feichtenhofer, Darrell, and Xie]{liu2022convnet}
Zhuang Liu, Hanzi Mao, Chao-Yuan Wu, Christoph Feichtenhofer, Trevor Darrell, and Saining Xie.
\newblock A convnet for the 2020s.
\newblock In \emph{Proceedings of the IEEE/CVF conference on computer vision and pattern recognition}, pages 11976--11986, 2022.

\bibitem[Merity et~al.(2017)Merity, McCann, and Socher]{merity2017revisiting}
Stephen Merity, Bryan McCann, and Richard Socher.
\newblock Revisiting activation regularization for language rnns.
\newblock \emph{arXiv preprint arXiv:1708.01009}, 2017.

\bibitem[Mescheder et~al.(2018)Mescheder, Geiger, and Nowozin]{mescheder2018training}
Lars Mescheder, Andreas Geiger, and Sebastian Nowozin.
\newblock Which training methods for gans do actually converge?
\newblock In \emph{International conference on machine learning}, pages 3481--3490. PMLR, 2018.

\bibitem[Metz et~al.(2016)Metz, Poole, Pfau, and Sohl-Dickstein]{metz2016unrolled}
Luke Metz, Ben Poole, David Pfau, and Jascha Sohl-Dickstein.
\newblock Unrolled generative adversarial networks.
\newblock \emph{arXiv preprint arXiv:1611.02163}, 2016.

\bibitem[M{\"u}ller et~al.(2019)M{\"u}ller, Kornblith, and Hinton]{muller2019does}
Rafael M{\"u}ller, Simon Kornblith, and Geoffrey~E Hinton.
\newblock When does label smoothing help?
\newblock \emph{Advances in neural information processing systems}, 32, 2019.

\bibitem[Nilsback and Zisserman(2008)]{Nilsback08}
Maria-Elena Nilsback and Andrew Zisserman.
\newblock Automated flower classification over a large number of classes.
\newblock In \emph{Indian Conference on Computer Vision, Graphics and Image Processing}, 2008.

\bibitem[Shen et~al.(2018)Shen, Qu, Zhang, and Yu]{shen2018wasserstein}
Jian Shen, Yanru Qu, Weinan Zhang, and Yong Yu.
\newblock Wasserstein distance guided representation learning for domain adaptation.
\newblock In \emph{Proceedings of the AAAI conference on artificial intelligence}, 2018.

\bibitem[Singhal et~al.(2023)Singhal, Walambe, Ramanna, and Kotecha]{singhal2023domain}
Peeyush Singhal, Rahee Walambe, Sheela Ramanna, and Ketan Kotecha.
\newblock Domain adaptation: challenges, methods, datasets, and applications.
\newblock \emph{IEEE access}, 11:\penalty0 6973--7020, 2023.

\bibitem[Smith et~al.(2021)Smith, Dherin, Barrett, and De]{smith2021origin}
Samuel~L Smith, Benoit Dherin, David~GT Barrett, and Soham De.
\newblock On the origin of implicit regularization in stochastic gradient descent.
\newblock \emph{arXiv preprint arXiv:2101.12176}, 2021.

\bibitem[Song et~al.(2019)Song, Kim, Park, and Lee]{song2019does}
Hwanjun Song, Minseok Kim, Dongmin Park, and Jae-Gil Lee.
\newblock How does early stopping help generalization against label noise?
\newblock \emph{arXiv preprint arXiv:1911.08059}, 2019.

\bibitem[Srivastava et~al.(2014)Srivastava, Hinton, Krizhevsky, Sutskever, and Salakhutdinov]{JMLR:v15:srivastava14a}
Nitish Srivastava, Geoffrey Hinton, Alex Krizhevsky, Ilya Sutskever, and Ruslan Salakhutdinov.
\newblock Dropout: A simple way to prevent neural networks from overfitting.
\newblock \emph{Journal of Machine Learning Research}, 15\penalty0 (56):\penalty0 1929--1958, 2014.

\bibitem[Szegedy et~al.(2016)Szegedy, Vanhoucke, Ioffe, Shlens, and Wojna]{szegedy2016rethinking}
Christian Szegedy, Vincent Vanhoucke, Sergey Ioffe, Jon Shlens, and Zbigniew Wojna.
\newblock Rethinking the inception architecture for computer vision.
\newblock In \emph{Proceedings of the IEEE conference on computer vision and pattern recognition}, pages 2818--2826, 2016.

\bibitem[Tang et~al.(2024)Tang, Lv, Zhang, Zhou, Duan, Wu, and Kuang]{tang2024aug}
Zihao Tang, Zheqi Lv, Shengyu Zhang, Yifan Zhou, Xinyu Duan, Fei Wu, and Kun Kuang.
\newblock Aug-kd: Anchor-based mixup generation for out-of-domain knowledge distillation.
\newblock \emph{arXiv preprint arXiv:2403.07030}, 2024.

\bibitem[Tian and Zhang(2022)]{tian2022comprehensive}
Yingjie Tian and Yuqi Zhang.
\newblock A comprehensive survey on regularization strategies in machine learning.
\newblock \emph{Information Fusion}, 80:\penalty0 146--166, 2022.

\bibitem[Tompson et~al.(2015)Tompson, Goroshin, Jain, LeCun, and Bregler]{tompson2015efficient}
Jonathan Tompson, Ross Goroshin, Arjun Jain, Yann LeCun, and Christoph Bregler.
\newblock Efficient object localization using convolutional networks.
\newblock In \emph{Proceedings of the IEEE conference on computer vision and pattern recognition}, pages 648--656, 2015.

\bibitem[Tzeng et~al.(2014)Tzeng, Hoffman, Zhang, Saenko, and Darrell]{tzeng2014deep}
Eric Tzeng, Judy Hoffman, Ning Zhang, Kate Saenko, and Trevor Darrell.
\newblock Deep domain confusion: Maximizing for domain invariance.
\newblock \emph{arXiv preprint arXiv:1412.3474}, 2014.

\bibitem[Tzeng et~al.(2017)Tzeng, Hoffman, Saenko, and Darrell]{tzeng2017adversarial}
Eric Tzeng, Judy Hoffman, Kate Saenko, and Trevor Darrell.
\newblock Adversarial discriminative domain adaptation.
\newblock In \emph{Proceedings of the IEEE conference on computer vision and pattern recognition}, pages 7167--7176, 2017.

\bibitem[Vapnik and Chervonenkis(2015)]{vapnik2015uniform}
Vladimir~N Vapnik and A~Ya Chervonenkis.
\newblock On the uniform convergence of relative frequencies of events to their probabilities.
\newblock In \emph{Measures of complexity: festschrift for alexey chervonenkis}, pages 11--30. Springer, 2015.

\bibitem[Welling and Teh(2011)]{welling2011bayesian}
Max Welling and Yee~W Teh.
\newblock Bayesian learning via stochastic gradient langevin dynamics.
\newblock In \emph{Proceedings of the 28th international conference on machine learning (ICML-11)}, pages 681--688. Citeseer, 2011.

\bibitem[Wu and He(2018)]{wu2018group}
Yuxin Wu and Kaiming He.
\newblock Group normalization.
\newblock In \emph{Proceedings of the European conference on computer vision (ECCV)}, pages 3--19, 2018.

\bibitem[Xu et~al.(2019)Xu, Mohtarami, and Glass]{xu2019adversarial}
Brian Xu, Mitra Mohtarami, and James Glass.
\newblock Adversarial domain adaptation for stance detection.
\newblock \emph{arXiv preprint arXiv:1902.02401}, 2019.

\bibitem[Zhang et~al.(2021)Zhang, Bengio, Hardt, Recht, and Vinyals]{zhang2021understanding}
Chiyuan Zhang, Samy Bengio, Moritz Hardt, Benjamin Recht, and Oriol Vinyals.
\newblock Understanding deep learning (still) requires rethinking generalization.
\newblock \emph{Communications of the ACM}, 64\penalty0 (3):\penalty0 107--115, 2021.

\bibitem[Zhong et~al.(2020)Zhong, Zheng, Kang, Li, and Yang]{zhong2020random}
Zhun Zhong, Liang Zheng, Guoliang Kang, Shaozi Li, and Yi Yang.
\newblock Random erasing data augmentation.
\newblock In \emph{Proceedings of the AAAI conference on artificial intelligence}, pages 13001--13008, 2020.

\end{thebibliography}
}

\ifarxiv \clearpage \appendix 
\label{sec:appendix_section}
\section{Appendix Section}

\begin{table}[ht]
\renewcommand{\arraystretch}{1.2} 

\resizebox{\textwidth}{!}{  
\begin{tabular}{l c c c c c c}
\hline
\textbf{Method} & \textbf{Min Val Loss} & \textbf{Max ACC1} & \textbf{Max ACC5} & \textbf{Avg Last 10 Val Loss} & \textbf{Avg Last 10 ACC1} & \textbf{Avg Last 10 ACC5} \\
\hline
Baseline   & 1.81 $\pm$ 0.04  & 56.6 $\pm$ 0.12  & 82.9 $\pm$ 0.31  & 3.6 $\pm$ 0.14 & 55.8 $\pm$ 0.37 & 82.3 $\pm$ 0.35 \\
CF 0.2   & \textbf{1.69 $\pm$ 0.017} & 54.4 $\pm$ 0.47  & 83.1 $\pm$ 0.53  & 2.85 $\pm$ 0.26 & 50.6 $\pm$ 0.46 & 77.5 $\pm$ 0.1 \\
Dropout 0.5     & 1.76 $\pm$ 0.08  & 56.3 $\pm$ 0.22  & 82.9 $\pm$ 0.56  & 3.65 $\pm$ 0.06 & 55.4 $\pm$ 0.34 & 82.0 $\pm$ 0.37 \\
Weightdecay 1e-4   & 1.71 $\pm$ 0.019 & 55.6 $\pm$ 0.26  & 84.0 $\pm$ 0.21  & 2.54 $\pm$ 0.06 & 53.4 $\pm$ 0.63 & 80.9 $\pm$ 0.18 \\
LabelSmooth 0.1    & 2.21 $\pm$ 0.01  & \textbf{62.6 $\pm$ 0.18}  & \textbf{85.0 $\pm$ 0.09}  & \textbf{2.24 $\pm$ 0.01} & \textbf{62.12 $\pm$ 0.29} & \textbf{84.5 $\pm$ 0.21} \\
\hline
CF + DP  & 1.74 $\pm$ 0.03  & 53.7 $\pm$ 0.81  & 82.4 $\pm$ 0.51  & 3.17 $\pm$ 0.11 & 51.0 $\pm$ 0.35 & 78.3 $\pm$ 0.27 \\
CF + LSM & 2.17 $\pm$ 0.02  & \textbf{59.3 $\pm$ 0.20}  & \textbf{84.9 $\pm$ 0.30}  & \textbf{2.29 $\pm$ 0.01} &\textbf{ 58.4 $\pm$ 0.59}& \textbf{82.3 $\pm$ 0.60} \\
CF + WD  &\textbf{ 1.67 $\pm$ 0.02}  & 55.1 $\pm$ 0.28  & 83.8 $\pm$ 0.09  & 2.42 $\pm$ 0.05 & 50.8 $\pm$ 0.45 & 78.8 $\pm$ 0.20 \\
\hline
\end{tabular}
}

\caption{Strategies with and without CF on CIFAR-100}
\label{tab:cifar100_performance}
\end{table}

\begin{table}[ht]
\renewcommand{\arraystretch}{1.2} 

\resizebox{\textwidth}{!}{  
\begin{tabular}{l c c c c c c}
\hline
\textbf{Method} & \textbf{Min Val Loss} & \textbf{Max ACC1} & \textbf{Max ACC5} & \textbf{Avg Last 10 Val Loss} & \textbf{Avg Last 10 ACC1} & \textbf{Avg Last 10 ACC5} \\
\hline
Baseline       & $2.43 \pm 0.03$ & $36.0 \pm 0.4$  & $68.5 \pm 0.21$ & $5.92 \pm 0.08$  & $34.2 \pm 0.27$  & $63.2 \pm 1.12$ \\
CF 0.2       & $2.35 \pm 0.05$ & $38.0 \pm 0.4$  & $68.8 \pm 0.52$ & $3.96 \pm 0.09$  & $31.68 \pm 1.3$ & $60.1 \pm 2.3$  \\
Dropout 0.5          & $2.35 \pm 0.01$ & $37.4 \pm 1.0$  & $69.7 \pm 0.42$ & $5.92 \pm 0.44$  & $33.7 \pm 0.82$ & $62.8 \pm 0.40$ \\
Weightdecay 1e-4        & $2.37 \pm 0.01$ & $37.3 \pm 0.80$ & $69.7 \pm 0.5$  & $4.91 \pm 0.17$  & $31.2 \pm 0.71$ & $60.9 \pm 0.32$ \\
LabelSmooth 0.1        & $2.72 \pm 0.01$ & $39.5 \pm 0.40$ & $67.4 \pm 0.65$ & $2.77 \pm 0.02$  & $37.9 \pm 0.80$ & $64.3 \pm 0.86$ \\
\hline
CF + DP        & $2.32 \pm 0.04$ & $38.8 \pm 0.37$ & $70.5 \pm 0.67$ & $4.26 \pm 0.18$  & $32.1 \pm 0.51$ & $62.5 \pm 0.26$ \\
CF + LSM       & $2.68 \pm 0.02$ & $39.4 \pm 0.99$ & $69.0 \pm 0.68$ & $2.79 \pm 0.02$  & $37.1 \pm 1.25$ & $64.2 \pm 0.47$ \\
CF + WD        & $2.51 \pm 0.04$ & $33.7 \pm 0.78$ & $66.3 \pm 0.77$ & $4.38 \pm 0.02$  & $30.0 \pm 0.86$ & $59.7 \pm 0.32$ \\
\hline
\end{tabular}
}

\caption{Strategies with and without CF on Webvision-mini}
\label{tab:webvision_performance}
\end{table}

\begin{figure*}[htbp]
    \centering
    \subsection*{CIFAR-100 Results}
    \begin{subfigure}[b]{0.3\textwidth}
        \centering
        \includegraphics[width=\textwidth]{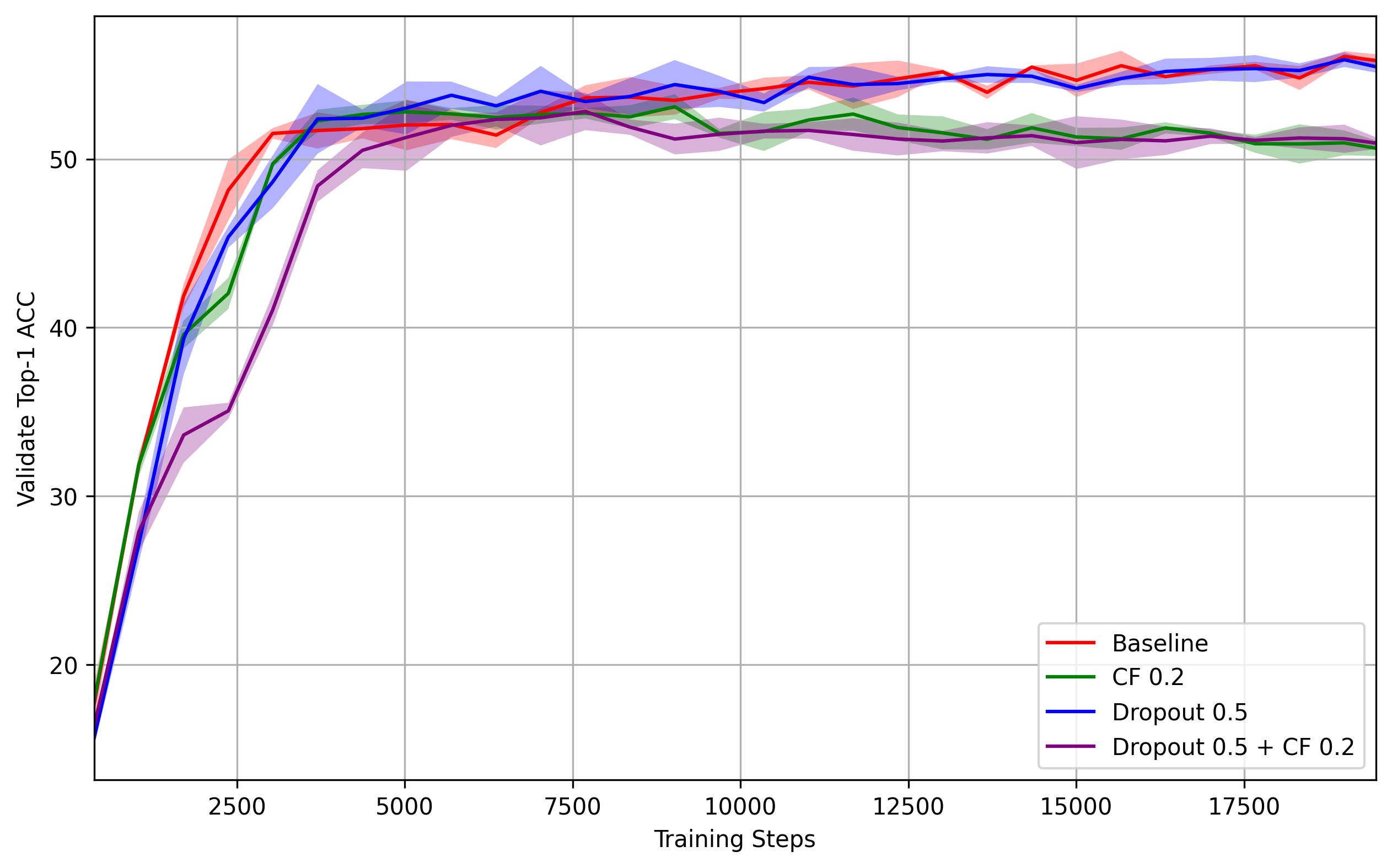}
    \end{subfigure} \hfill
    \begin{subfigure}[b]{0.3\textwidth}
        \centering
        \includegraphics[width=\textwidth]{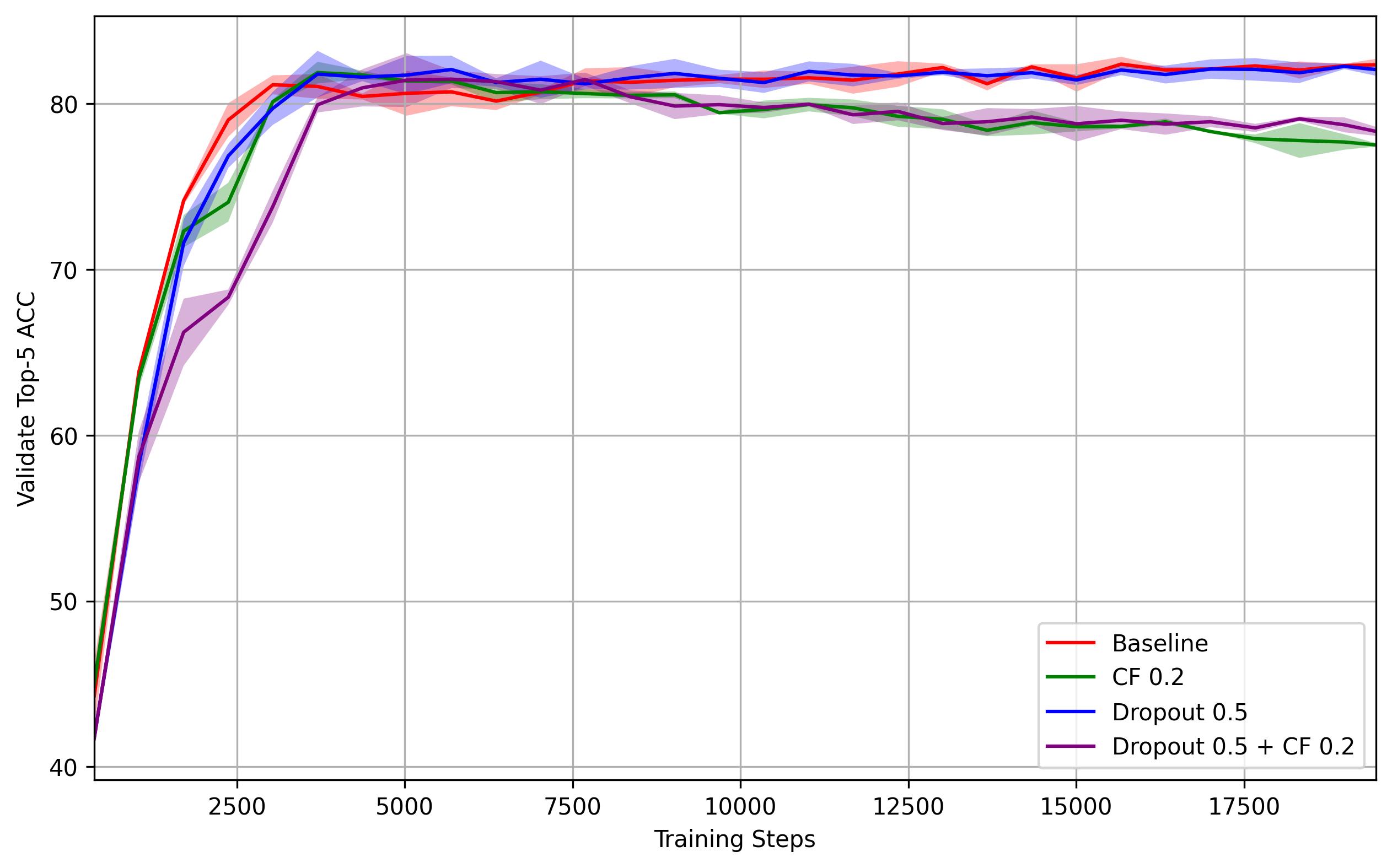}
    \end{subfigure} \hfill
    \begin{subfigure}[b]{0.3\textwidth}
        \centering
        \includegraphics[width=\textwidth]{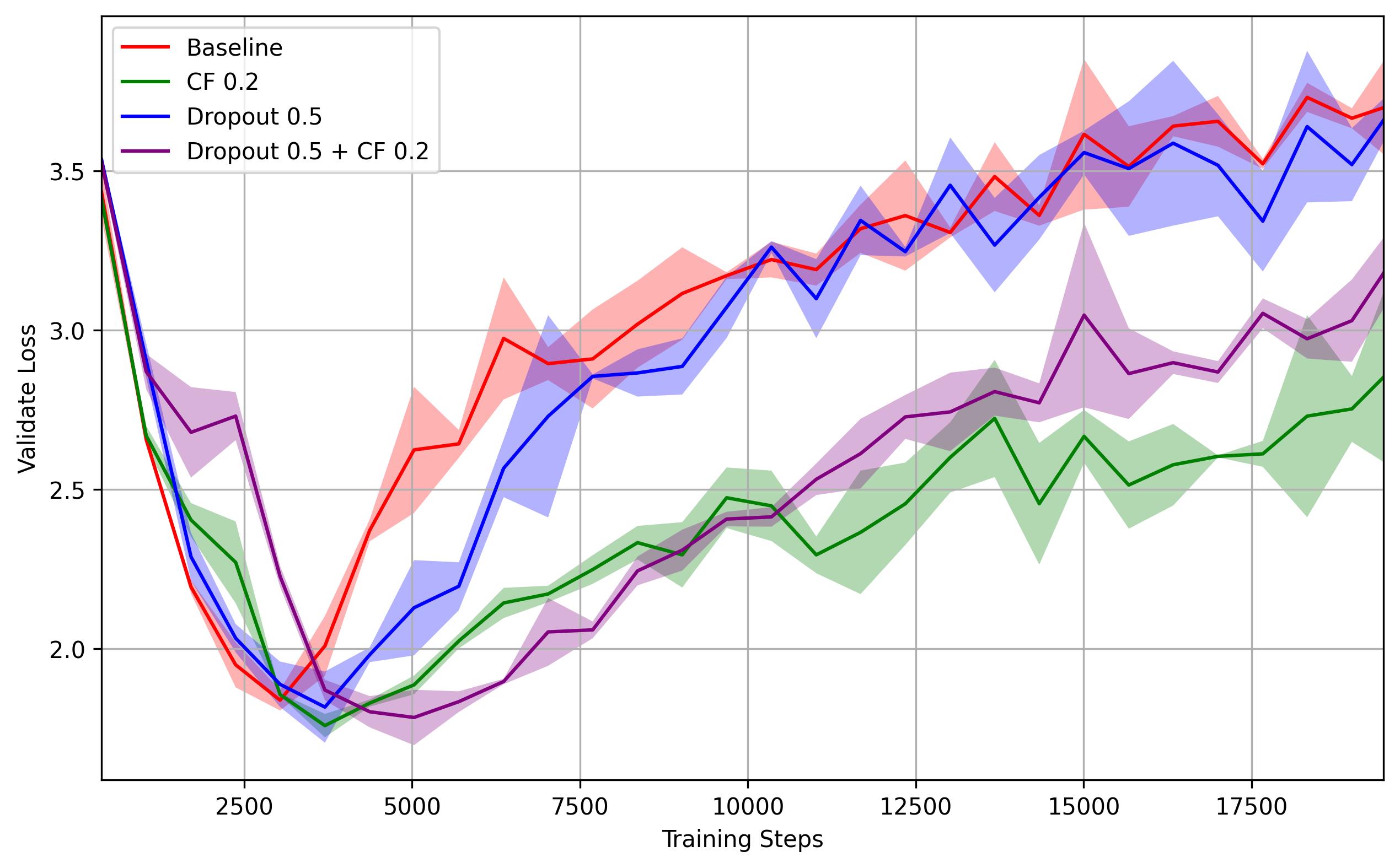}
    \end{subfigure} \hfill

    \begin{subfigure}[b]{0.3\textwidth}
        \centering
        \includegraphics[width=\textwidth]{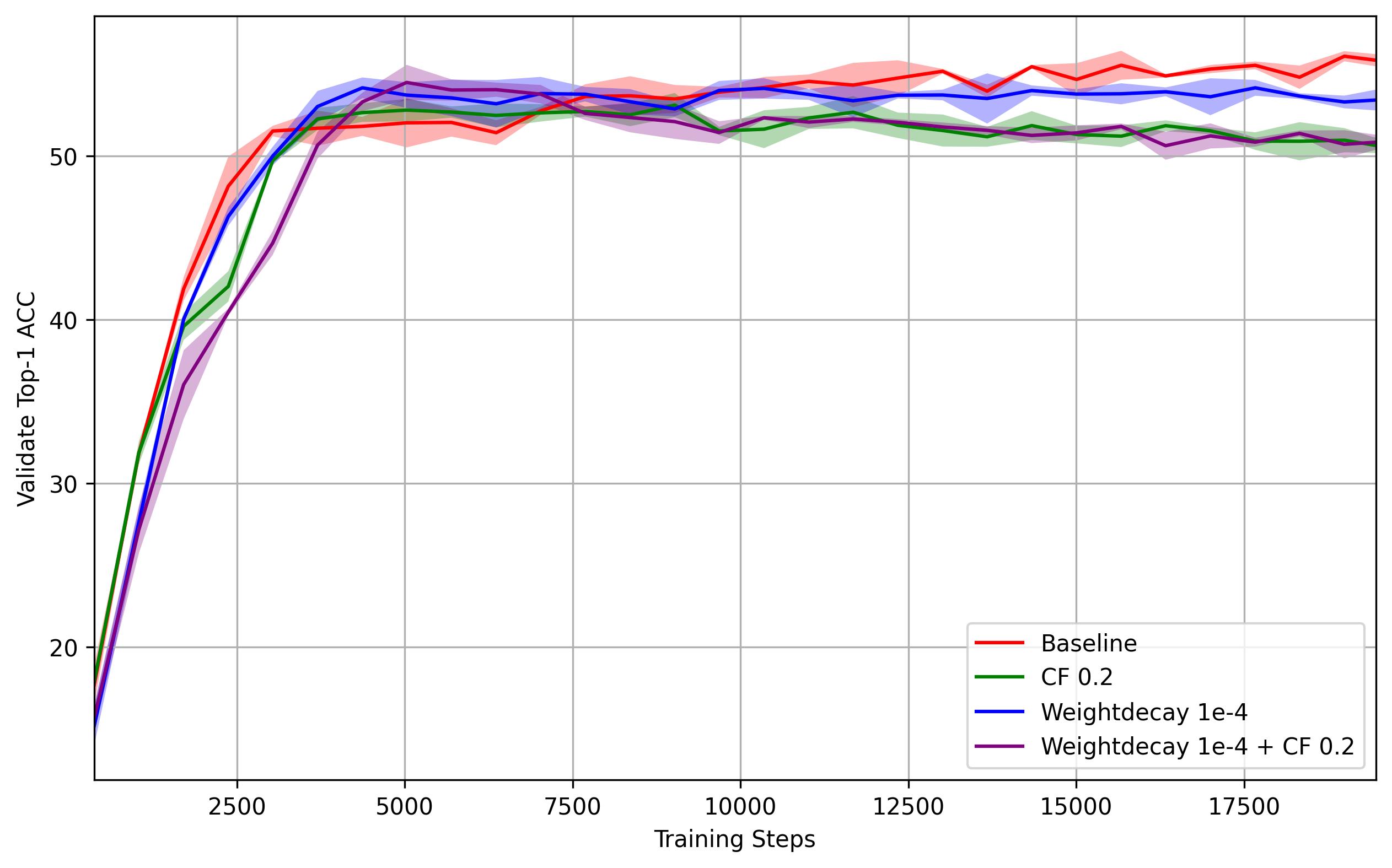}
    \end{subfigure} \hfill
    \begin{subfigure}[b]{0.3\textwidth}
        \centering
        \includegraphics[width=\textwidth]{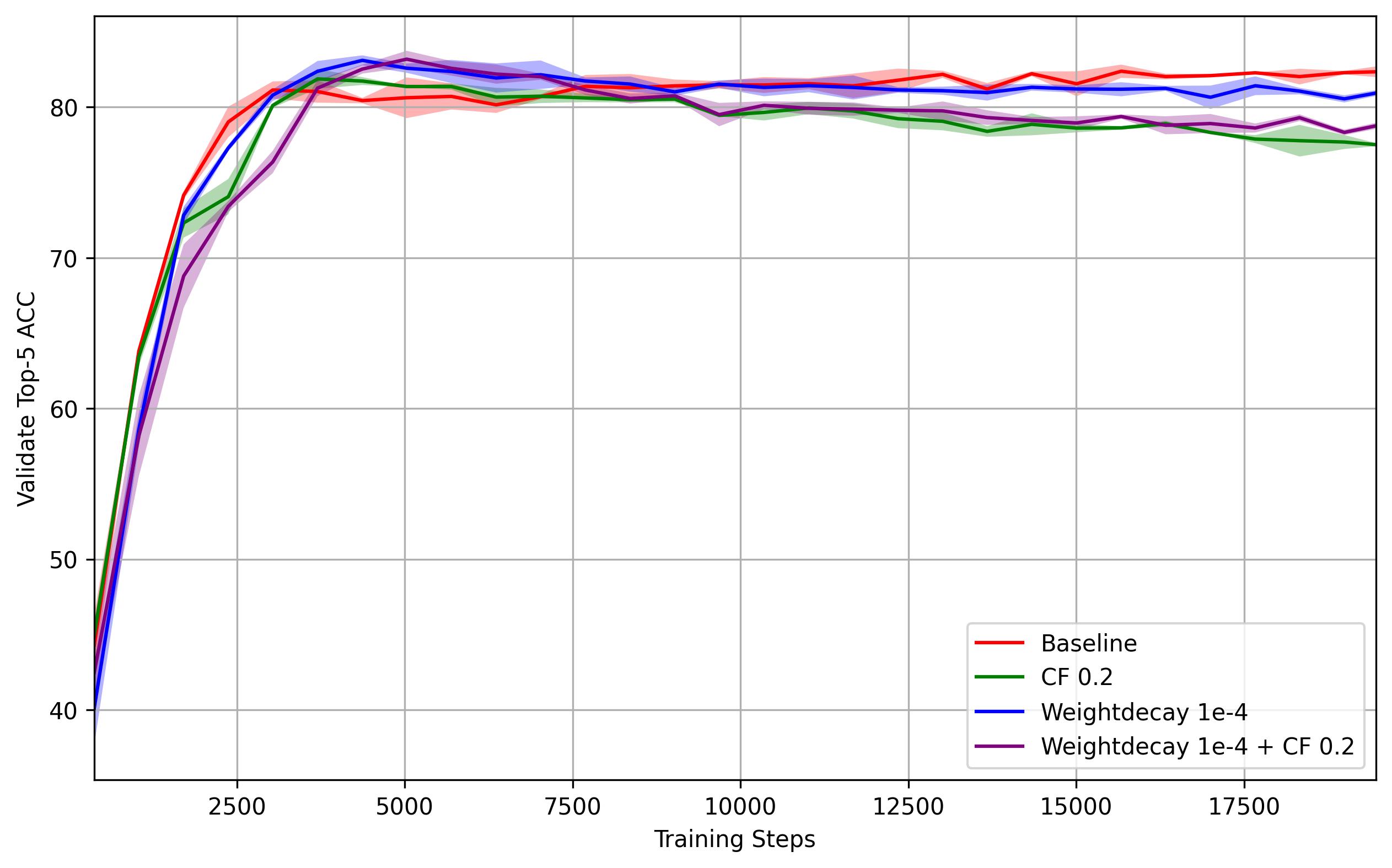}
    \end{subfigure} \hfill
    \begin{subfigure}[b]{0.3\textwidth}
        \centering
        \includegraphics[width=\textwidth]{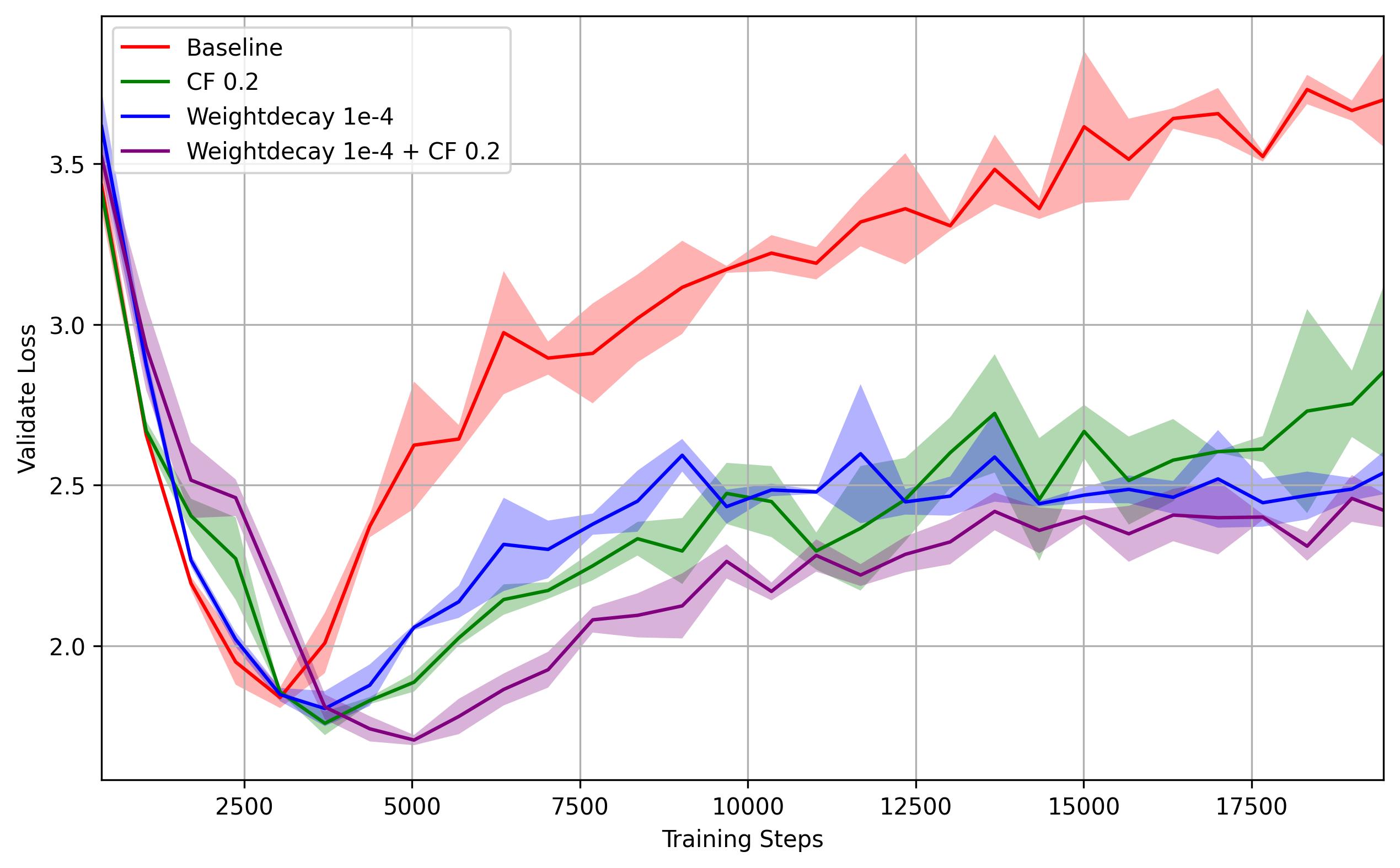}
    \end{subfigure} \hfill

    \begin{subfigure}[b]{0.3\textwidth}
        \centering
        \includegraphics[width=\textwidth]{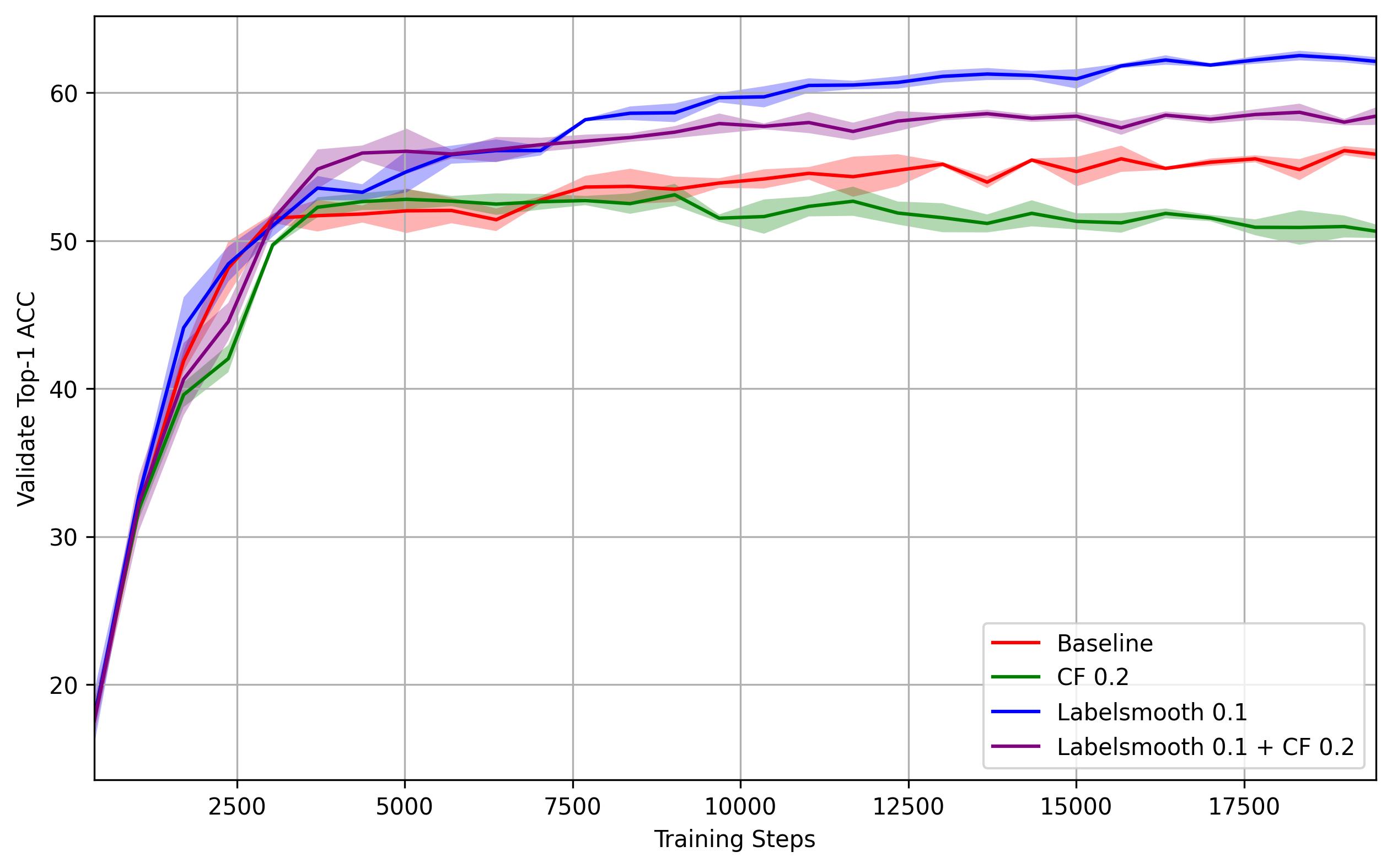}
    \end{subfigure} \hfill
    \begin{subfigure}[b]{0.3\textwidth}
        \centering
        \includegraphics[width=\textwidth]{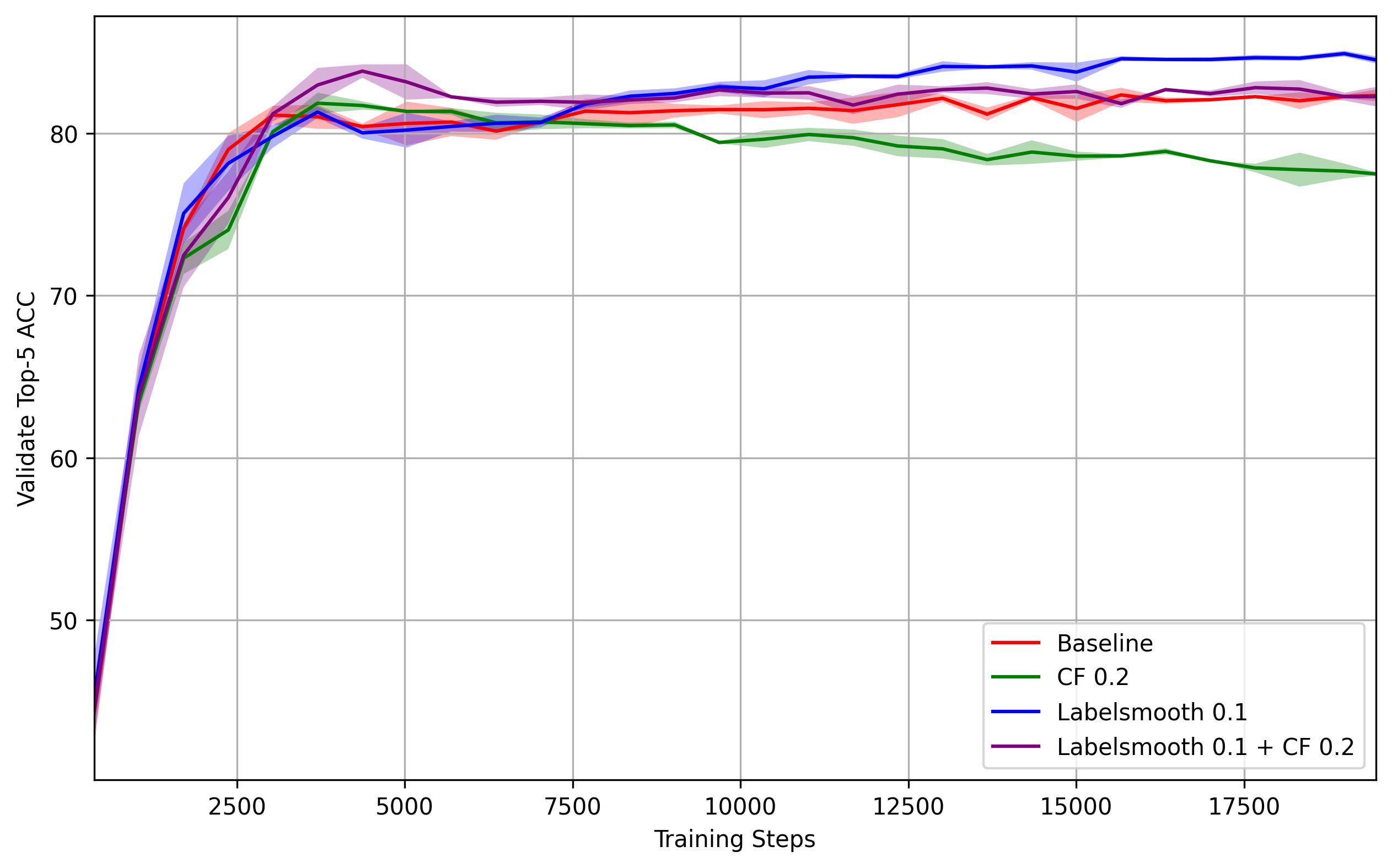}
    \end{subfigure} \hfill
    \begin{subfigure}[b]{0.3\textwidth}
        \centering
        \includegraphics[width=\textwidth]{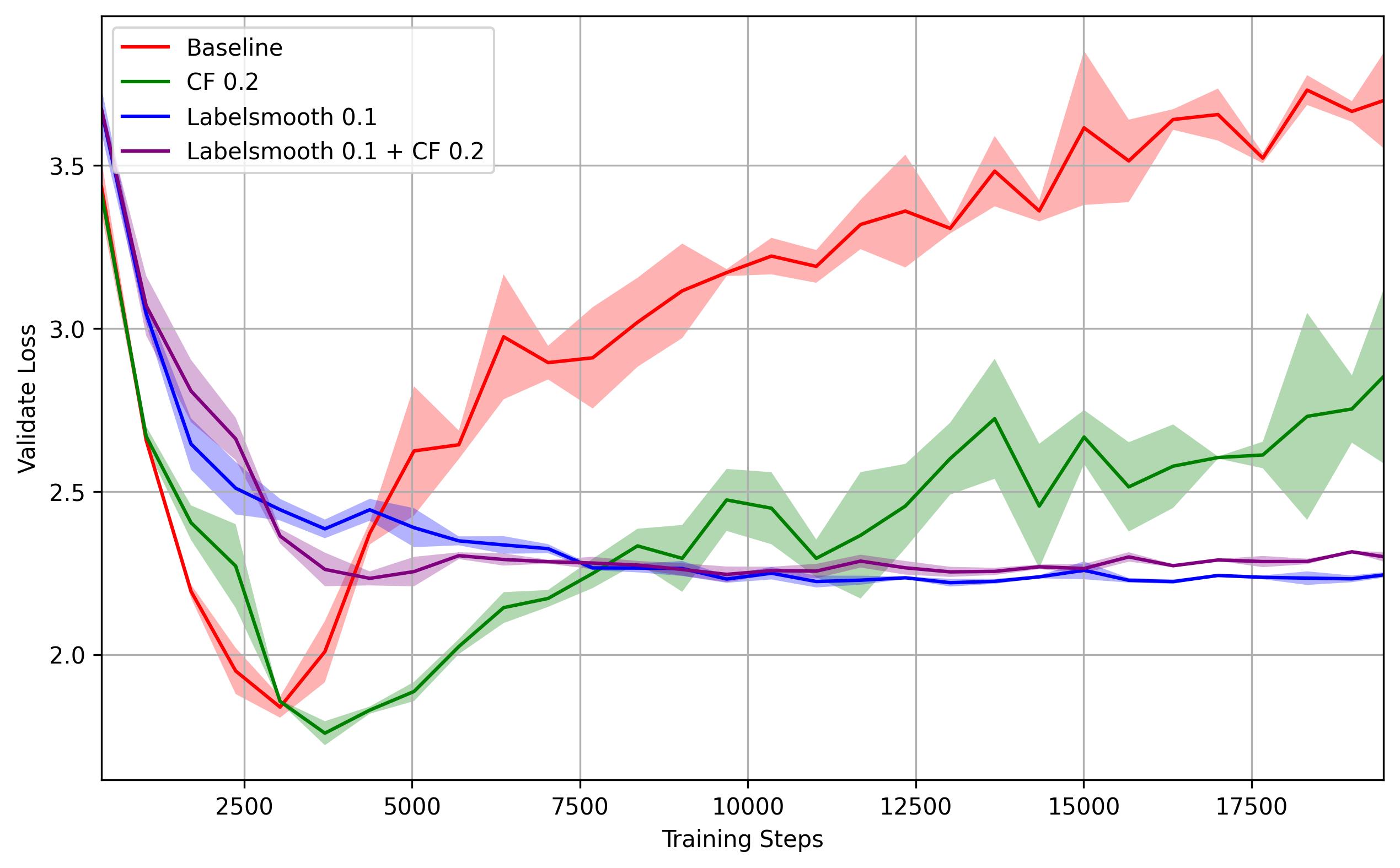}
    \end{subfigure} \hfill

    \subsection*{WebVision Results}
    \begin{subfigure}[b]{0.3\textwidth}
        \centering
        \includegraphics[width=\textwidth]{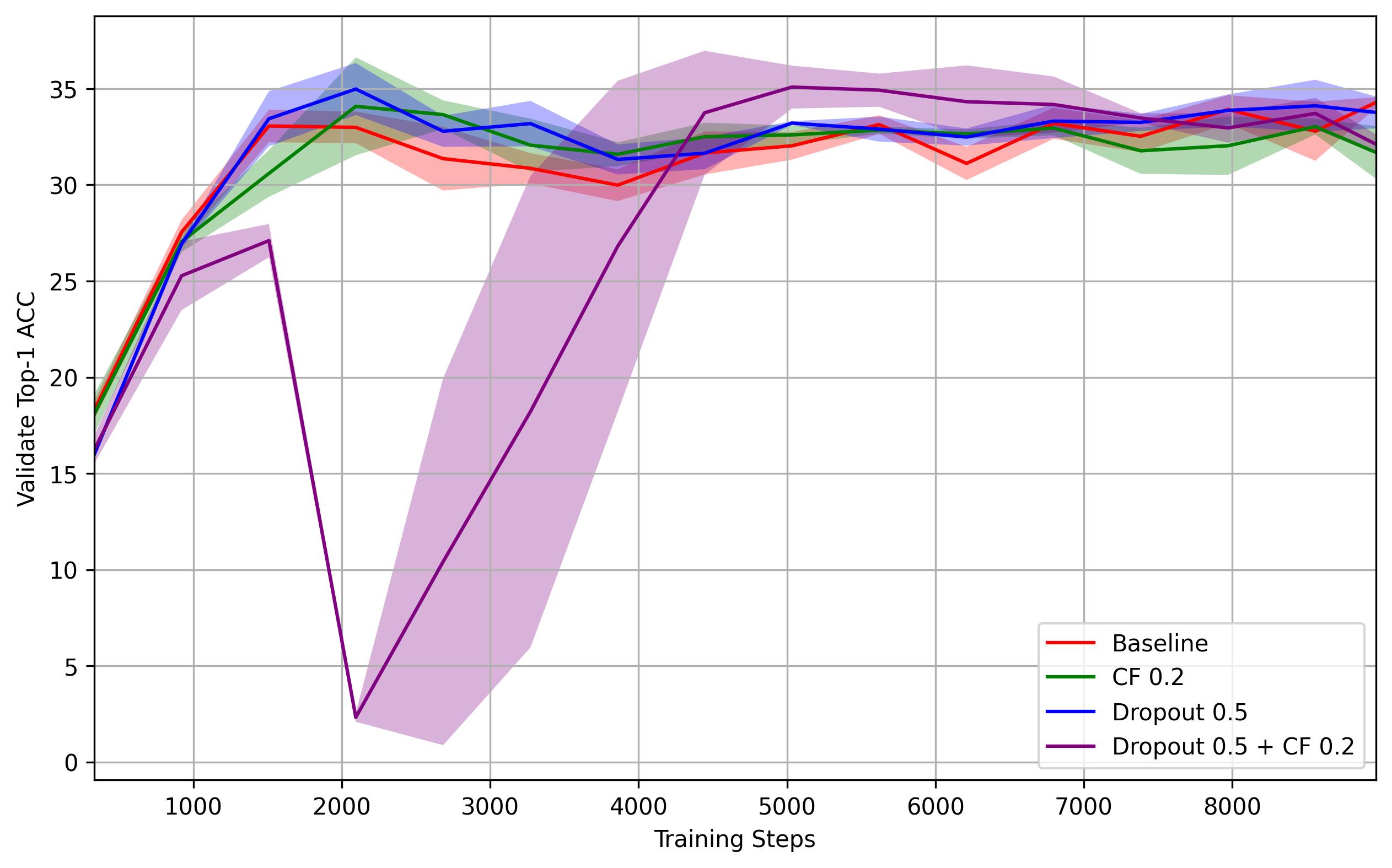}
    \end{subfigure} \hfill
    \begin{subfigure}[b]{0.3\textwidth}
        \centering
        \includegraphics[width=\textwidth]{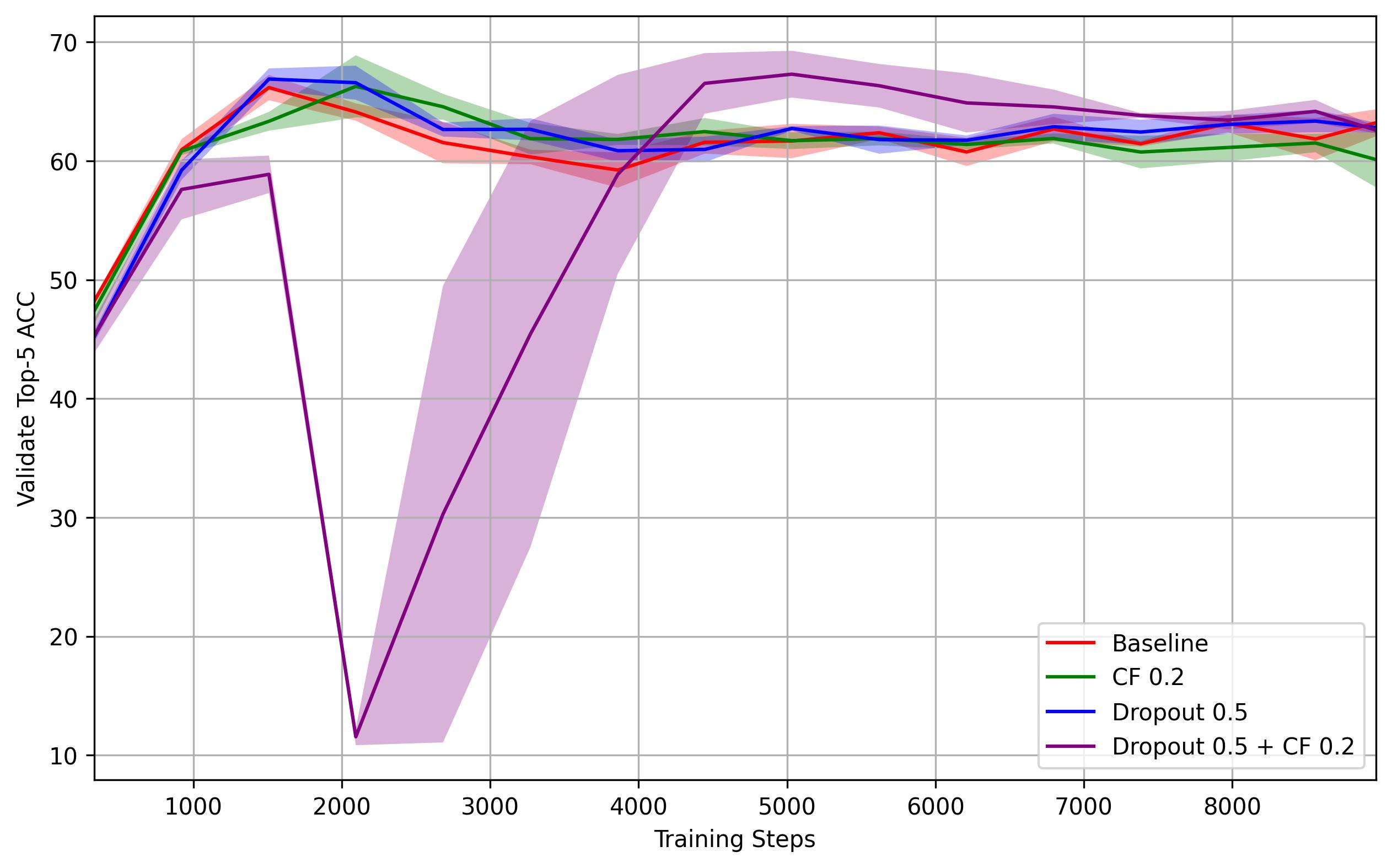}
    \end{subfigure} \hfill
    \begin{subfigure}[b]{0.3\textwidth}
        \centering
        \includegraphics[width=\textwidth]{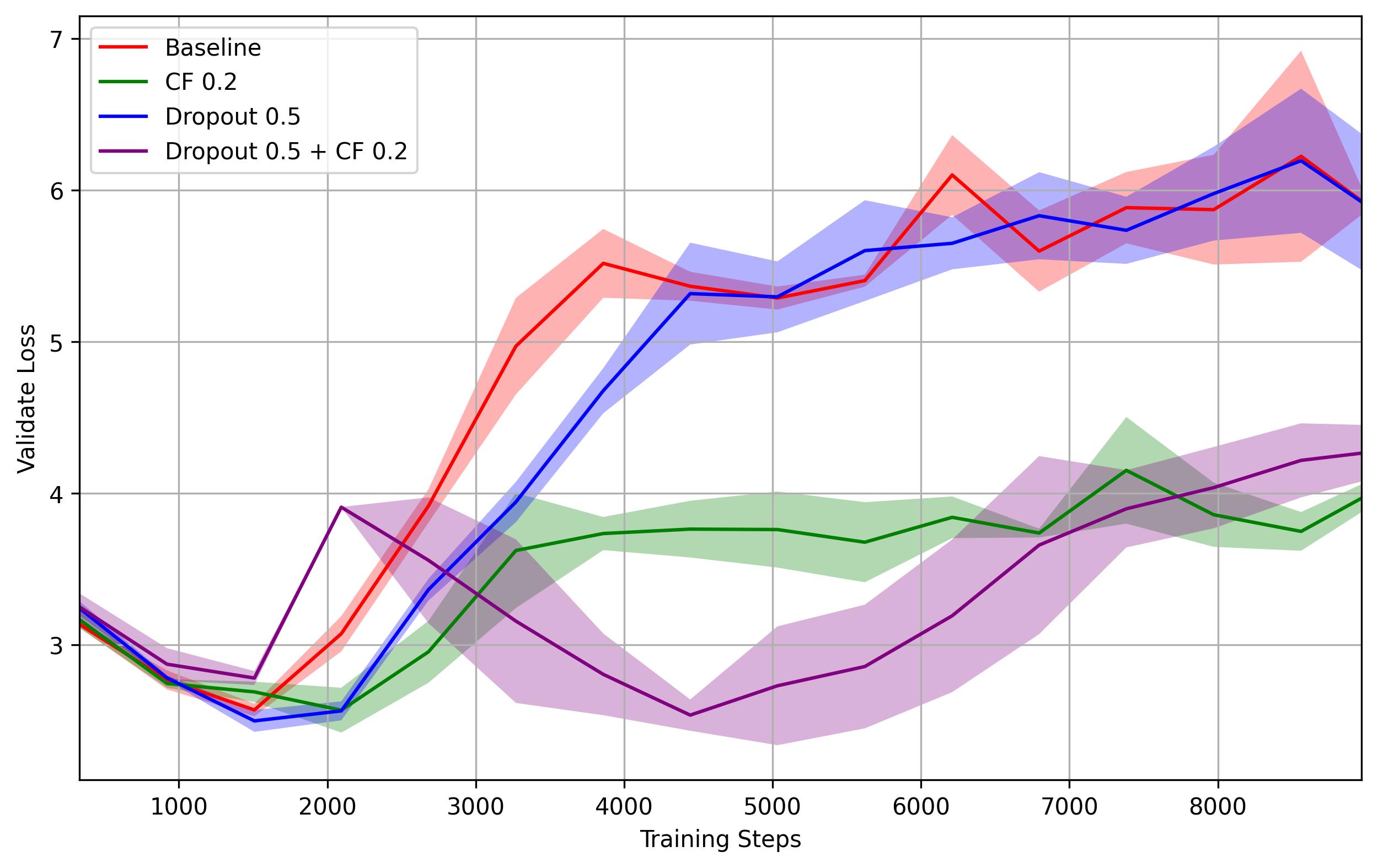}
    \end{subfigure} \hfill

    \begin{subfigure}[b]{0.3\textwidth}
        \centering
        \includegraphics[width=\textwidth]{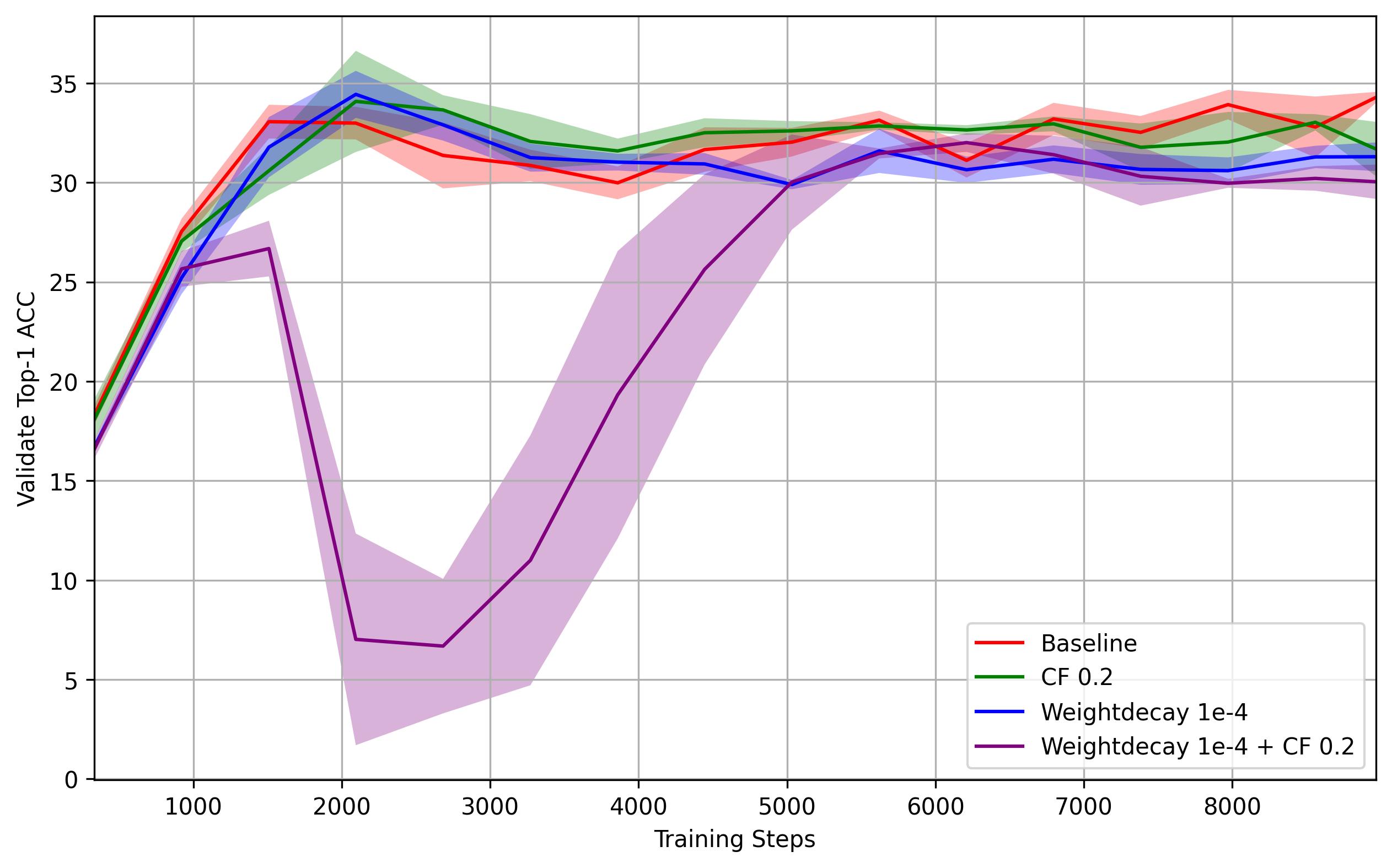}
    \end{subfigure} \hfill
    \begin{subfigure}[b]{0.3\textwidth}
        \centering
        \includegraphics[width=\textwidth]{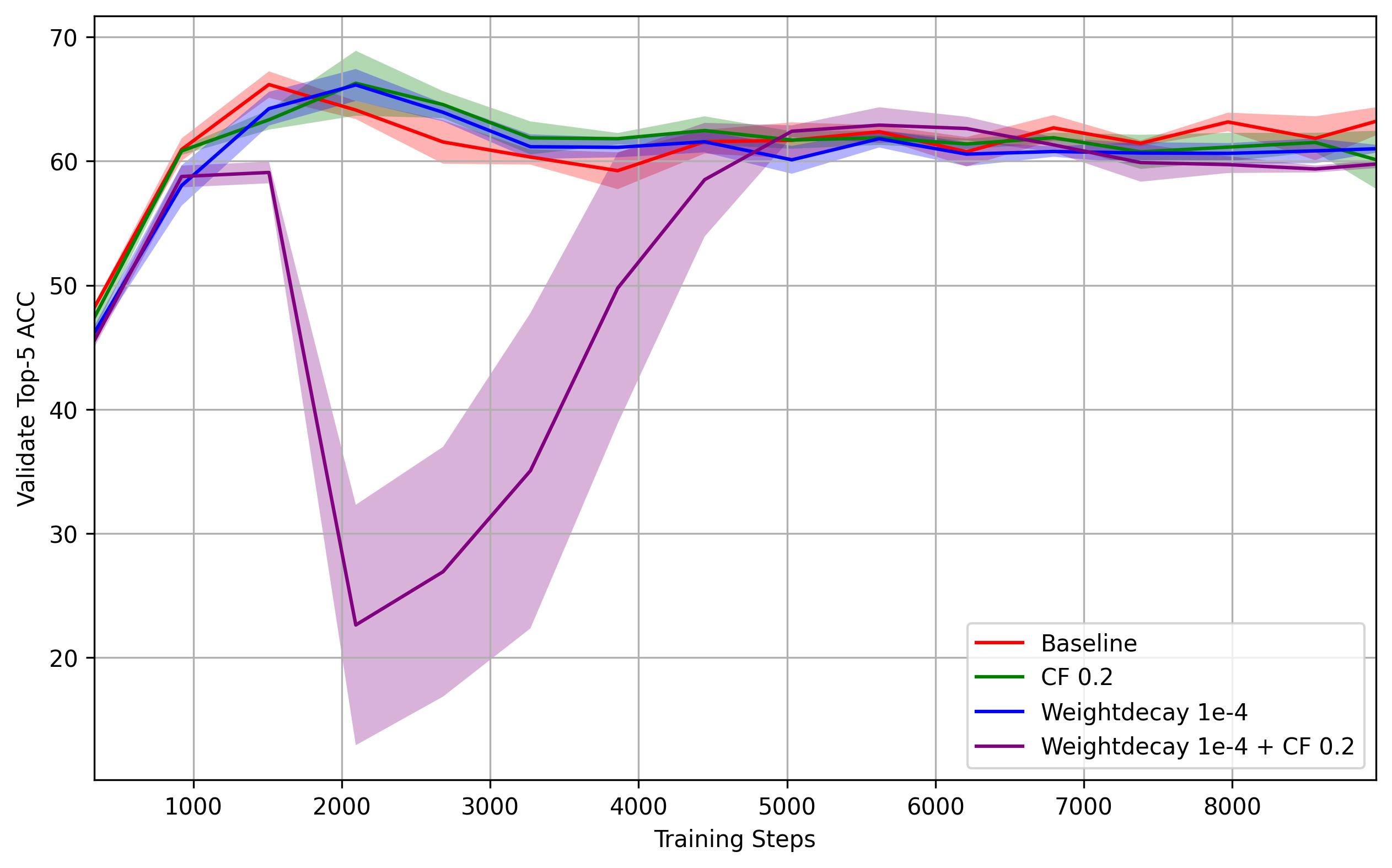}
    \end{subfigure} \hfill
    \begin{subfigure}[b]{0.3\textwidth}
        \centering
        \includegraphics[width=\textwidth]{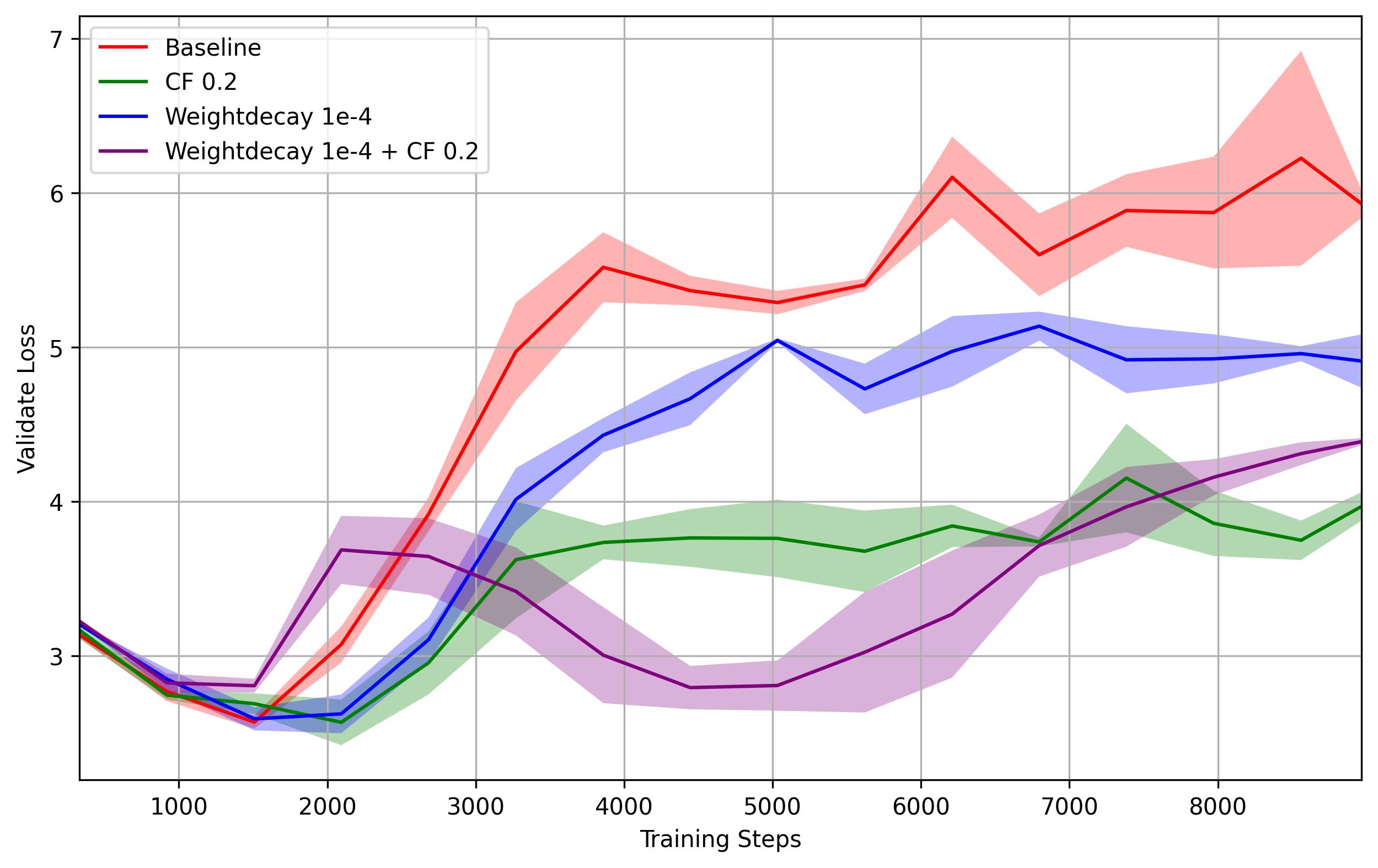}
    \end{subfigure} \hfill

    \begin{subfigure}[b]{0.3\textwidth}
        \centering
        \includegraphics[width=\textwidth]{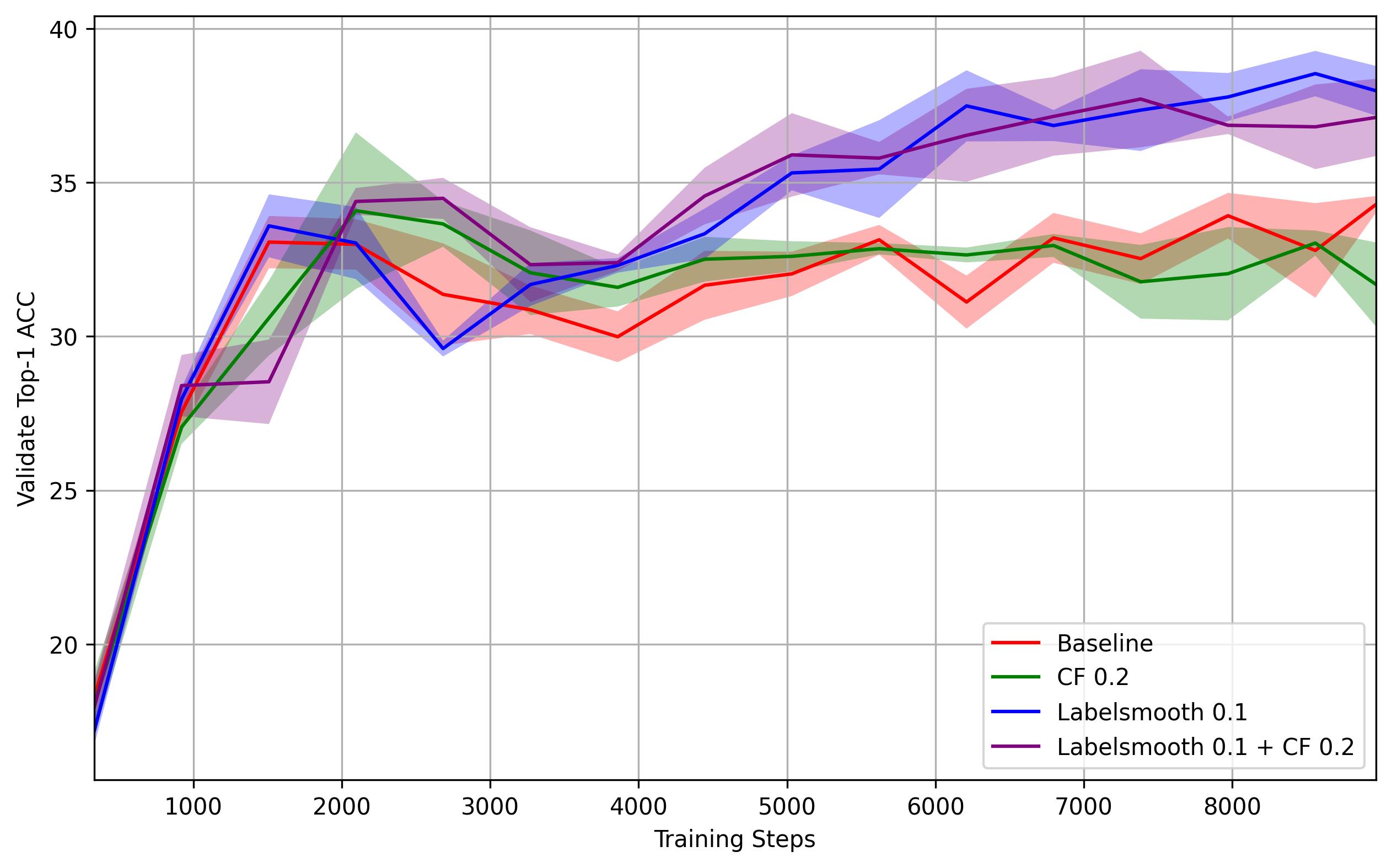}
    \end{subfigure} \hfill
    \begin{subfigure}[b]{0.3\textwidth}
        \centering
        \includegraphics[width=\textwidth]{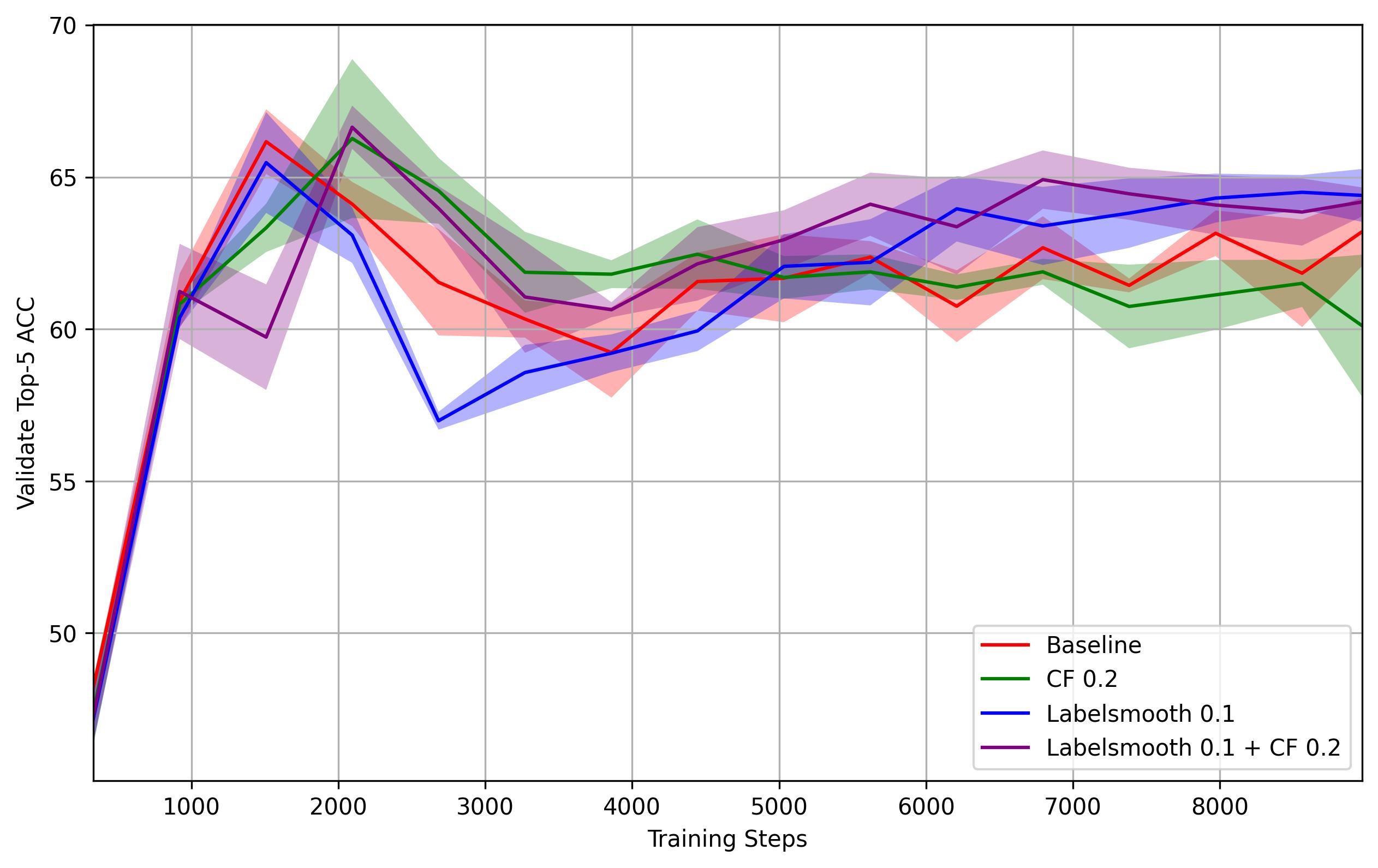}
    \end{subfigure} \hfill

    \caption{Training Curves with Combined Methods on CIFAR-100 and WebVision}
    \label{fig:compatibility}
\end{figure*}

\begin{figure*}[htbp]
    \centering
    
    \begin{subfigure}[b]{0.45\textwidth}
        \centering
        \includegraphics[width=\textwidth]{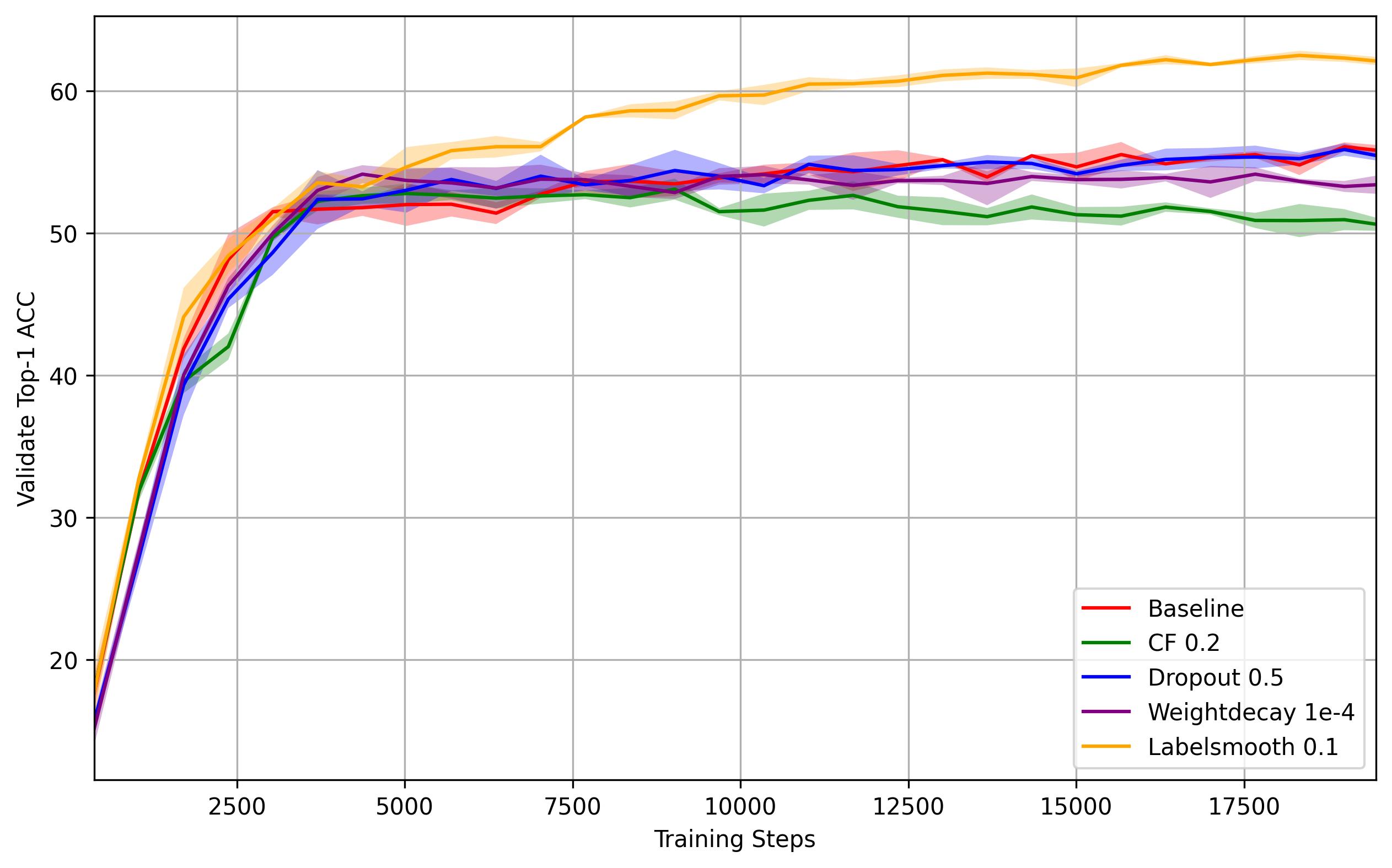} 
        \caption{CIFAR-100 Top-1 ACC} 
        \label{fig:memo_noise}
    \end{subfigure} \hfill
    \begin{subfigure}[b]{0.45\textwidth}
        \centering
        \includegraphics[width=\textwidth]{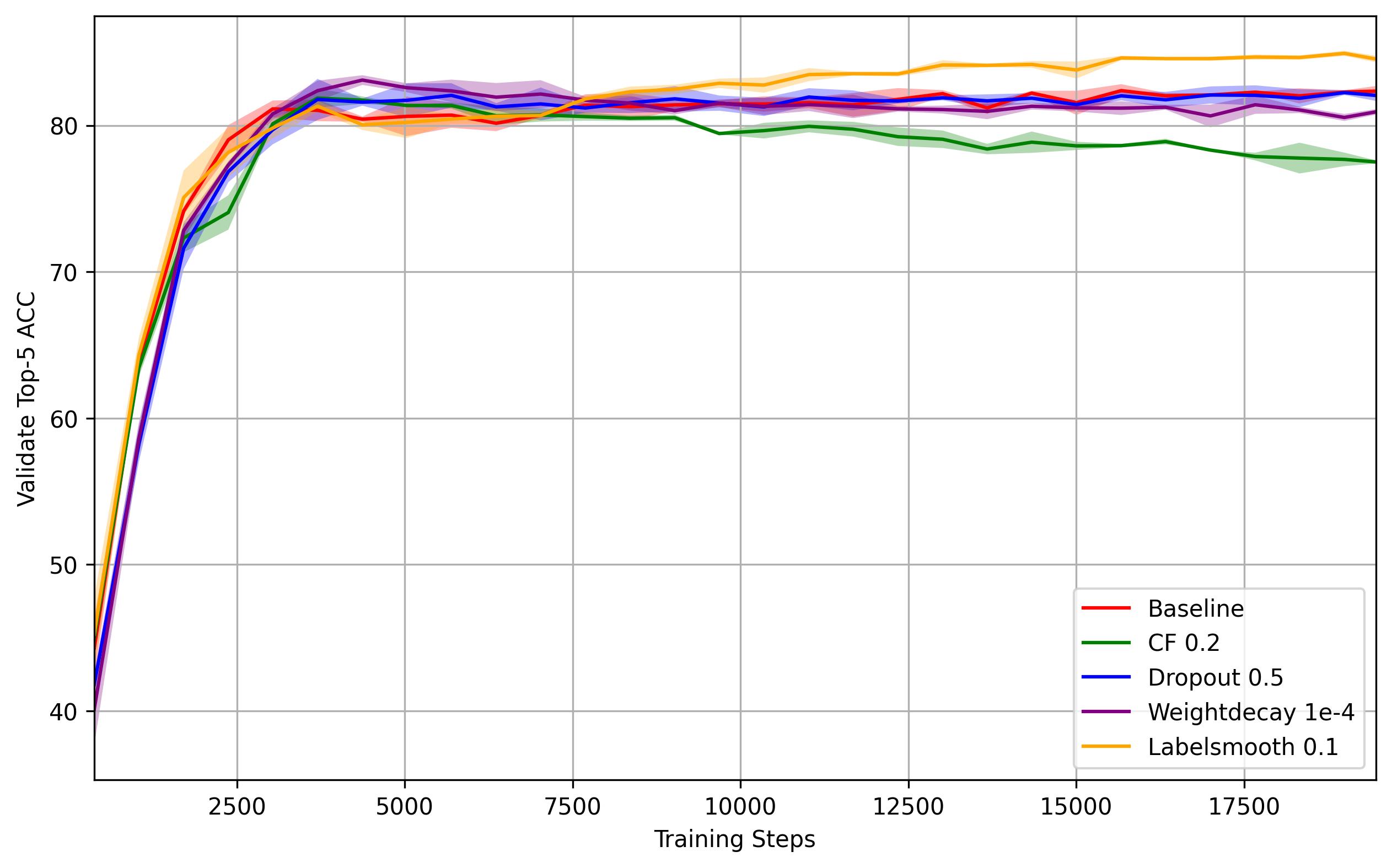} 
        \caption{CIFAR-100 Top-5 ACC} 
        \label{fig:memo_normal}
    \end{subfigure} \hfill
    \begin{subfigure}[b]{0.45\textwidth}
        \centering
        \includegraphics[width=\textwidth]{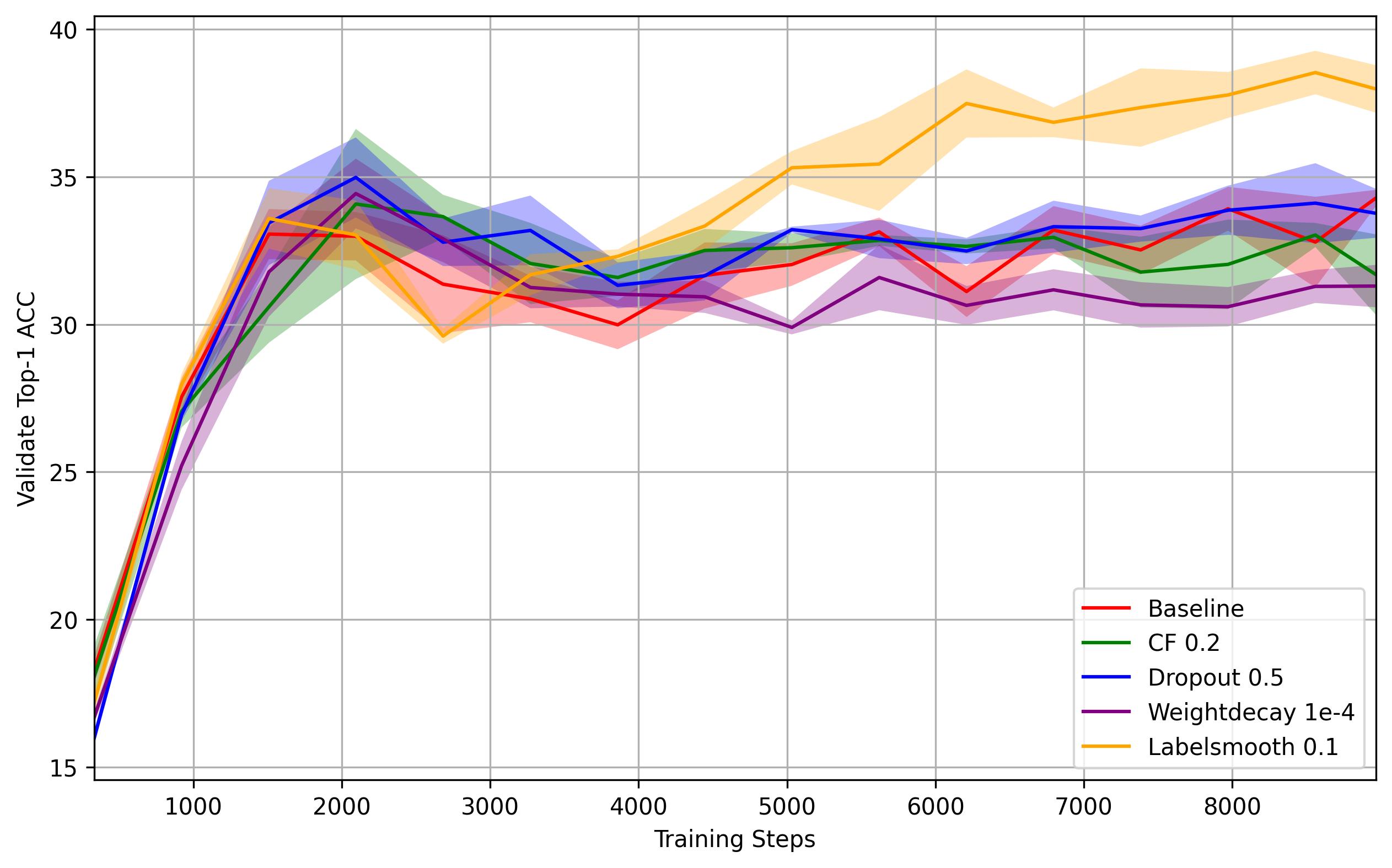} 
        \caption{Webvision-mini Top-1 ACC} 
        \label{fig:warmup}
    \end{subfigure} \hfill
    \begin{subfigure}[b]{0.45\textwidth}
        \centering
        \includegraphics[width=\textwidth]{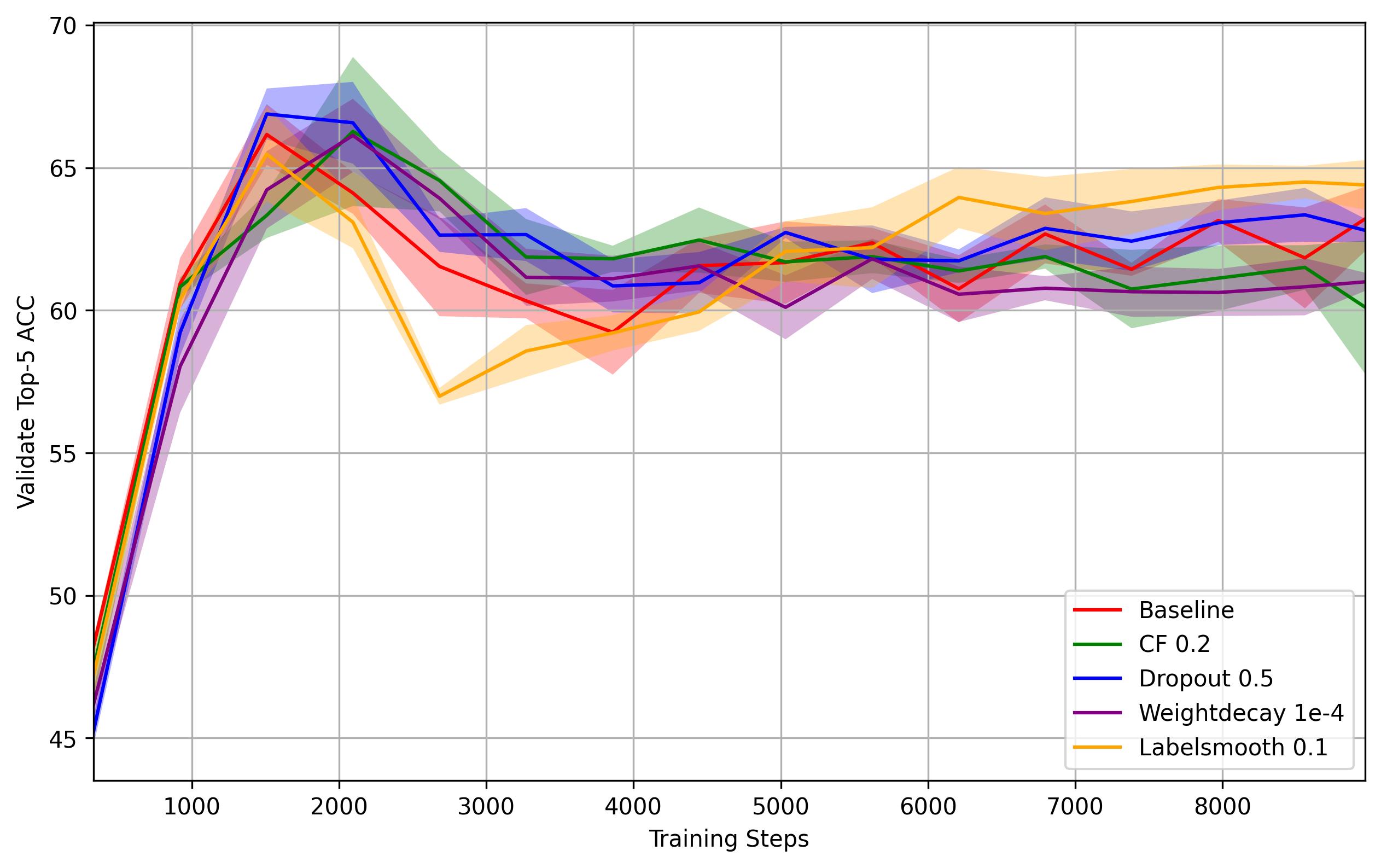} 
        \caption{Webvision-mini Top-5 ACC} 
        \label{fig:memo_noise}
    \end{subfigure} \hfill

    \caption{Performance of label smoothing, dropout, label smoothing, consistent feature, weight decay on CIFAR-100 and Webvision-mini.}
    \label{fig:compatibility}
\end{figure*}

\begin{figure*}[htbp]
    \centering
    
    \begin{subfigure}[b]{0.45\textwidth}
        \centering
        \includegraphics[width=\textwidth]{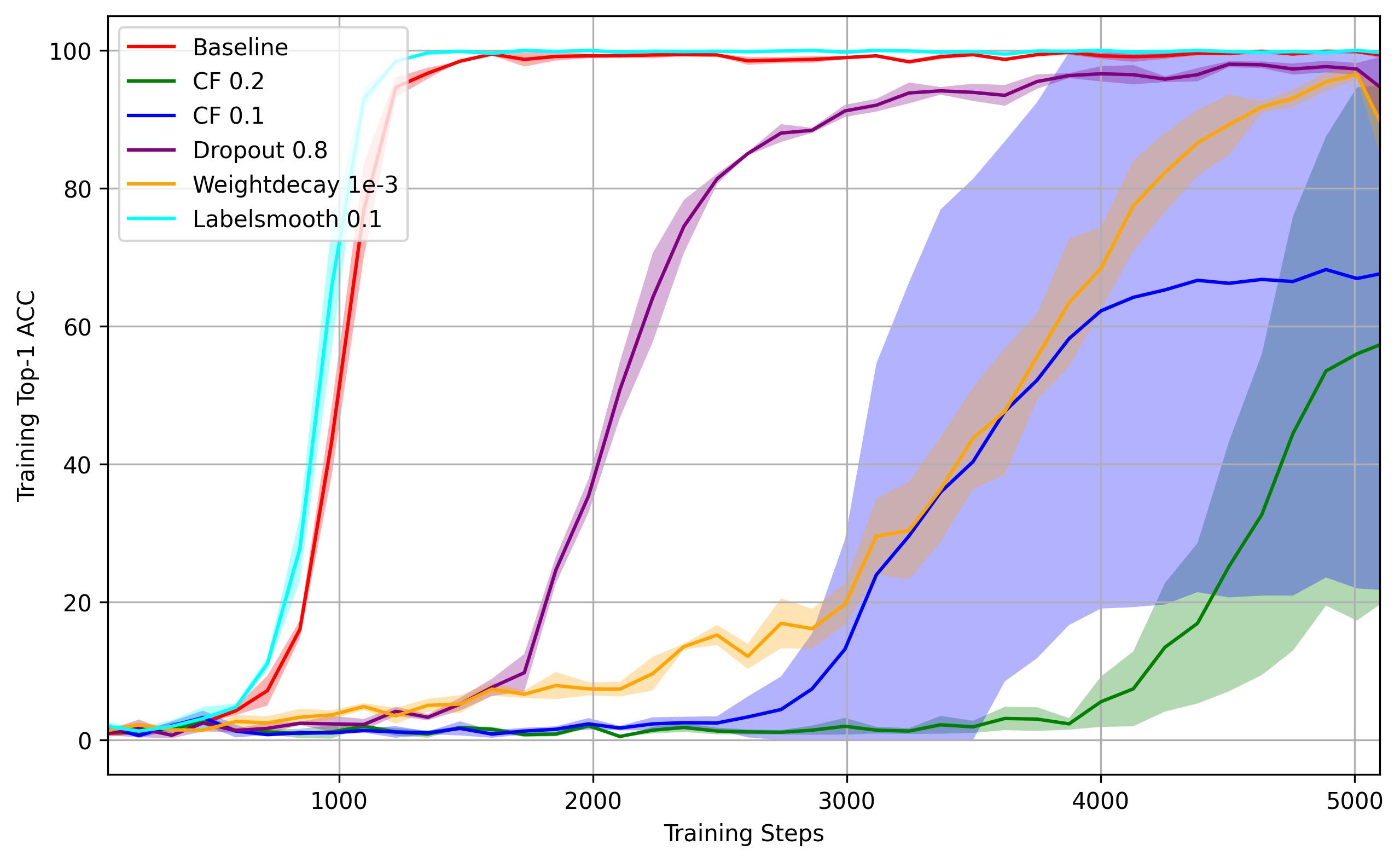} 
        \caption{} 
    \end{subfigure} \hfill
    \begin{subfigure}[b]{0.45\textwidth}
        \centering
        \includegraphics[width=\textwidth]{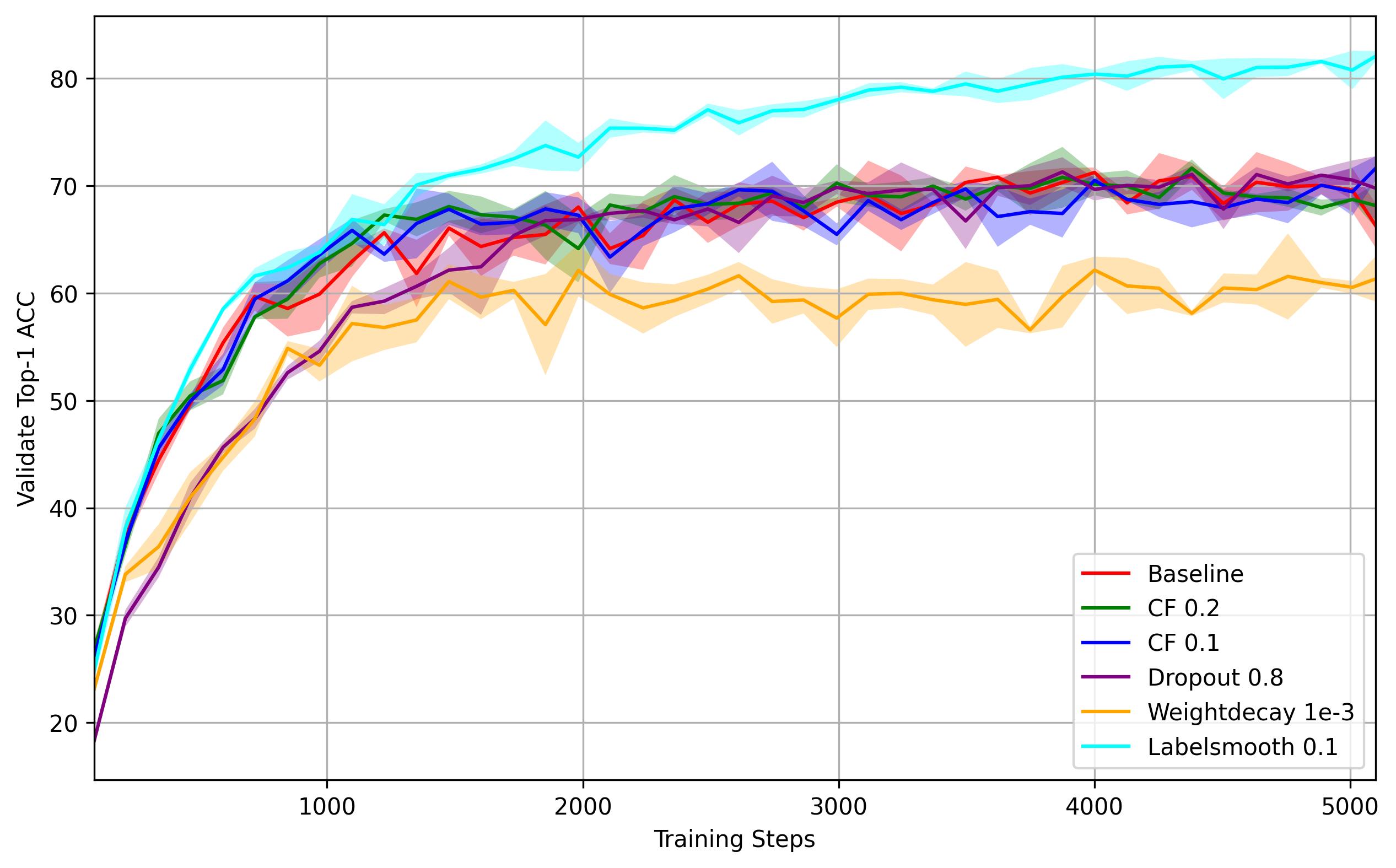} 
        \caption{} 
    \end{subfigure} \hfill
    \caption{Training accuracy (a) on the Flower102-Random dataset with different methods, and training accuracy (b) on the standard Flower102 dataset under the same settings, demonstrating the memory suppression ability and impact on normal convergence of different methods.}
    \label{fig:compatibility}
\end{figure*}

\begin{figure*}[htbp]
    \centering

    \begin{subfigure}[b]{0.32\textwidth}
        \centering
        \includegraphics[width=\textwidth]{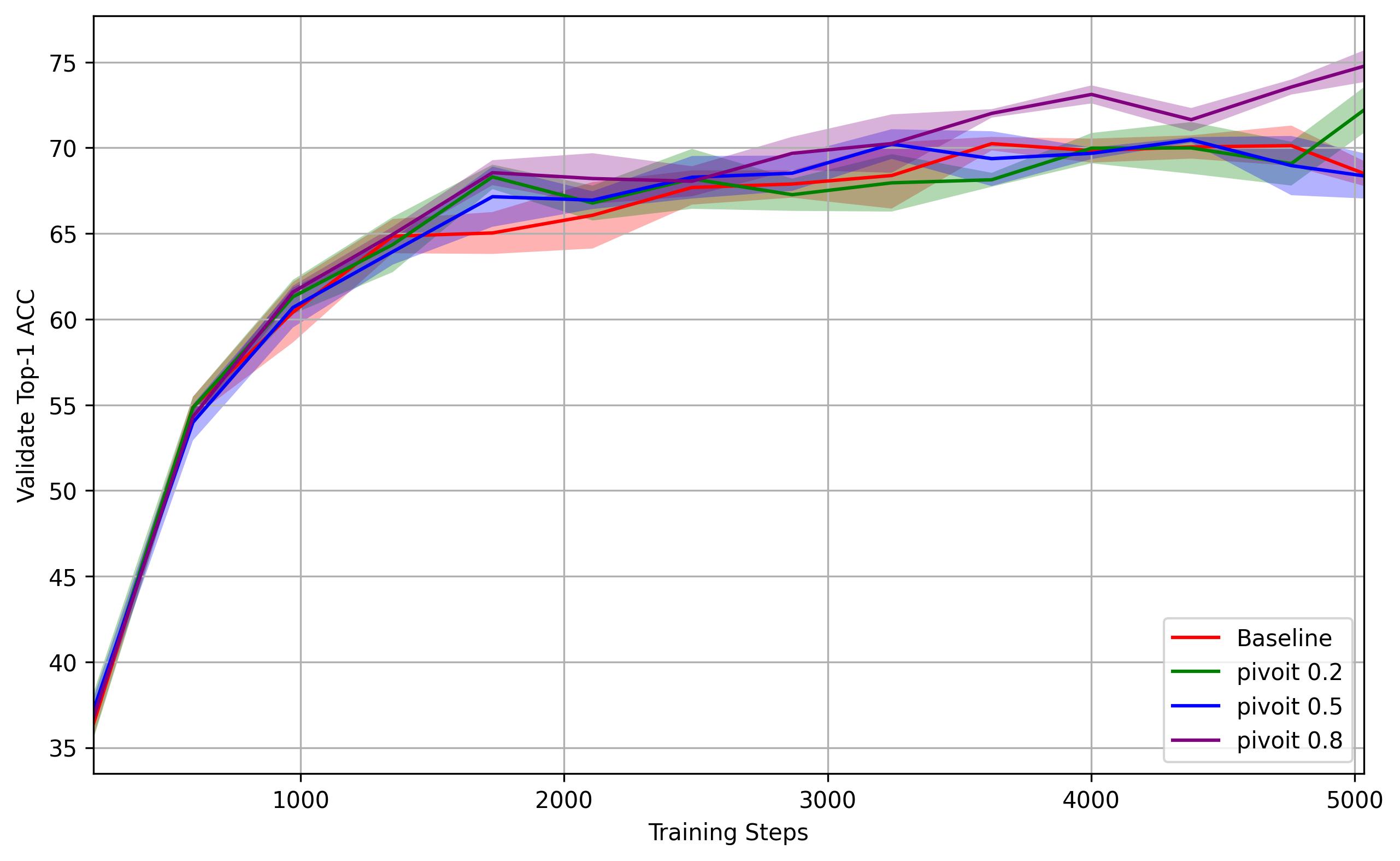} 
        \caption{Effect of parameter $p$ on Top-1 Accuracy} 
        \label{fig:p_acc}
    \end{subfigure} \hfill
    \begin{subfigure}[b]{0.32\textwidth}
        \centering
        \includegraphics[width=\textwidth]{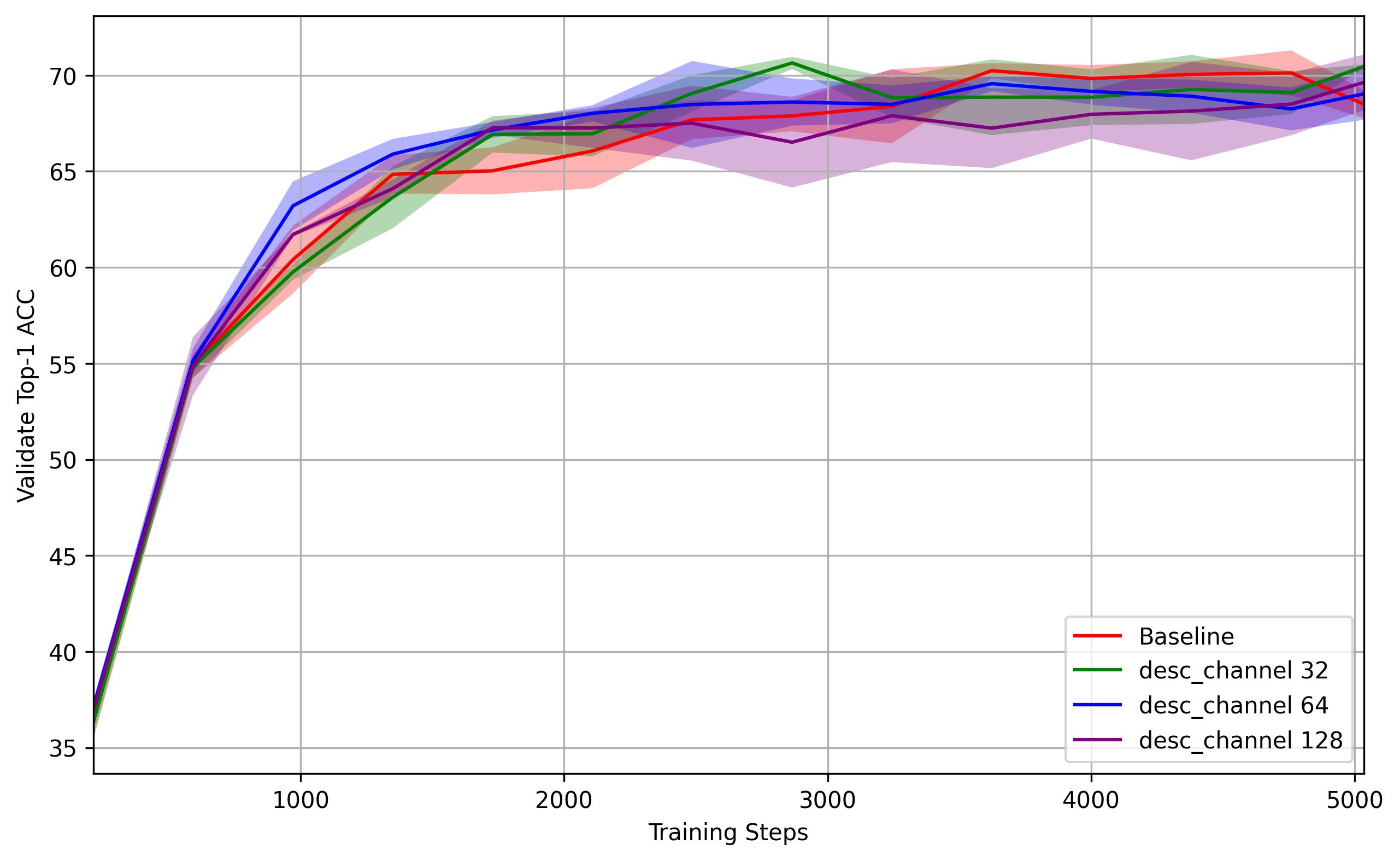} 
        \caption{Effect of parameter $desc$ on Top-1 Accuracy}
        \label{fig:desc_acc}
    \end{subfigure} \hfill
    \begin{subfigure}[b]{0.32\textwidth}
        \centering
        \includegraphics[width=\textwidth]{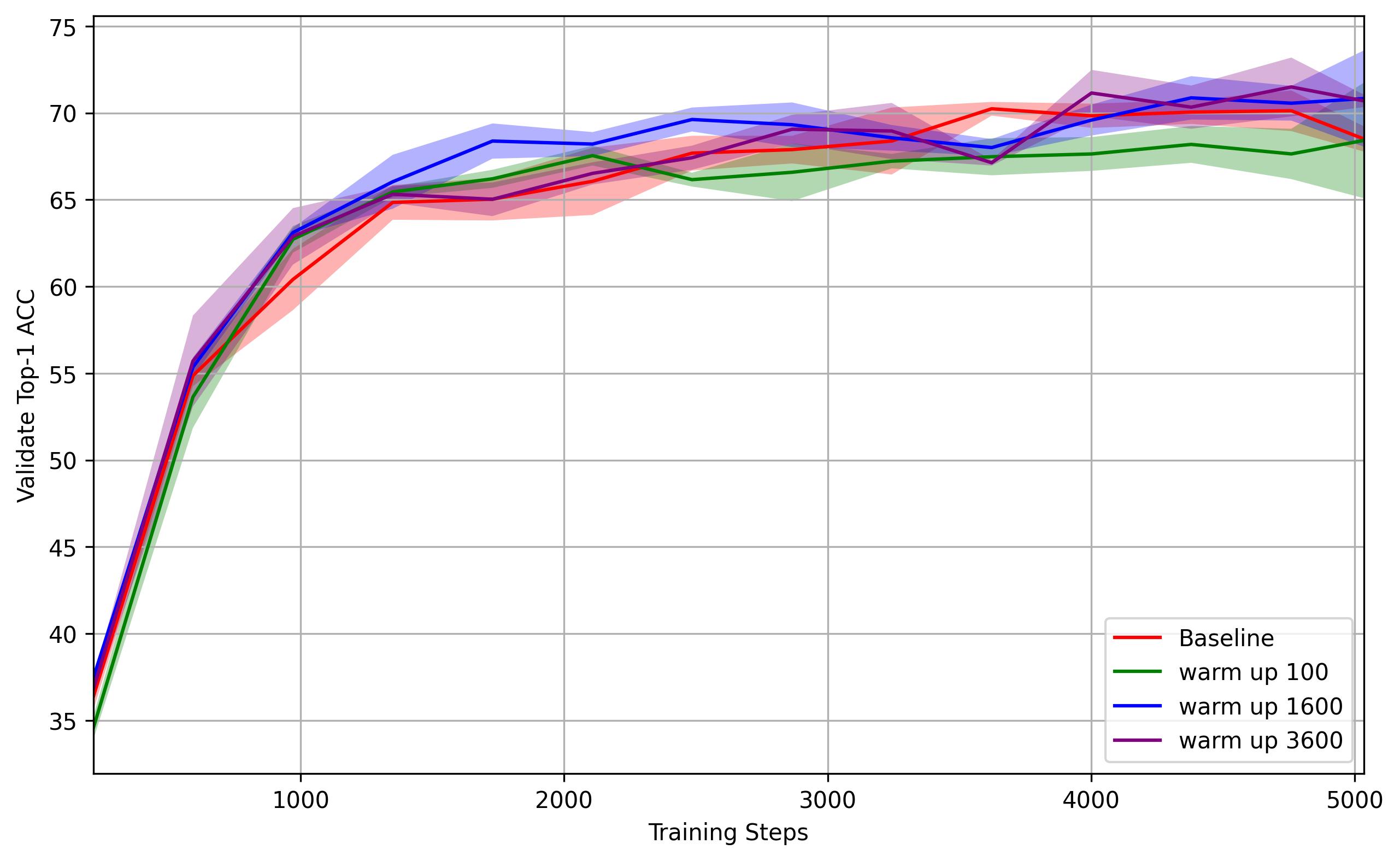} 
        \caption{Effect of parameter $warm$ on Top-1 Accuracy}
        \label{fig:warm_acc}
    \end{subfigure} \hfill

    \begin{subfigure}[b]{0.32\textwidth}
        \centering
        \includegraphics[width=\textwidth]{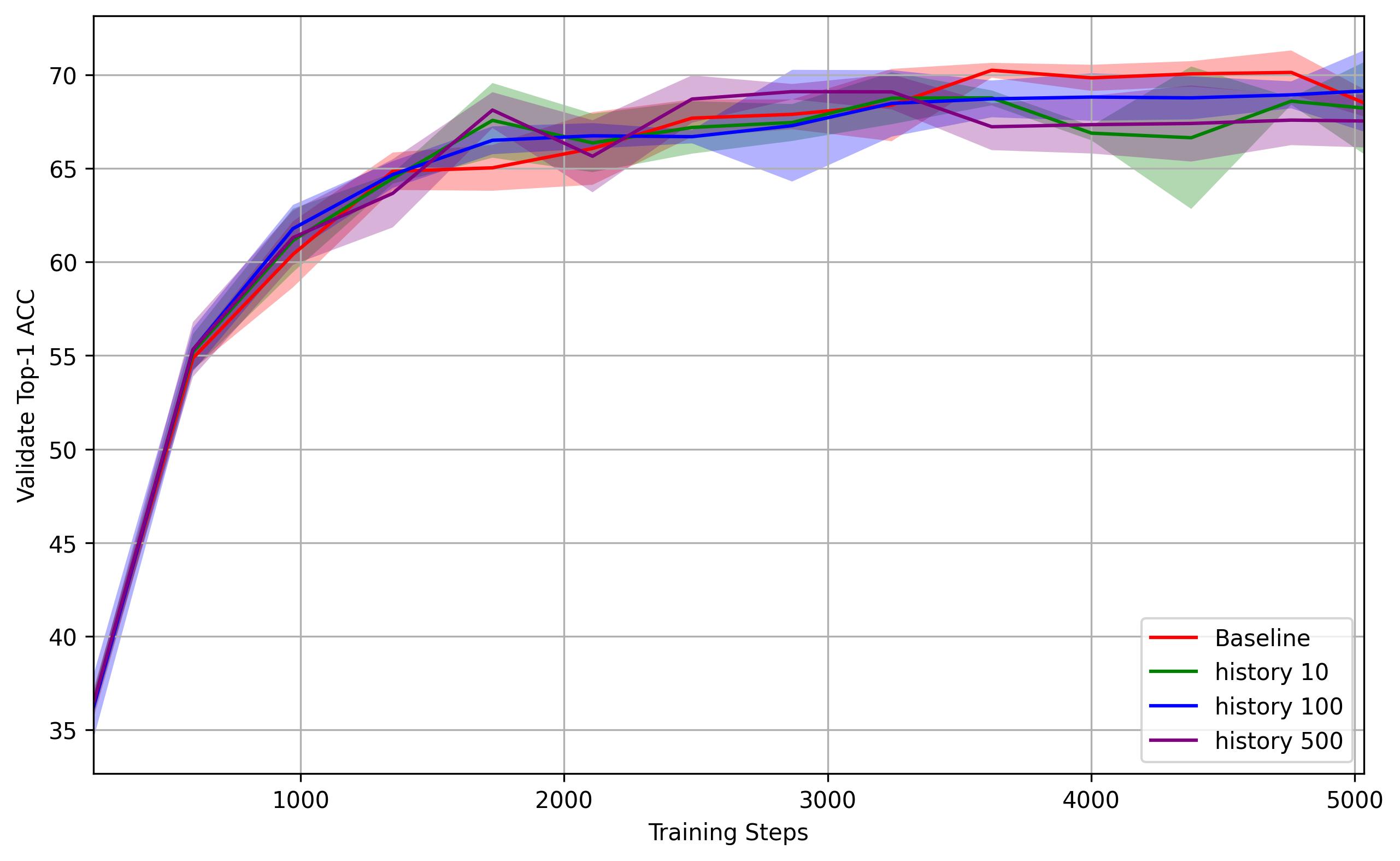} 
        \caption{Effect of parameter $history$ on Top-1 Accuracy}
        \label{fig:history_acc}
    \end{subfigure} \hfill
    \begin{subfigure}[b]{0.32\textwidth}
        \centering
        \includegraphics[width=\textwidth]{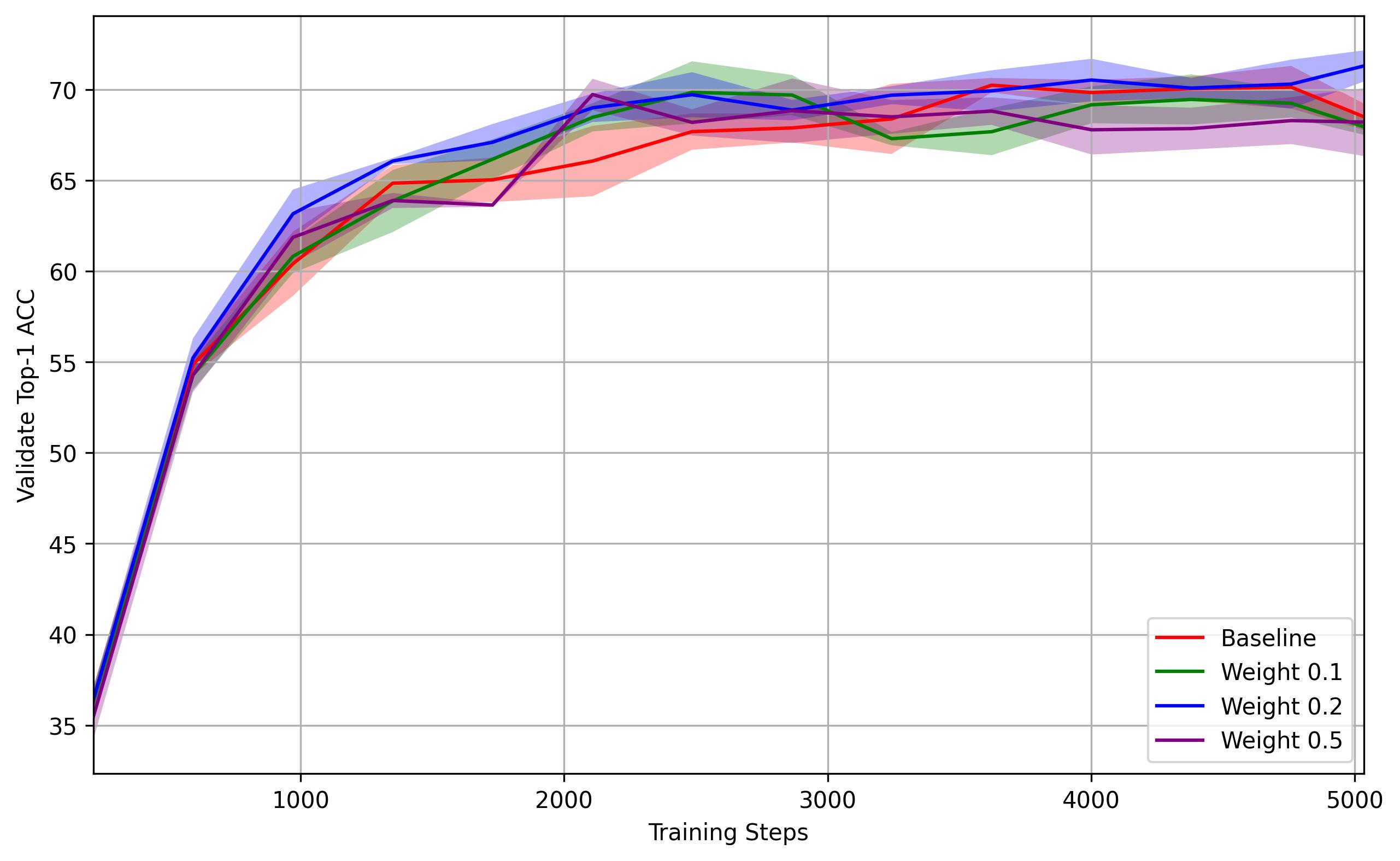} 
        \caption{Effect of parameter $weight$ on Top-1 Accuracy}
        \label{fig:weight_acc}
    \end{subfigure} \hfill
    \begin{subfigure}[b]{0.32\textwidth}
        \centering
        \includegraphics[width=\textwidth]{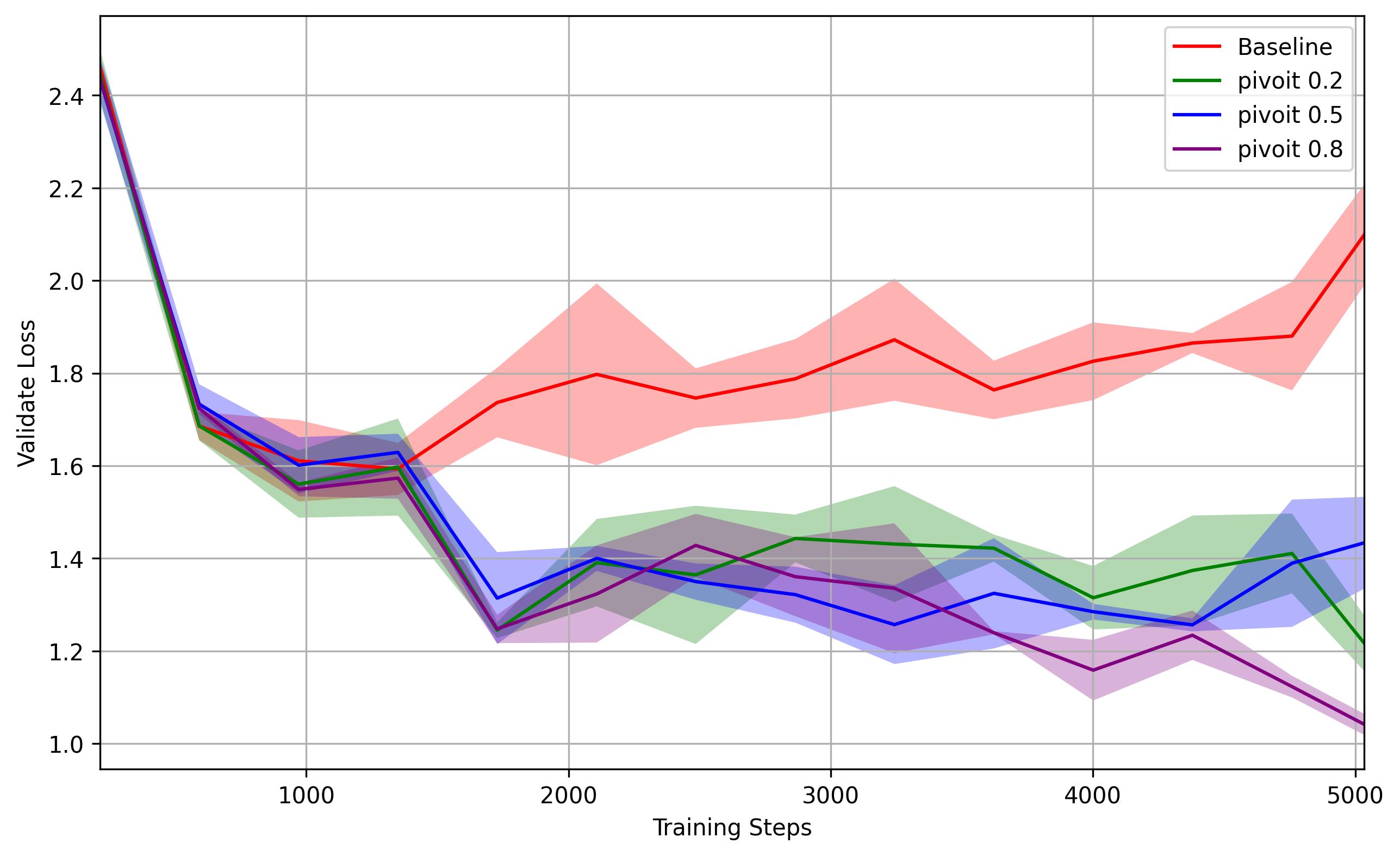} 
        \caption{Effect of parameter $p$ on Validation Loss}
        \label{fig:p_valloss}
    \end{subfigure} \hfill
    \begin{subfigure}[b]{0.32\textwidth}
        \centering
        \includegraphics[width=\textwidth]{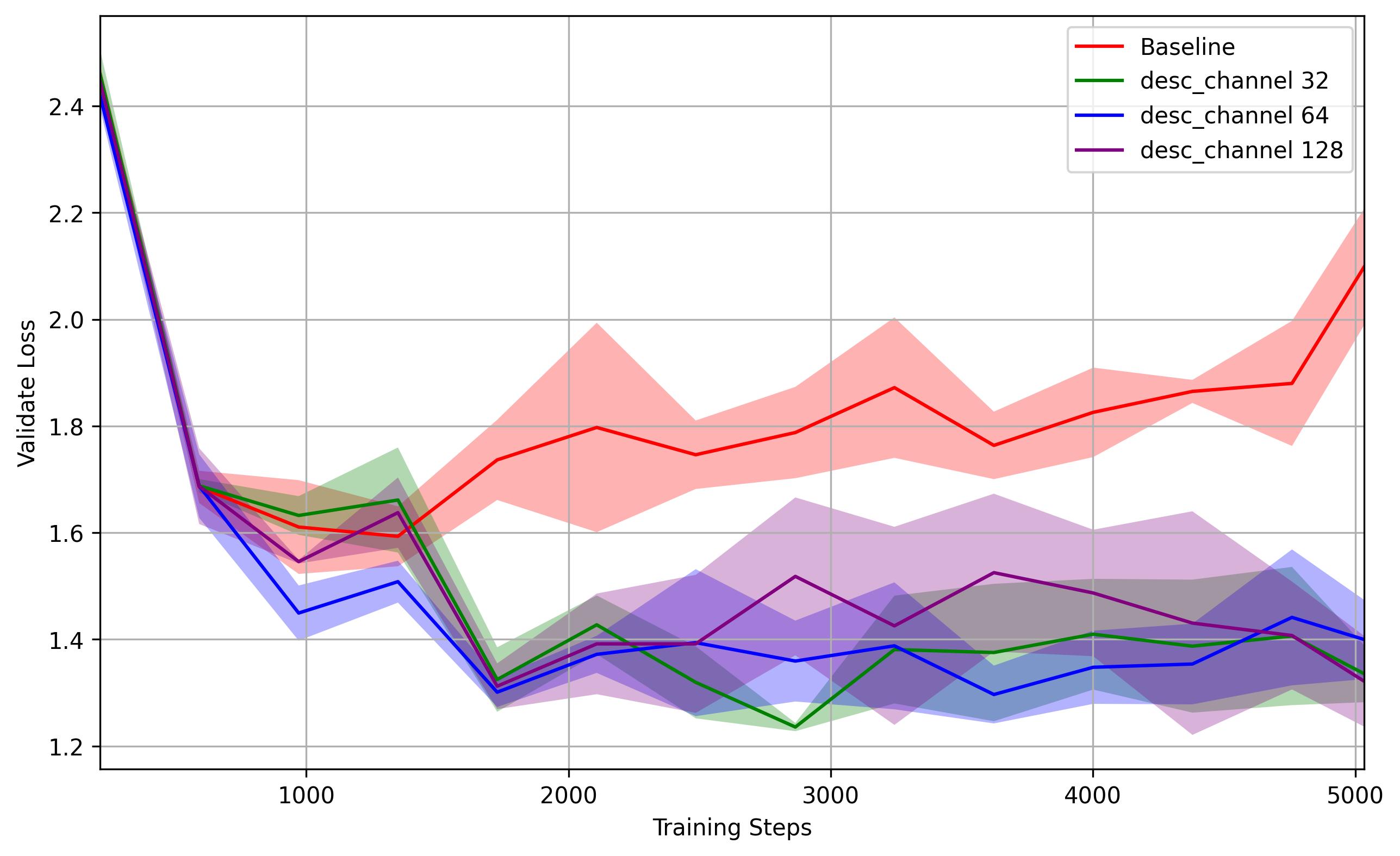} 
        \caption{Effect of parameter $desc$ on Validation Loss}
        \label{fig:desc_valloss}
    \end{subfigure} \hfill
    \begin{subfigure}[b]{0.32\textwidth}
        \centering
        \includegraphics[width=\textwidth]{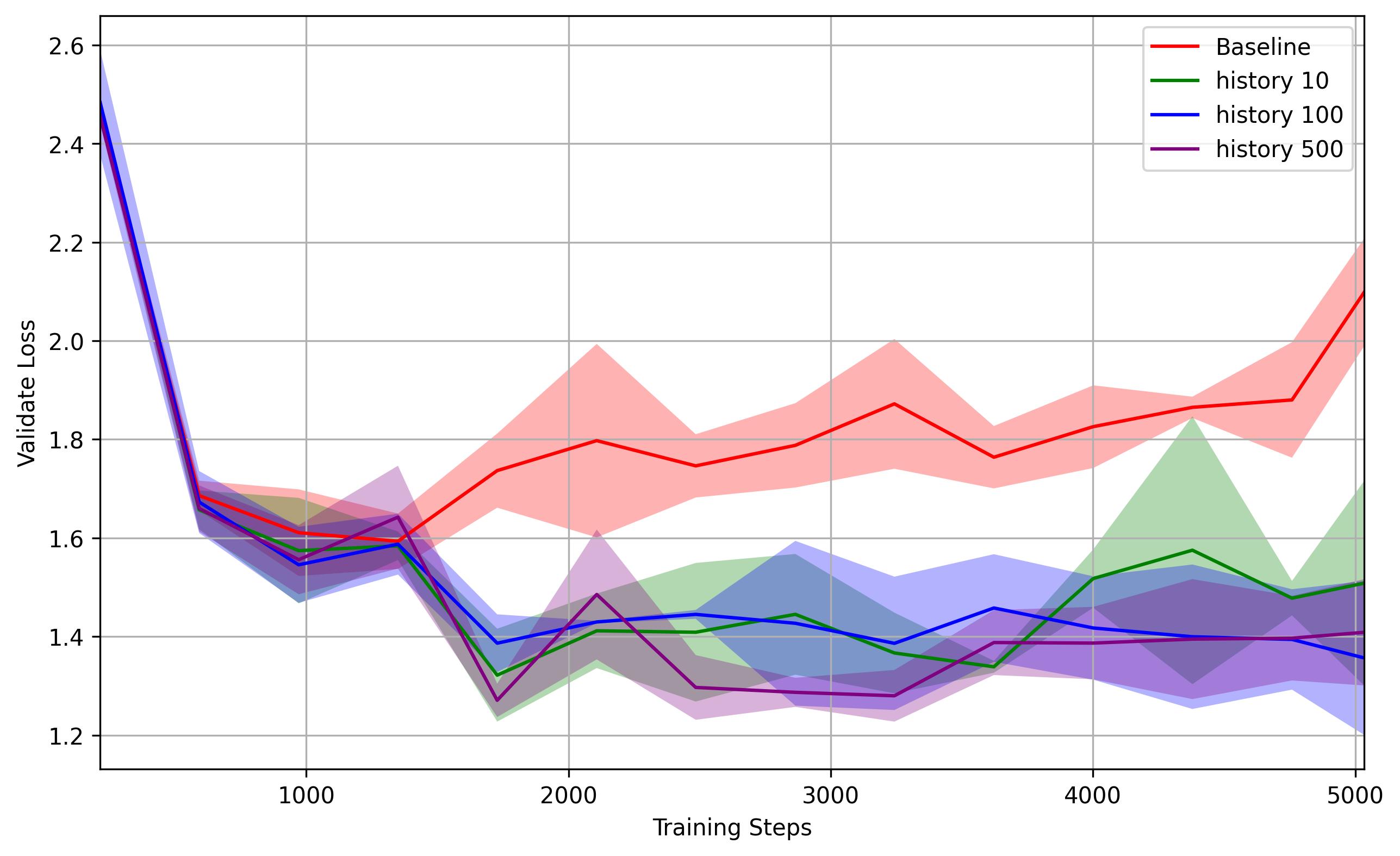} 
        \caption{Effect of parameter $history$ on Validation Loss}
        \label{fig:history_valloss}
    \end{subfigure} \hfill
    \begin{subfigure}[b]{0.32\textwidth}
        \centering
        \includegraphics[width=\textwidth]{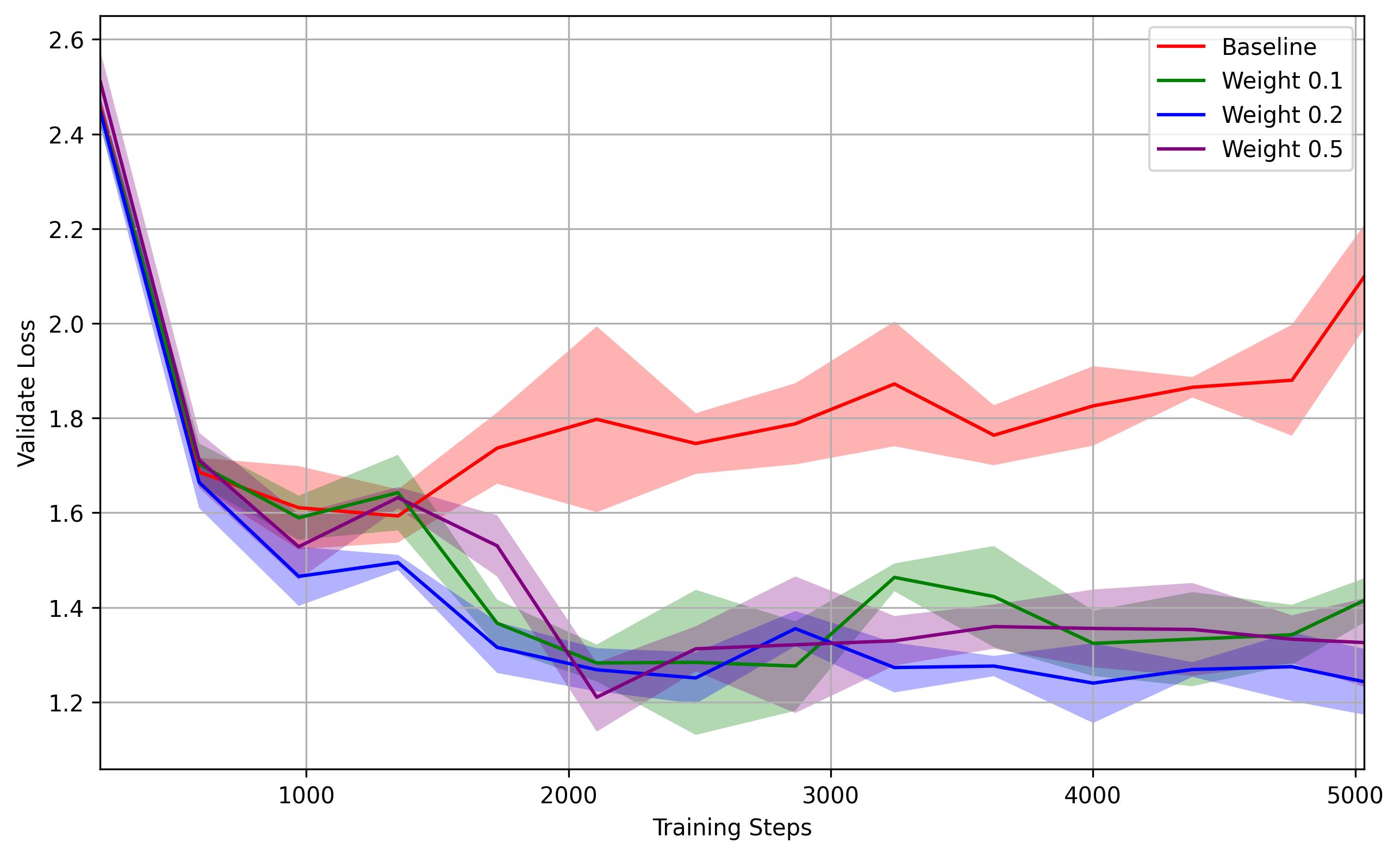} 
        \caption{Effect of parameter $weight$ on Validation Loss}
        \label{fig:weight_valloss}
    \end{subfigure} \hfill

    \caption{Ablation Study on CF parameters: Impact on Top-1 Accuracy and Validation Loss.}
    \label{fig:ablation}
\end{figure*}

\begin{figure*}[htbp]
    \centering
    \begin{subfigure}[b]{0.32\textwidth}
        \centering
        \includegraphics[width=\textwidth]{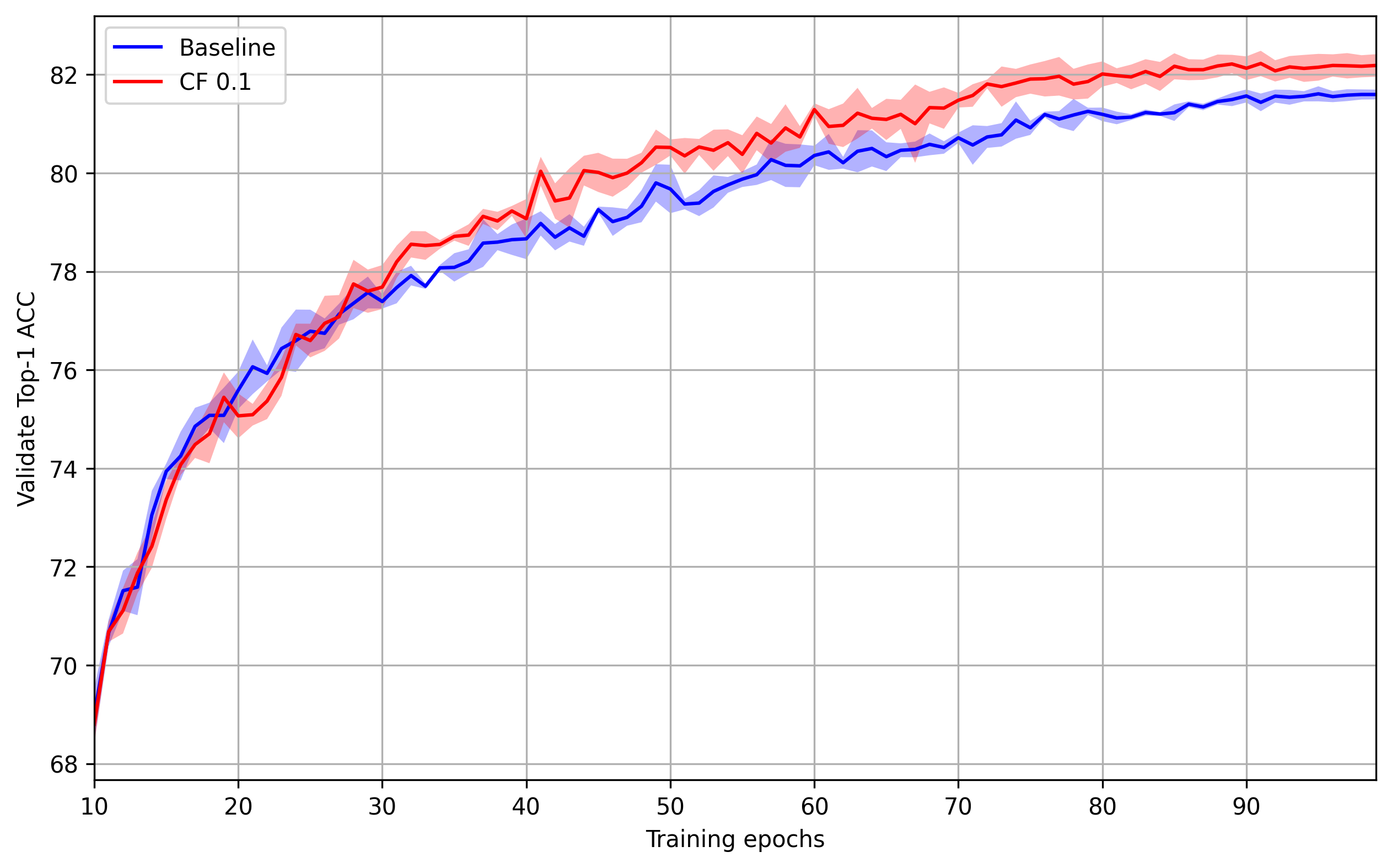} 
        \caption{Convnext\_Tiny Top-1 ACC} 
        \label{fig:memo_noise}
    \end{subfigure} \hfill
    \begin{subfigure}[b]{0.32\textwidth}
        \centering
        \includegraphics[width=\textwidth]{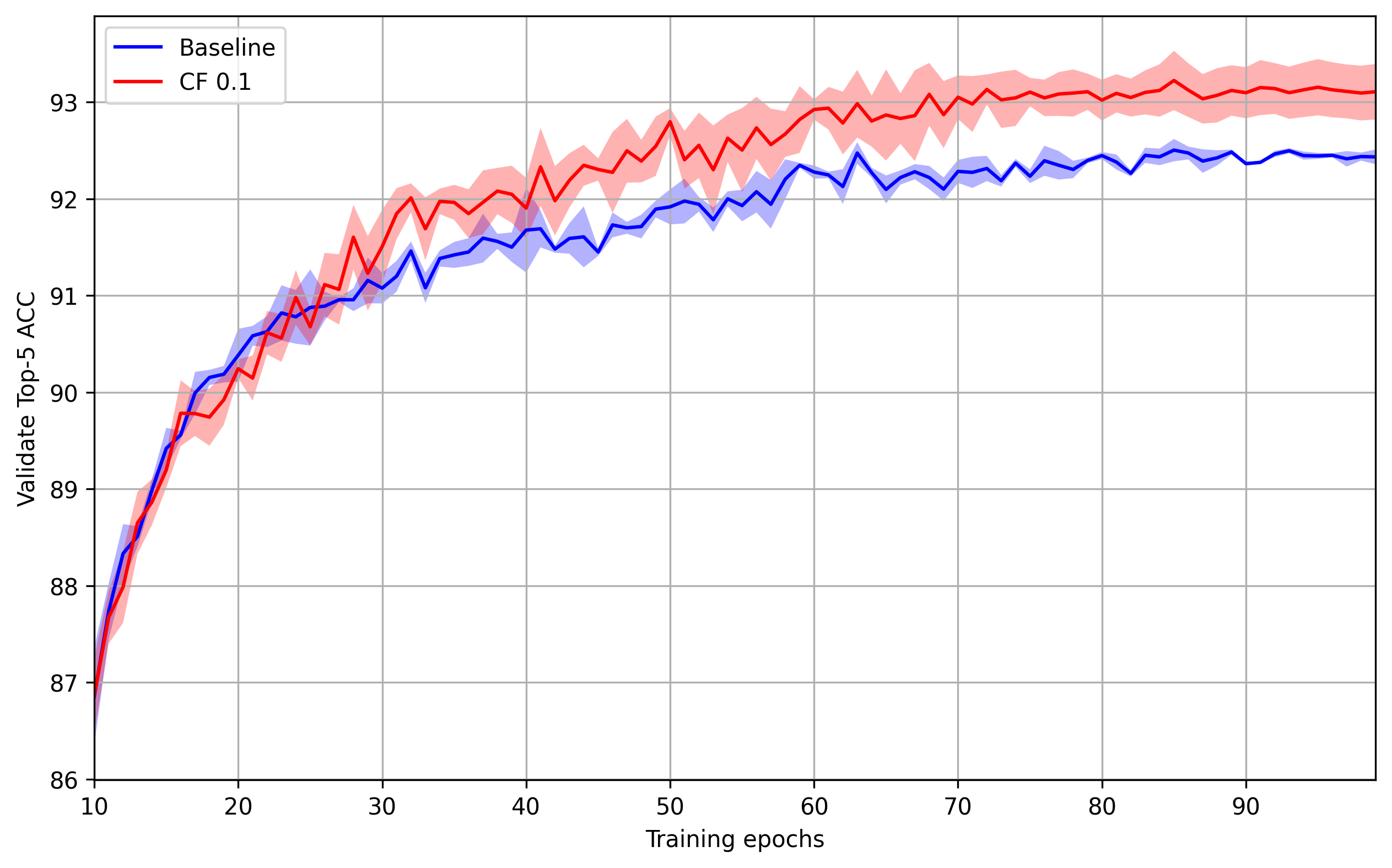} 
        \caption{Convnext\_Tiny Top-5 ACC} 
        \label{fig:memo_normal}
    \end{subfigure} \hfill
    \begin{subfigure}[b]{0.32\textwidth}
        \centering
        \includegraphics[width=\textwidth]{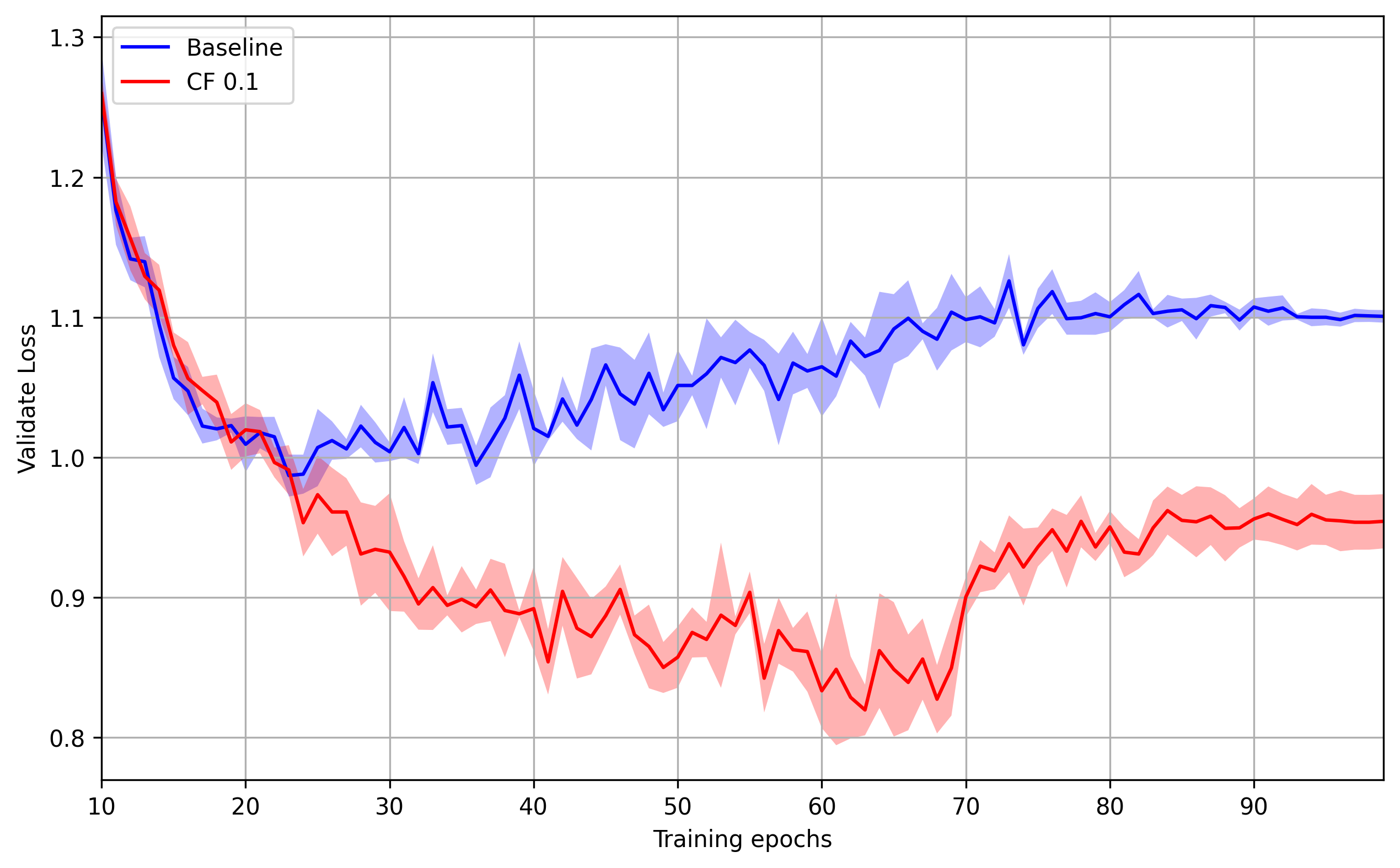} 
        \caption{Convnext\_Tiny Validate Loss} 
        \label{fig:warmup}
    \end{subfigure} \hfill
    \begin{subfigure}[b]{0.32\textwidth}
        \centering
        \includegraphics[width=\textwidth]{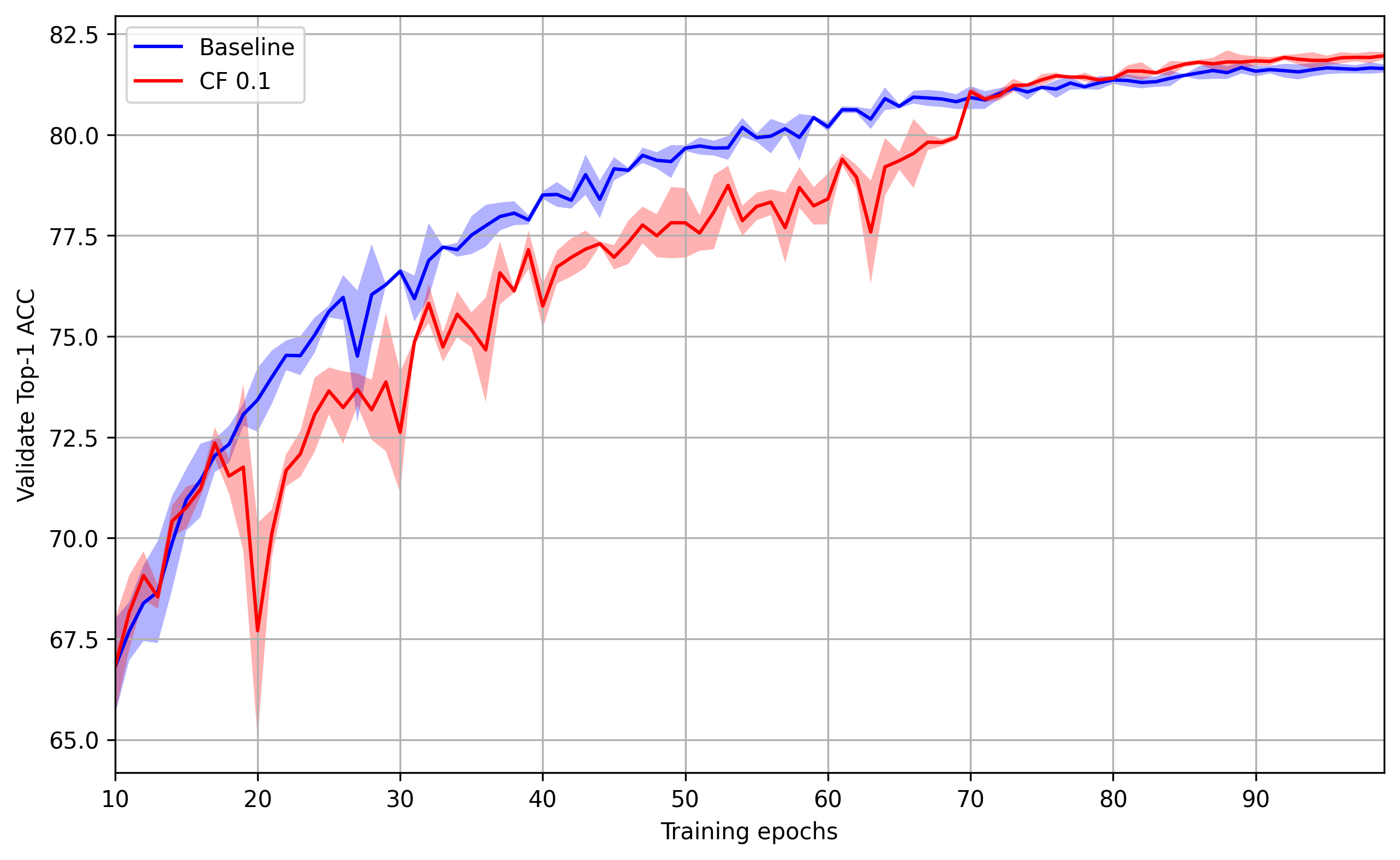} 
        \caption{MobileNetV3\_Large Top-1 ACC} 
        \label{fig:memo_noise}
    \end{subfigure} \hfill
    \begin{subfigure}[b]{0.32\textwidth}
        \centering
        \includegraphics[width=\textwidth]{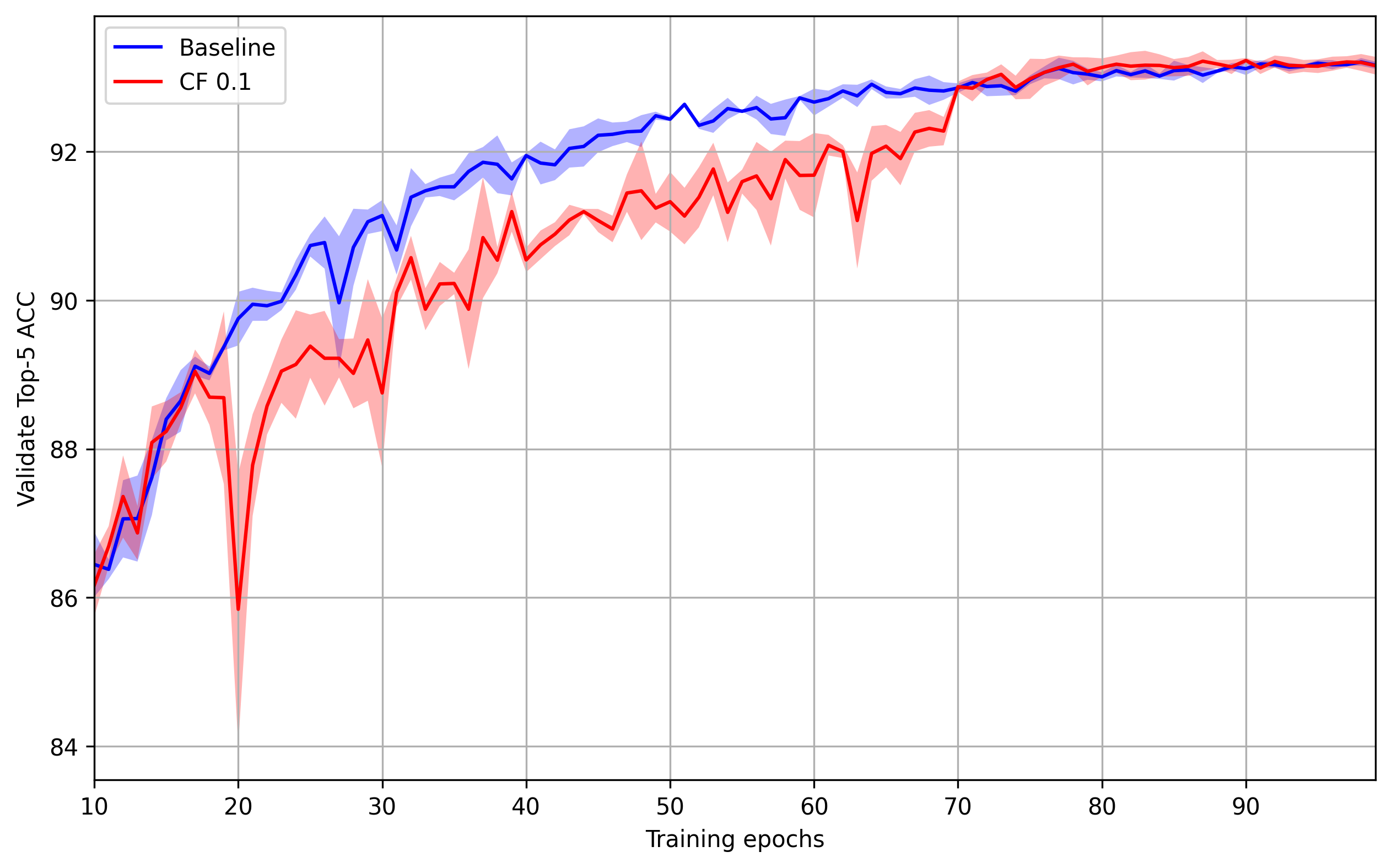} 
        \caption{MobileNetV3\_Large Top-5 ACC} 
        \label{fig:memo_normal}
    \end{subfigure} \hfill
    \begin{subfigure}[b]{0.32\textwidth}
        \centering
        \includegraphics[width=\textwidth]{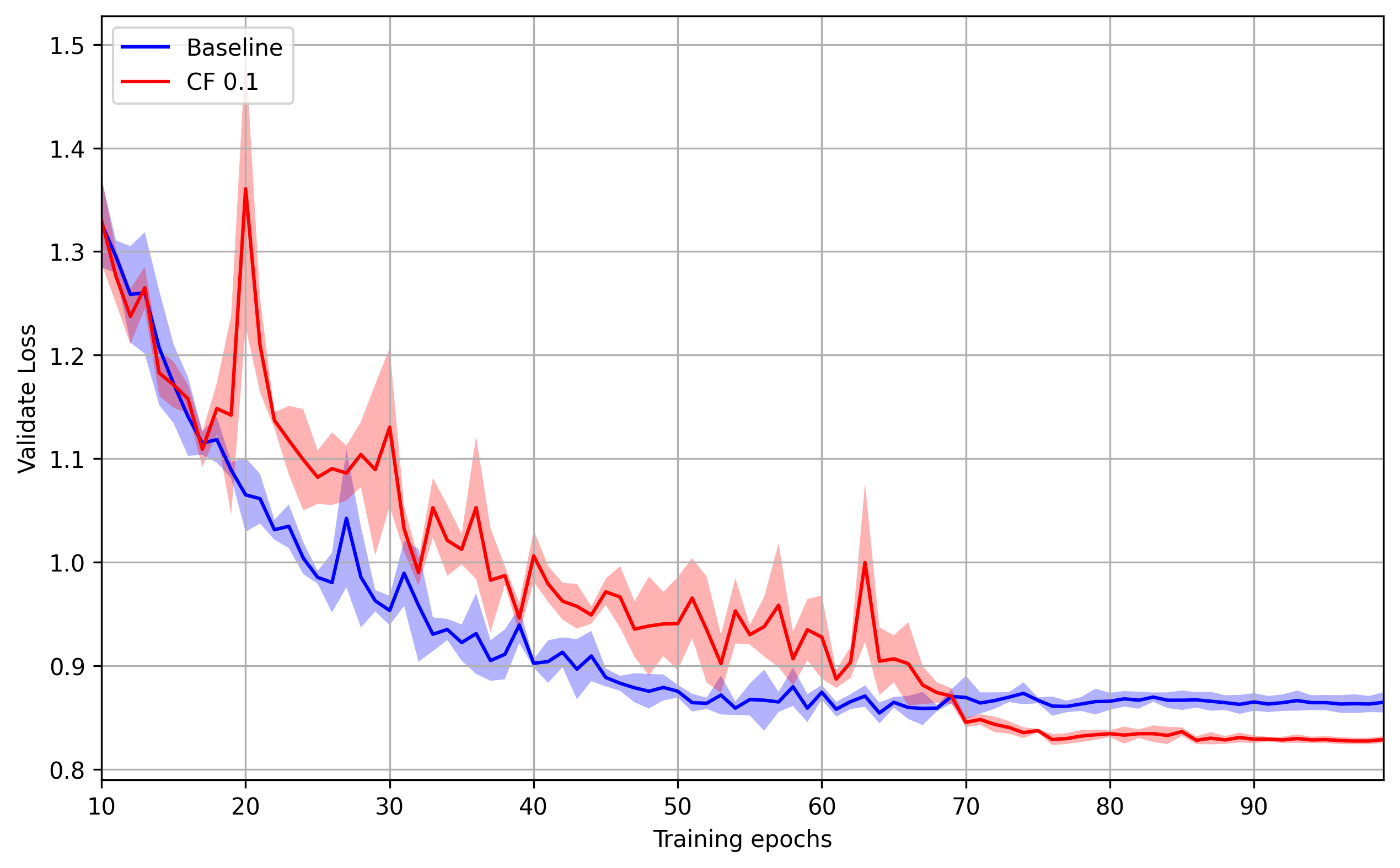} 
        \caption{MobileNetV3\_Large Validate Loss} 
        \label{fig:warmup}
    \end{subfigure} \hfill

    \begin{subfigure}[b]{0.32\textwidth}
        \centering
        \includegraphics[width=\textwidth]{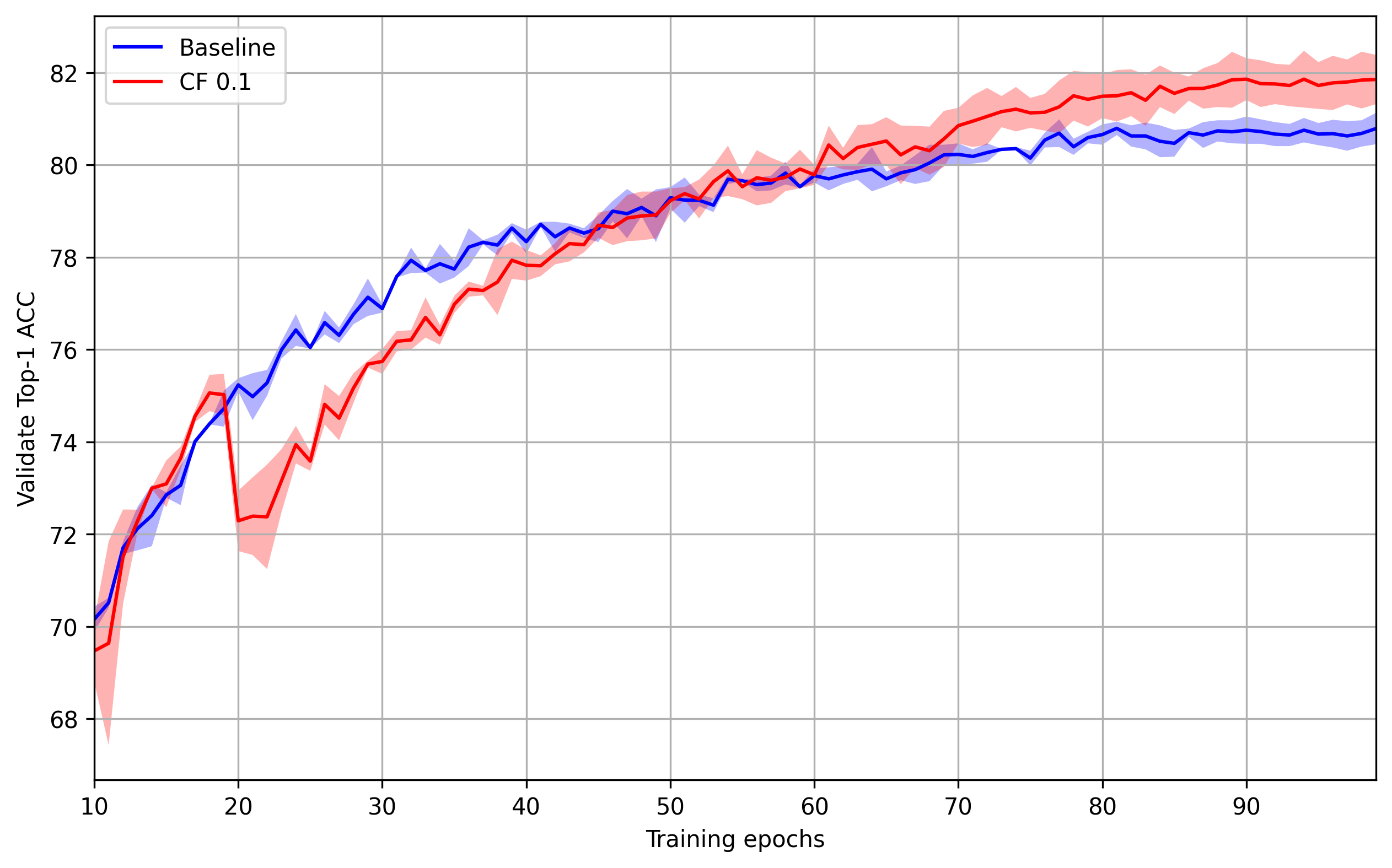} 
        \caption{ResNet18 Top-1 ACC} 
        \label{fig:memo_noise}
    \end{subfigure} \hfill
    \begin{subfigure}[b]{0.32\textwidth}
        \centering
        \includegraphics[width=\textwidth]{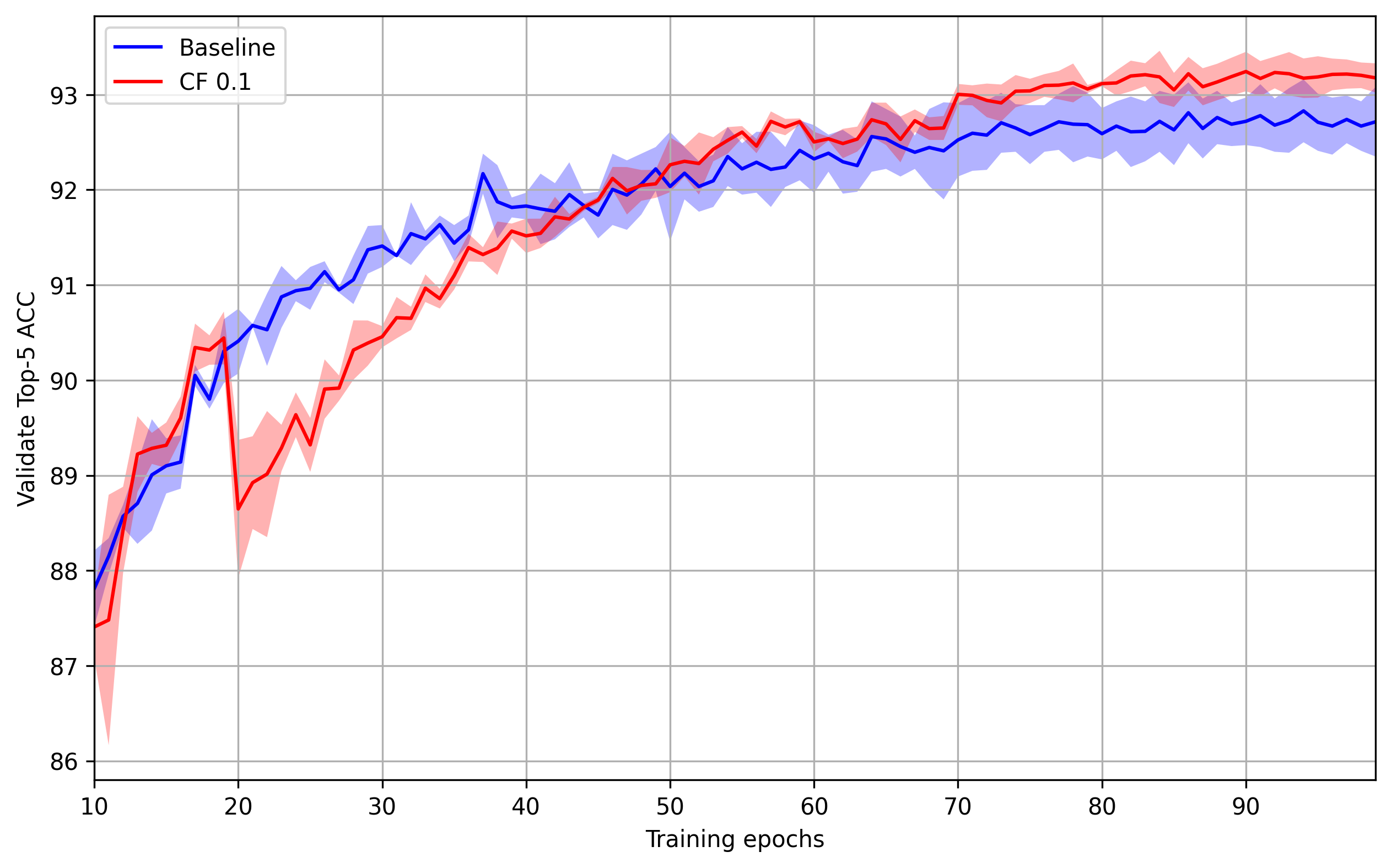} 
        \caption{ResNet18 Top-5 ACC} 
        \label{fig:memo_normal}
    \end{subfigure} \hfill
    \begin{subfigure}[b]{0.32\textwidth}
        \centering
        \includegraphics[width=\textwidth]{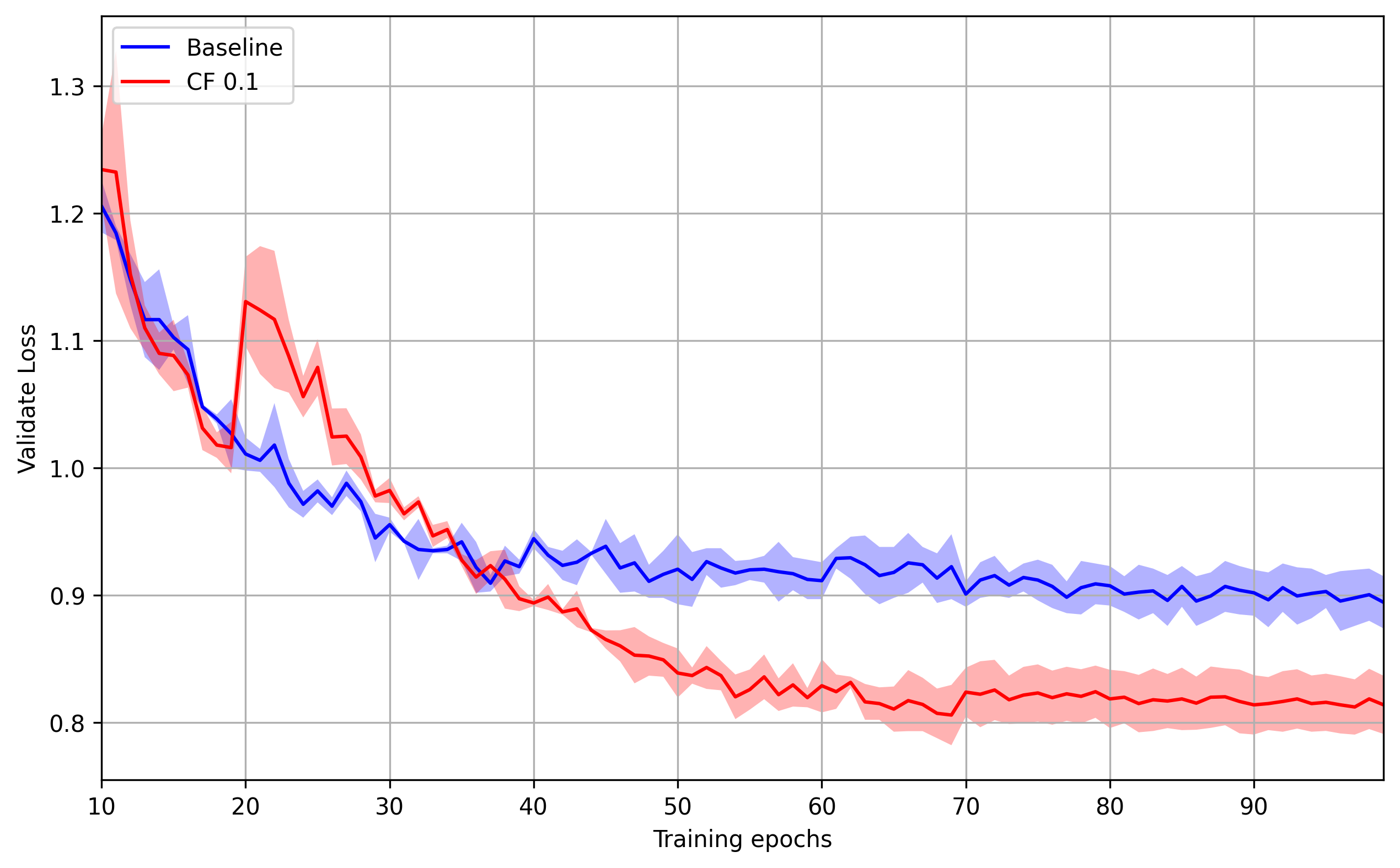} 
        \caption{ResNet18 Validate Loss} 
        \label{fig:warmup}
    \end{subfigure} \hfill

    \begin{subfigure}[b]{0.32\textwidth}
        \centering
        \includegraphics[width=\textwidth]{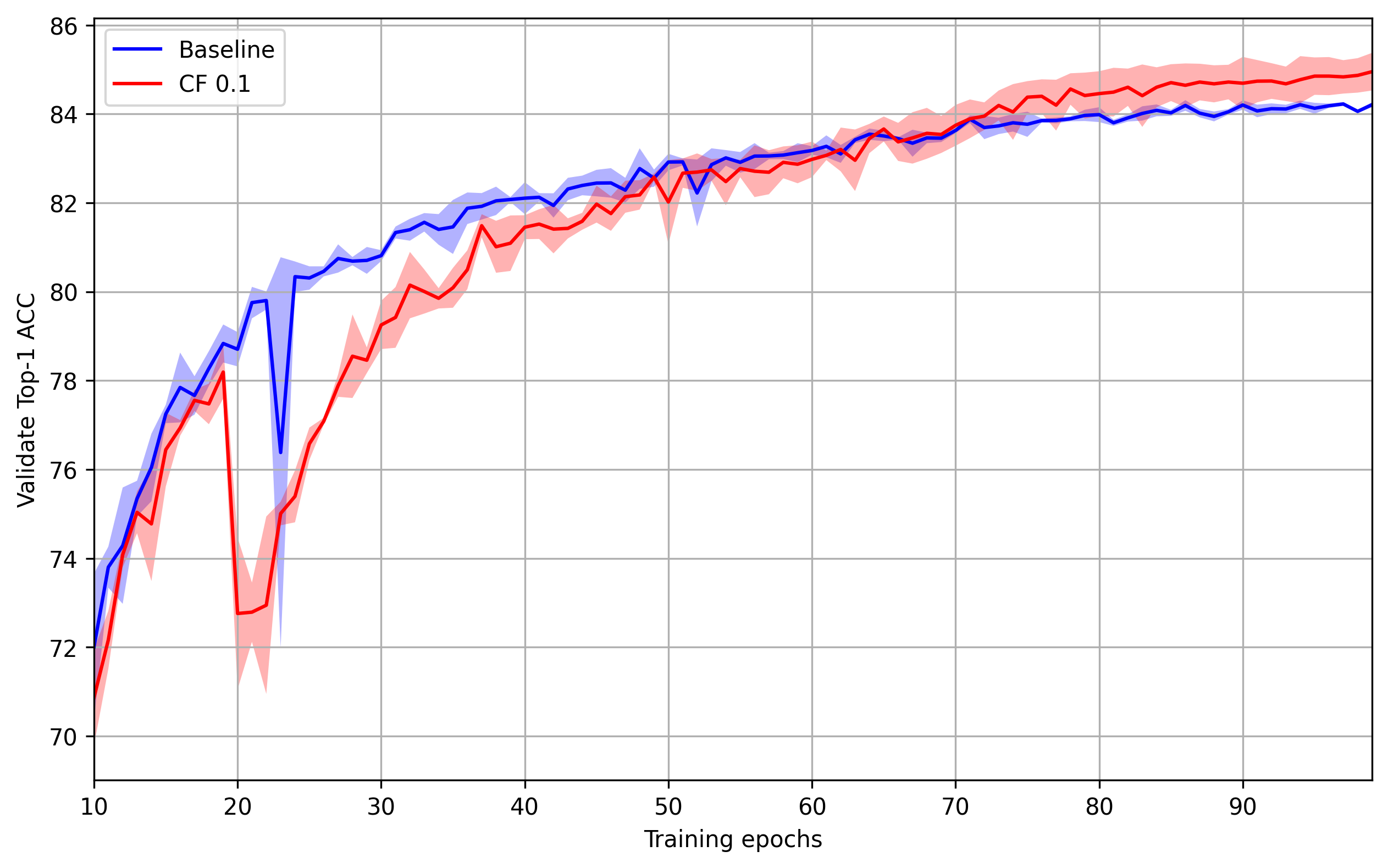} 
        \caption{ResNet50 Top-1 ACC} 
        \label{fig:memo_noise}
    \end{subfigure} \hfill
    \begin{subfigure}[b]{0.32\textwidth}
        \centering
        \includegraphics[width=\textwidth]{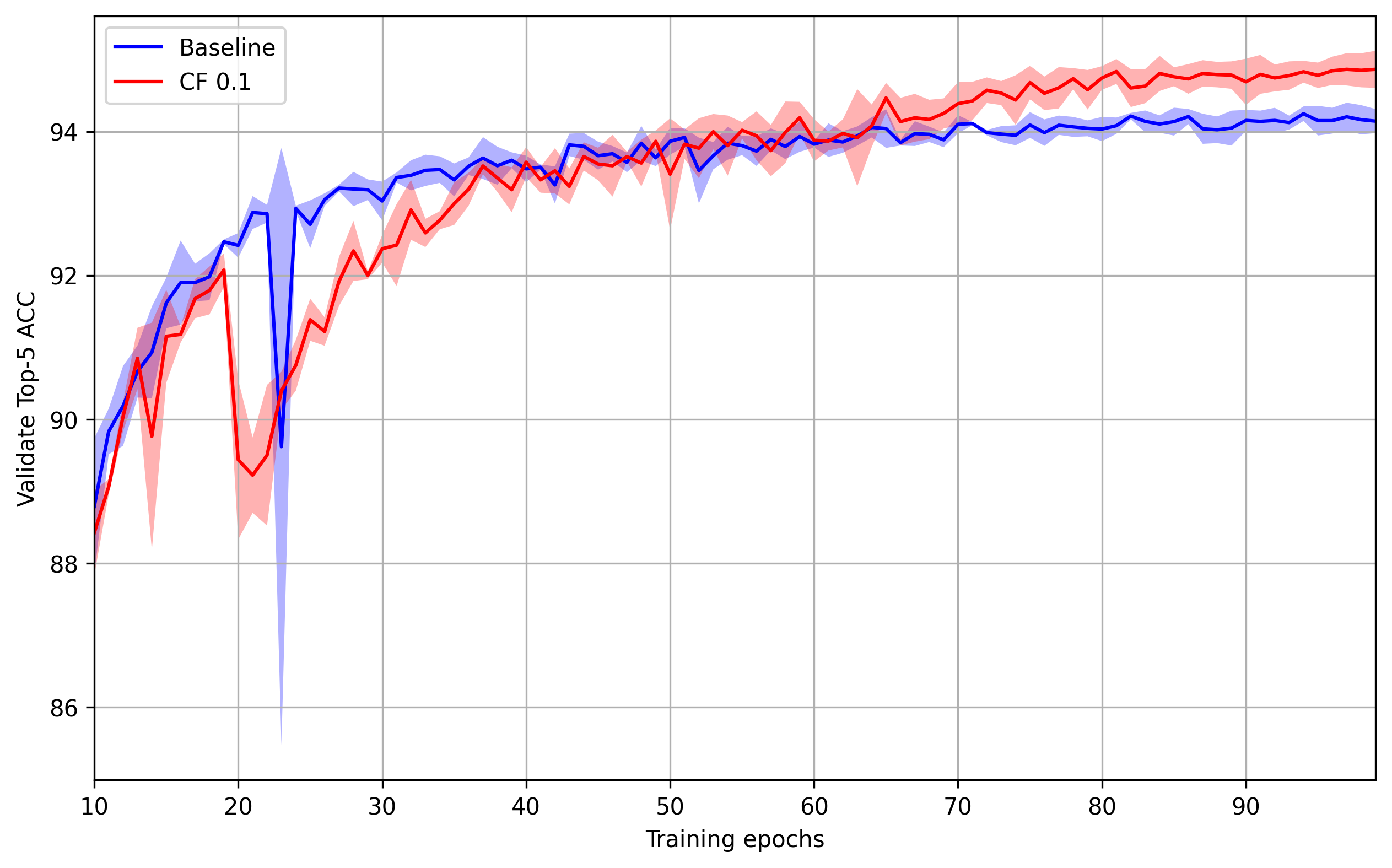} 
        \caption{ResNet50 Top-5 ACC} 
        \label{fig:memo_normal}
    \end{subfigure} \hfill
    \begin{subfigure}[b]{0.32\textwidth}
        \centering
        \includegraphics[width=\textwidth]{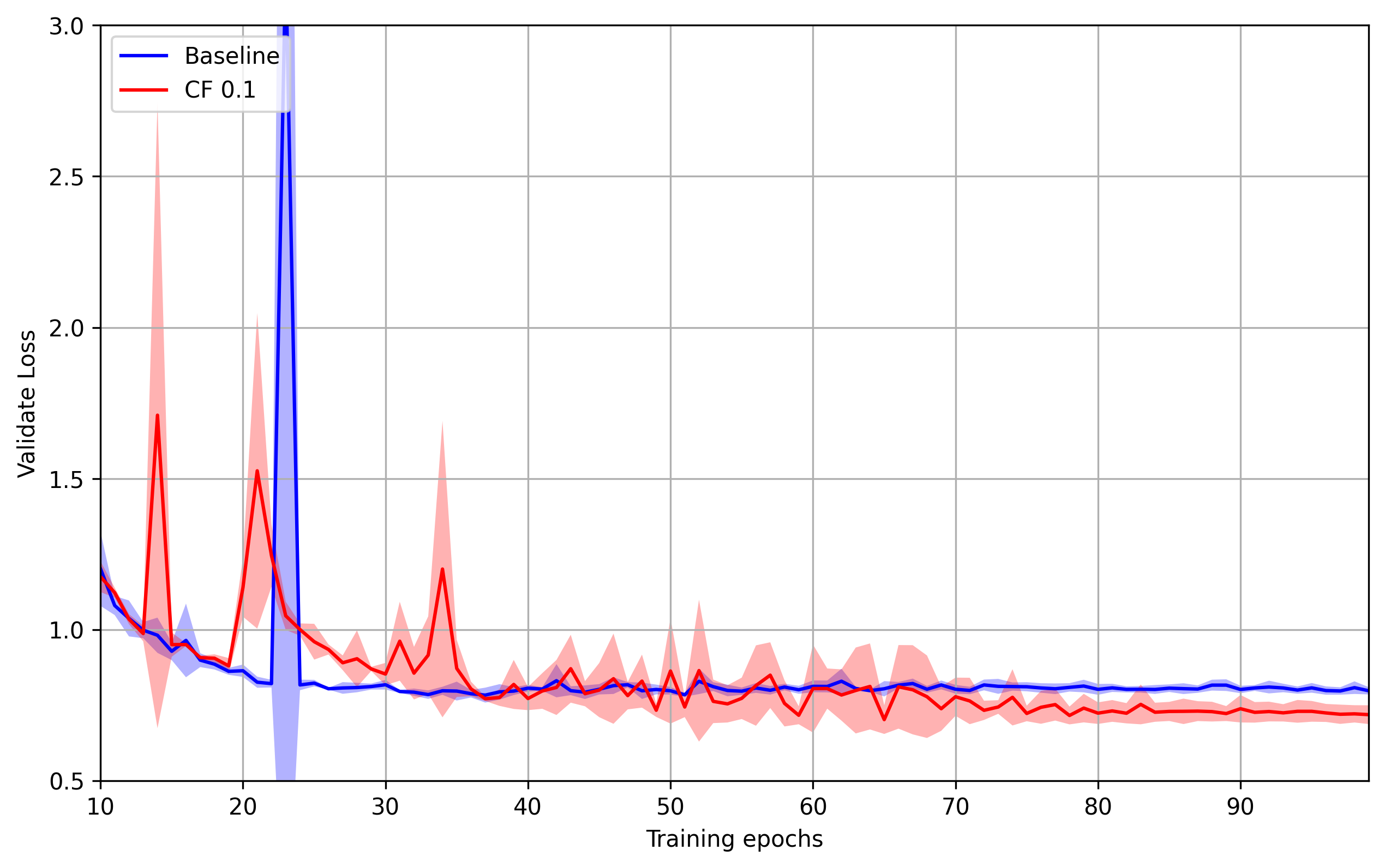} 
        \caption{ResNet50 Validate Loss} 
        \label{fig:warmup}
    \end{subfigure} \hfill
    
    \caption{Training Curves Across Different Model Architectures on ImageNet200.}
    \label{fig:compatibility}
\end{figure*}

\begin{table*}[h]
\centering
\begin{tabular}{@{}ll@{}}
\toprule
\textbf{Hyperparameters} & \textbf{Values} \\ \midrule
Number of nodes       & 1               \\
GPUs per node         & 2               \\ \midrule
Training epochs          & 100             \\
Optimizer                & Adam            \\
Learning rate        & 0.001           \\
Adam $\beta$             & (0.9, 0.999)    \\
Learning rate schedule   & CosineAnnealingLR \\
Minimal learning rate    & 0               \\ \midrule
Random resized crop      & Size=(224, 224), Scale=(0.08, 1.0), Ratio=(0.75, 1.3333) \\
Horizontal flip          & Probability=0.5 \\
Normalize mean           & (0.485, 0.456, 0.406) \\
Normalize std            & (0.229, 0.224, 0.225) \\
Validation resize size   & 256             \\
Validation crop size     & 224             \\
Input resolution         & $224 \times 224$ \\ \midrule
CF warm up epochs         & 20 \\
CF shut-off epochs        & 70  \\

\bottomrule
\end{tabular}
\caption{Hyperparameters for training models on ImageNet200}

\end{table*}



 \fi

\end{document}